\documentclass{article}

\usepackage{PRIMEarxiv}
\usepackage{titletoc}
\usepackage[page,header]{appendix}

\usepackage[utf8]{inputenc} 
\usepackage[T1]{fontenc}    
\usepackage{hyperref}       
\usepackage{url}            
\usepackage{booktabs}       
\usepackage{amsfonts}       
\usepackage{nicefrac}       
\usepackage{microtype}      
\usepackage{lipsum}
\usepackage{fancyhdr}       
\usepackage{graphicx}       

\usepackage{wrapfig}
\usepackage{amsmath}
\usepackage{float}
\usepackage{multicol}
\usepackage{algorithm}
\usepackage{algorithmic}
\usepackage{amsthm}
\usepackage{subfigure}
\usepackage{makecell}
\usepackage{multirow}

\theoremstyle{plain}
\newtheorem{theorem}{Theorem}[section]

\theoremstyle{definition}

\theoremstyle{remark}

\title{Value-Based Deep Multi-Agent Reinforcement Learning with Dynamic Sparse Training}
\author{
  Pihe Hu$^\ast$ \\
  Tsinghua University\\
  Beijing, China\\
  \texttt{hupihe@gmail.com} \\
  \And
  Shaolong Li\thanks{Equal contribution.} \\
  Central South University\\
  Changsha, China\\
  \texttt{shaolongli16@gmail.com} \\
  \And
  Zhuoran Li\\
  Tsinghua University\\
  Beijing, China\\
  \texttt{lizr20@mails.tsinghua.edu.cn}
  \And
  Ling Pan\\
  Hong Kong University of\\
  Science and Technology\\
  Hong Kong, China\\
  \texttt{lingpan@ust.hk}
  \And
  Longbo Huang\thanks{Corresponding author.}\\
  Tsinghua University\\
  Beijing, China\\
  \texttt{longbohuang@tsinghua.edu.cn}
}

\begin{document}
\maketitle
\begin{abstract}
Deep Multi-agent Reinforcement Learning (MARL) relies on neural networks with numerous parameters in multi-agent scenarios, often incurring substantial computational overhead. Consequently, there is an urgent need to expedite training and enable model compression in MARL. This paper proposes the utilization of dynamic sparse training (DST), a technique proven effective in deep supervised learning tasks, to alleviate the computational burdens in MARL training. However, a direct adoption of DST fails to yield satisfactory MARL agents, leading to breakdowns in value learning within deep sparse value-based MARL models. Motivated by this challenge, we introduce an innovative Multi-Agent Sparse Training (MAST) framework aimed at simultaneously enhancing the reliability of learning targets and the rationality of sample distribution to improve value learning in sparse models. Specifically, MAST incorporates the Soft Mellowmax Operator with a hybrid TD-($\lambda$) schema to establish dependable learning targets. Additionally, it employs a dual replay buffer mechanism to enhance the distribution of training samples. Building upon these aspects, MAST utilizes gradient-based topology evolution to exclusively train multiple MARL agents using sparse networks. Our comprehensive experimental investigation across various value-based MARL algorithms on multiple benchmarks demonstrates, for the first time, significant reductions in redundancy of up to $20\times$ in Floating Point Operations (FLOPs) for both training and inference, with less than $3\%$ performance degradation.
\end{abstract}

\section{Introduction}
Multi-agent reinforcement learning (MARL) \cite{shoham2008multiagent}, coupled with deep neural networks, has not only revolutionized artificial intelligence but also showcased remarkable success across a wide range of critical applications. From mastering multi-agent video games such as Quake III Arena \cite{jaderberg2019human}, StarCraft II \cite{mathieu2021starcraft}, Dota 2 \cite{berner2019dota}, and Hide and Seek \cite{baker2019emergent} to guiding autonomous robots through intricate real-world environments \cite{shalev2016safe, da2017simultaneously, chen2020autonomous}, deep MARL has emerged as an indispensable and versatile tool for addressing complex, multifaceted challenges. Its unique capability to capture intricate interactions and dependencies among multiple agents has spurred novel solutions, solidifying its position as a transformative paradigm across various domains \cite{zhang2021multi, albrecht2023multi}. 

However, the exceptional success of deep MARL comes at a considerable computational cost. Training these agents involves the intricate task of adapting neural networks to accommodate an expanded parameter space, especially in scenarios with a substantial number of agents. For instance, the training regimen for AlphaStar \cite{mathieu2021starcraft}, tailored for StarCraft II, extended over a grueling 14-day period, employing 16 TPUs per agent. Similarly, the OpenAI Five \cite{berner2019dota} model for Dota 2 underwent an extensive training cycle spanning 180 days and harnessing thousands of GPUs. This exponential increase in computational demands as the number of agents grows (and the corresponding joint action and state spaces)  poses a significant challenge during MARL deployment.

To tackle these computational challenges, researchers have delved into dynamic sparse training (DST), a method that trains neural network models with dynamically sparse topology. For example, RigL \cite{evci2020rigging} can train a 90\%-sparse network from scratch in deep supervised learning without performance degradation. However, in deep reinforcement learning (DRL), the learning target evolves in a bootstrapping manner \cite{tesauro1995temporal}, and the distribution of training data is path-dependent \cite{desai2019evaluating}, posing additional challenges to sparse training. Improper sparsification can result in irreversible damage to the learning path \cite{igl2020transient}. Initial attempts at DST in sparse single-agent DRL training have faced difficulties in achieving consistent model compression across diverse environments, as documented in \cite{sokar2021dynamic, graesser2022state}. This is mainly because sparse models may introduce significant bias, leading to unreliable learning targets and exacerbating training instability as agents learn through bootstrapping. Moreover, the partially observable nature of each agent makes training non-stationarity inherently more severe in multi-agent settings. Collectively, these factors pose significant challenges for value learning in each agent under sparse models.

\begin{wrapfigure}{r}{.3\linewidth}
\centering
\includegraphics[width=\linewidth]{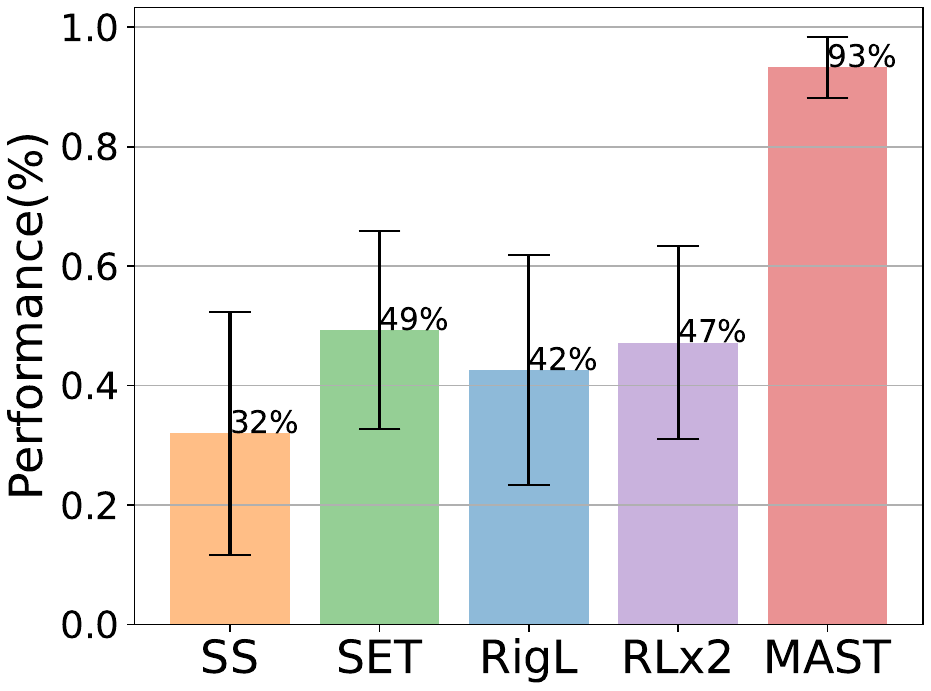}
\caption{Comparison of different sparse training methods.}
\label{fig:intro}
\end{wrapfigure}
We present a motivating experiment in Figure~\ref{fig:intro}, where we evaluated various sparse training methods on the {\ttfamily 3s5z} task from SMAC \cite{samvelyan19smac} using a neural network with only $10\%$ of its original parameters. Classical DST methods, including SET \cite{mocanu2018scalable} and RigL \cite{evci2020rigging}, demonstrate poor performance in MARL scenarios, along with using static sparse networks (SS). Additionally, RLx2 as proposed in \cite{tan2022rlx2} proves ineffective for multi-agent settings, despite enabling DST for single-agent settings with $90\%$ sparsity. In contrast, our MAST framework in this work achieves a win rate of over $90\%$. In addition, the only prior attempt to train sparse MARL agents, as described in \cite{yang2022learninggroup}, prunes agent networks during training with weight grouping \cite{wang2019fully}. However, this approach fails to maintain sparsity throughout training, and only achieves a final model sparsity of only $80\%$. Moreover, their experiment is confined to a two-user environment, PredatorPrey-v2, in MuJoCo \cite{todorov2012mujoco}. These observations highlight the fact that, despite its potential, the application of sparse networks in the context of MARL remains largely unexplored (A comprehensive literature review is deferred in Appendix~\ref{app:rw}). Consequently, a critical and intriguing question arises:
\begin{center}
\textbf{\emph{Can we train MARL agents effectively using ultra-sparse networks throughout?}}
\end{center}
We affirmatively address the question by introducing a novel sparse training framework, Multi-Agent Sparse Training (MAST). Since improper sparsification results in network fitting errors in the learning targets and incurs large policy inconsistency errors in the training samples, MAST ingeniously integrates the Soft Mellowmax Operator with a hybrid TD-($\lambda$) schema to establish reliable learning targets. Additionally, it incorporates a novel dual replay buffer mechanism to enhance the distribution of training samples. Leveraging these components, MAST employs gradient-based topology evolution to exclusively train multiple MARL agents using sparse networks. Consequently, MAST facilitates the training of highly efficient MARL agents with minimal performance compromise, employing ultra-sparse networks throughout the training process.

Our extensive experimental investigation across various value-based MARL algorithms on multiple  SMAC benchmarks reveals MAST's ability to achieve model compression ranging from $5\times$ to $20\times$, while incurring minimal performance trade-offs (under $3\%$). Moreover, MAST demonstrates an impressive capability to reduce the Floating Point Operations (FLOPs) required for both training and inference by up to $20\times$, showcasing a significant margin over other baselines (detailed in Section~\ref{sec:exp_2}).

\section{Preliminaries}
\paragraph{Deep MARL} We model the MARL problem as a decentralized partially observable Markov decision process (Dec-POMDP) \cite{oliehoek2016concise}, represented by a tuple $\langle \mathcal{N}, \mathcal{S}, \mathcal{U}, P, r, \mathcal{Z}, O, \gamma\rangle$. Deep Multi-Agent $Q$-learning extends the deep $Q$ learning method \cite{mnih2013playing} to multi-agent scenarios \cite{sunehag2018value, rashid2020monotonic, son2019qtran}. The agent-wise Q function is defined over its history $\tau_i$ as $Q_i$ for agent $i$. Subsequently, the joint action-value function $Q_\text{tot}(\boldsymbol{\tau}, \boldsymbol{u})$ operates over the joint action-observation history $\boldsymbol{\tau}$ and joint action $\boldsymbol{u}$. The objective, given transitions $(\boldsymbol{\tau}, \boldsymbol{u}, r, \boldsymbol{\tau}')$ sampled from the experience replay buffer $\mathcal{B}$, is to minimize the mean squared error loss $\mathcal{L}(\theta)$ on the temporal-difference (TD) error $\delta = y - Q_{\text{tot}}(\boldsymbol{\tau}, \boldsymbol{u})$. Here, the TD target $y = r + \gamma \max_{\boldsymbol{u}'} \bar{Q}_{\text{tot}}(\boldsymbol{\tau}', \boldsymbol{u}')$, where $\bar{Q}_{\text{tot}}$ is the target network for the joint action $Q$-function, periodically copied from $Q_{\text{tot}}$. Parameters of $Q_{\text{tot}}$ are updated using $\theta' = \theta - \alpha \nabla_\theta \mathcal{L}(\theta)$, with $\alpha$ representing the learning rate.

We focus on algorithms that adhere to the Centralized Training with Decentralized Execution (CTDE) paradigm \cite{kraemer2016multi}, within which, agents undergo centralized training, where the complete action-observation history and global state are available. However, during execution, they are constrained to individual local action-observation histories. To efficiently implement CTDE, the Individual-Global-Maximum (IGM) property \cite{son2019qtran} in Eq.~(\ref{eq:igm}), serves as a key mechanism: 
\begin{equation}\label{eq:igm}
\arg \max _{\boldsymbol{u}} Q_{\text{tot}}(s, \boldsymbol{u})=\big(\arg \max _{u_1} Q_1\left(s, u_1\right), \cdots, \arg \max _{u_N} Q_N\left(s, u_N\right)\big).
\end{equation}
Many deep MARL algorithms adhere to the IGM criterion, such as the QMIX series algorithms \cite{rashid2020monotonic, rashid2020weighted,pan2021regularized}. These algorithms employ a mixing network $f_s$ with non-negative weights, enabling the joint Q-function to be expressed as $Q_{\text{tot}}(s, \boldsymbol{u})=f_s\left(Q_1\left(s, u_1\right), \cdots, Q_N\left(s, u_N\right)\right)$.

\begin{figure}[H]
\centering
\includegraphics[width=.9\linewidth]{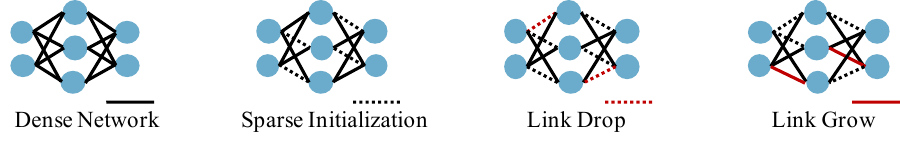}
\caption{Illustration of dynamic sparse training.}
\label{fig:rigl}
\end{figure}

\paragraph{Dynamic Sparse Training} Dynamic sparse training (DST), initially proposed in deep supervised learning, can train a 90\% sparse network without performance degradation from scratch, such as in ResNet-50 \cite{he2016deep} and MobileNet \cite{howard2017mobilenets}. In DST, the dense network is randomly sparsified at initialization, as shown in Figure~\ref{fig:rigl}, and its topology is dynamically changed during training by link dropping and growing. Specifically, the topology evolution mechanism in MAST follows the RigL method \cite{evci2020rigging}, which improves the optimization of sparse neural networks by leveraging weight magnitude and gradient information to jointly optimize model parameters and connectivity.

RigL periodically and dynamically drops a subset of existing connections with the smallest absolute weight values and concurrently grows an equivalent number of empty connections with the largest gradients. The pseudo-code of RigL is given in Algorithm~\ref{alg:updatemask}, where the symbol $\odot$ denotes the element-wise multiplication operator, $M_{\theta}$ symbolizes the binary mask that delineates the sparse topology for the network $\theta$, and $\zeta_t$ is the update fraction in training step $t$. This process maintains the network sparsity throughout the training with a strong evolutionary ability that saves training FLOPs and gives a sparse model after training.
\begin{algorithm}[H]
\caption{Topology Evoltion\cite{evci2020rigging}}\label{alg:updatemask}
\begin{algorithmic}[1]
\STATE $\theta_l,N_l,s_l$: parameters, number of parameters, sparsity of layer $l$.
\FOR{ each layer $l$}
\STATE$k=\zeta_t (1-s_l)N_l$
\STATE$\mathbb{I}_{\text{drop}}=\text{ArgTopK}(-|\theta_l\odot M_{\theta_l}|,k)$ \label{line:alg1-drop}
\STATE$\mathbb{I}_{\text{grow}}=\text{ArgTopK}_{i\notin \theta_l\odot M_{\theta_l} \backslash \mathbb{I}_{\text{drop}}}(|\nabla_{\theta_l}L|,k)$ \label{line:alg1-grow}
\STATE Update $M_{\theta_l}$ according to $\mathbb{I}_{\text{drop}}$ and $\mathbb{I}_{\text{grow}}$
\STATE$\theta_l\leftarrow\theta_l\odot M_{\theta_l}$
\ENDFOR
\end{algorithmic} 
\end{algorithm}
However, in DRL, the learning target evolves in a bootstrapping manner \cite{tesauro1995temporal}, such that the distribution of training data is path-dependent \cite{desai2019evaluating}, posing additional challenges to sparse training. Moreover, the partially observable nature of each agent exacerbates training non-stationarity, particularly in multi-agent settings. These factors collectively present significant hurdles for value learning in each agent under sparse models. As illustrated in Figure~\ref{fig:intro}, attempts to train ultra-sparse MARL models using simplistic topology evolution or the sparse training framework for single-agent RL have failed to achieve satisfactory performance. Therefore, MAST introduces innovative solutions to enhance value learning in ultra-sparse models by simultaneously improving the reliability of learning targets and the rationality of sample distribution.

\section{Enhancing Value Learning in Sparse Models}\label{sec:mast}
\begin{figure}[H]
\centering
\includegraphics[width=\linewidth]{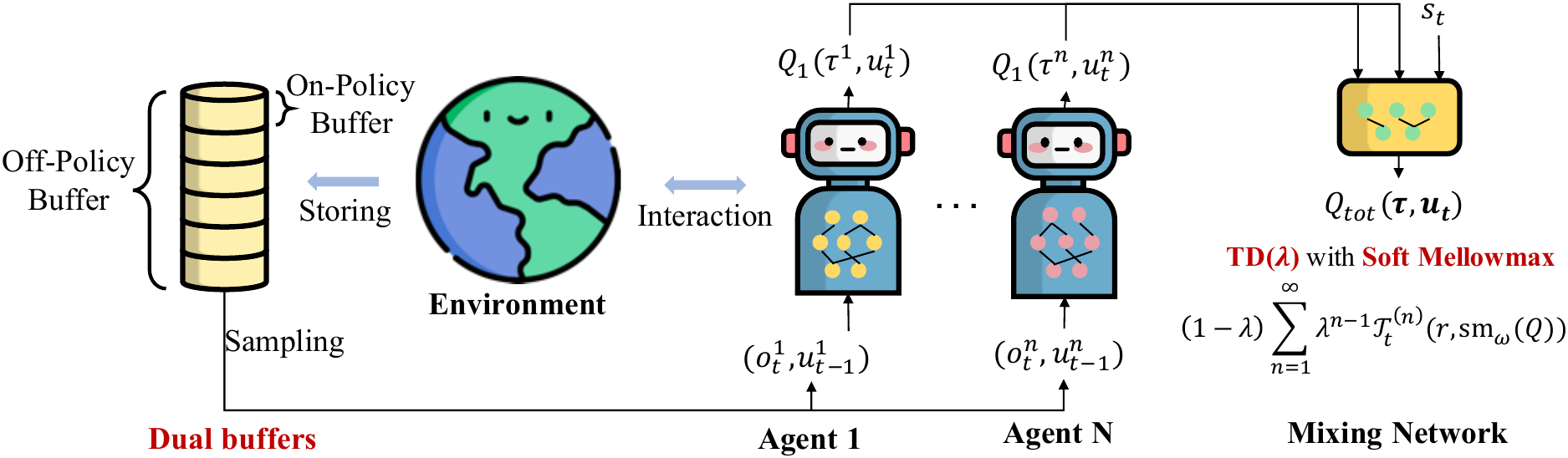}
\caption{An example of the MAST framework based on QMIX.}
\label{fig:framework}
\end{figure}

This section outlines the pivotal components of the MAST framework for training sparse MARL agents. MAST introduces innovative solutions to enhance the accuracy of value learning in ultra-sparse models by concurrently refining training data targets and distributions. Consequently, the topology evolution in MAST effectively identifies appropriate ultra-sparse network topologies. This approach aligns with single-agent DRL, where sparse training necessitates co-design with the value learning method as described in \cite{tan2022rlx2}. However, the partially observable nature of each agent exacerbates training non-stationarity in multi-agent settings. As illustrated in Figure~\ref{fig:framework}, MAST implements two key innovations to achieve accurate value learning in ultra-sparse models:
i) hybrid TD($\lambda$) targets combined with the Soft Mellowmax operator to mitigate estimation errors arising from network sparsity, and 
ii) dual replay buffers to reduce policy inconsistency errors due to sparsification.

\subsection{Improving the Reliability of Training Targets}
Initially, we observe that the expected error is amplified under sparse models, motivating the introduction of multi-step TD targets. However, different environments may require varying values of step lengths to achieve optimal performance. Consequently, we focus on \textbf{hybrid TD($\lambda$) targets}, which can achieve performance comparable to the best step length across different settings. Additionally, we find that the overestimation problem remains significant in sparse models. To address this, we propose the use of the \textbf{Soft Mellowmax operator} in constructing learning targets. This operator is effective in reducing overestimation bias without incurring additional computational costs.

\paragraph{Hybrid TD($\lambda$) Targets.} In deep multi-agent Q-learning, temporal difference (TD) learning is a fundamental method for finding an optimal policy, where the joint action-value network is iteratively updated by minimizing a squared loss driven by the TD target. Let $M_{\theta}$ be a binary mask representing the network's sparse topology, and denote the sparse network as $\hat{\theta} = \theta \odot M_{\theta}$, where $\odot$ signifies element-wise multiplication. Since sparse networks operate within a reduced hypothesis space with fewer parameters, the sparse network $\hat{\theta}$ may induce a large bias, making the learning targets unreliable, as evidenced in \cite{sokar2021dynamic}. 
 
\begin{wrapfigure}{r}{.5\linewidth}
\centering
\includegraphics[width=\linewidth]{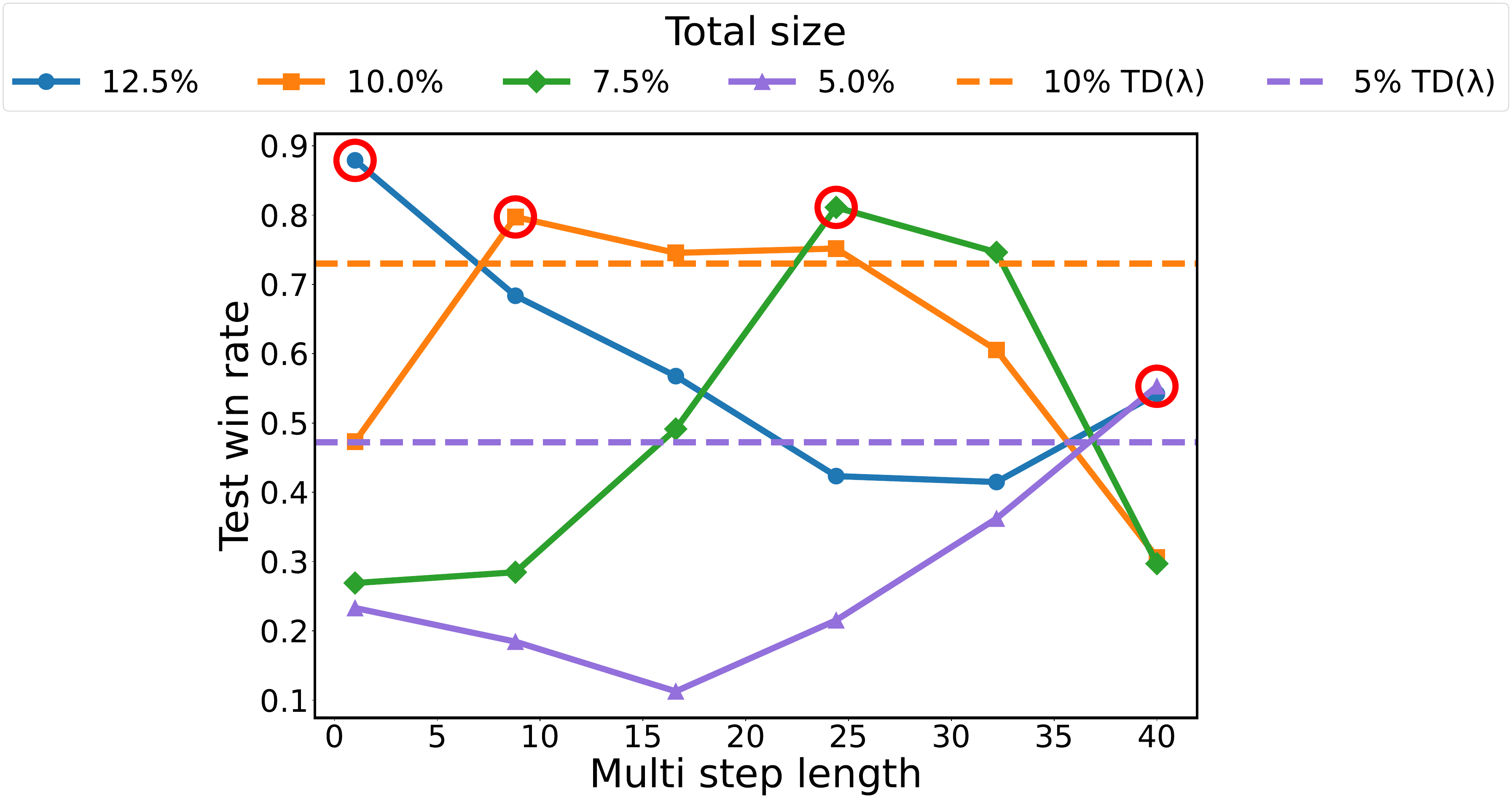}
\caption{Performances of different step lengths.}
\vspace{-.2cm}
\label{fig:nstep_reward}
\end{wrapfigure}

Moreover, we establish Theorem~\ref{th:tderror} to characterize the upper bound of the expected multi-step TD error under sparse models, where the multi-step return at $(s_t, \boldsymbol{u}_t)$ is $\mathcal{T}_n(s_t, \boldsymbol{u}_t) = \sum_{k=0}^{n-1} \gamma^k r_{t+k} + \gamma^n \max_{\boldsymbol{u}} Q_\text{tot}(s_{t+n}, \boldsymbol{u}; \hat{\theta})$ under sparse models.
As Eq~(\ref{eq:tderror}) shows, the expected multi-step TD error comes from two parts, intrinsical policy inconsistency error and network fitting error.
Thus, Eq~(\ref{eq:tderror}) implies that the expected TD error will be enlarged if the network is sparsified improperly with a larger network fitting error.
Indeed, the upper bound of the expected TD error will be enlarged if the network fitting error is increased. Subsequently, it is infeasible for the model to learn a good policy. Eq.~(\ref{eq:tderror}) also shows that introducing a multi-step return target discounts the network fitting error by a factor of $\gamma^n$ in the upper bound of the expected TD error. Thus, employing a multi-step return $\mathcal{T}_{t}^{(n)}$ with a sufficiently large $n$, or even Monte Carlo methods \cite{sutton2018reinforcement}, can effectively diminish the TD error caused by network sparsification for $\gamma < 1$. 
\begin{theorem}\label{th:tderror}
Denote $\pi$ as the target policy at timestep $t$, and $\rho$ as the behavior policy generating the transitions $(s_t,\boldsymbol{u}_t, \ldots, s_{t+n}, \boldsymbol{u}_{t+n})$. Denote the network fitting error as $\epsilon(s,\boldsymbol{u}) = |Q_\text{tot}(s,\boldsymbol{u};\hat{\theta}) - Q_\text{tot}^{\pi}(s,\boldsymbol{u})|$. Then, the expected error between the multi-step TD target $\mathcal{T}_n$ conditioned on transitions from the behavior policy $\rho$ and the true joint action-value function $Q_\text{tot}^\pi$ is
\begin{equation}\label{eq:tderror}
\begin{aligned}
&|\mathbb{E}_{\rho}[\mathcal{T}_n(s_t, \boldsymbol{u}_t)] - Q_\text{tot}^{\pi}(s_t, \boldsymbol{u}_t)| \le \gamma^n \mathbb{E}_{\rho}[\underbrace{2\epsilon(s_{t+n}, \rho(s_{t+n})) + \epsilon(s_{t+n}, \pi(s_{t+n}))}_{\text{Network fitting error}}] \\
&+ \underbrace{|Q_\text{tot}^{\rho}(s_{t}, \boldsymbol{u}_t) - Q_\text{tot}^{\pi}(s_t, \boldsymbol{u}_t)|}_{\text{Policy inconsistency error}} + \underbrace{\gamma^n \mathbb{E}_{\rho}[|Q_\text{tot}^{\pi}(s_{t+n}, \pi(s_{t+n})) - Q_\text{tot}^{\rho}(s_{t+n}, \rho(s_{t+n}))|]}_{\text{Discounted policy inconsistency error}}.
\end{aligned}
\end{equation}
\begin{proof}
Please refer to Appendix~\ref{app:pf}.
\end{proof}
\end{theorem}
However, the Monte Carlo method is prone to high variance, suggesting that an optimal TD target in sparse models should be a multi-step return with a judiciously chosen step length, balancing network fitting error due to sparsification and training variance. Figure~\ref{fig:nstep_reward} illustrates model performance across different step lengths and model sizes, revealing that an optimal step length exists for various model sizes. Moreover, the optimal step length increases as model size decreases, which aligns with Theorem~\ref{th:tderror} due to the increased network fitting error in models with higher sparsity.
The above facts suggest the need to increase the step length in the learning targets to maintain the performance of sparse models. However, the optimal step length varies across different settings. To address this, we introduce the TD($\lambda$) target \cite{sutton2018reinforcement} to achieve a good trade-off: $\mathcal{T}_\lambda = (1-\lambda) \sum_{n=1}^{\infty} \lambda^{n-1} \mathcal{T}_n$ for $\lambda\in[0,1]$. This target averages all possible multi-step returns $\{\mathcal{T}_n\}_{n=1}^\infty$ into a single return using exponentially decaying weights, providing a computationally efficient approach with episode-form data. In Figure~\ref{fig:nstep_reward}, we also plot two representative performances of TD($\lambda$) under $10\%$ and $5\%$ model sizes, which are both close to the optimal step length under different model sizes.

Previous studies \cite{fedus2020revisiting} have highlighted that an immediate shift to multi-step targets can exacerbate policy inconsistency error as shown in Eq.~(\ref{eq:tderror}). Since the TD($\lambda$) target $\mathcal{T}_\lambda$ averages all potential multi-step returns $\{\mathcal{T}_n\}_{n=1}^\infty$, an immediate transition to this target may encounter similar issues. To address this challenge, we adopt a hybrid strategy inspired by the delayed mechanism proposed in \cite{tan2022rlx2}. Initially, when the training step is less than a threshold $T_0$, we employ one-step TD targets ($\mathcal{T}_1$) to minimize policy inconsistency errors. As training progresses and the policy stabilizes, we transit to TD($\lambda$) targets to mitigate sparse network fitting errors. This mechanism ensures consistent and reliable learning targets throughout the sparse training process.

\paragraph{Soft Mellowmax Operator.} The max operator in the Bellman operator poses a well-known theoretical challenge, namely overestimation, which hinders the convergence of various linear and non-linear approximation schemes \cite{tsitsiklis1996analysis}. Deep MARL algorithms, including QMIX \cite{rashid2020monotonic}, also grapple with the overestimation problem. Several works have addressed this issue in dense MARL algorithms, such as double critics \cite{ackermann2019reducing}, weighted critic updates \cite{sarkar2021weighted}, the Softmax operator \cite{pan2021regularized}, and the Sub-Avg operator \cite{wu2022sub}. However, these methods introduce additional computational costs, sometimes even doubling the computational budget, which is infeasible for our sparse training framework.

We turn our attention to the Soft Mellowmax operator, which has been proven effective in reducing overestimation for dense MARL algorithms in \cite{gan2021stabilizing}. For MARL algorithms satisfying the IGM property in Eq.~(\ref{eq:igm}), we replace the max operator in $Q_i$ with the Soft Mellowmax operator in Eq.~(\ref{eq:sm}) to mitigate overestimation bias in the joint-action Q function within sparse models:
\begin{equation}\label{eq:sm}
    \operatorname{sm}_\omega (Q_i(\tau, \cdot))=\frac{1}{\omega}\log \left[\sum_{u\in\mathcal{U}} \frac{\exp \left(\alpha Q_i\left(\tau, u\right)\right)}{ \sum_{u'\in\mathcal{U}}\exp \left(\alpha Q_i\left(\tau, u'\right)\right)}\exp \left(\omega Q_i\left(\tau, u\right)\right)\right],
\end{equation}
where $\omega>0$, and $\alpha \in \mathbb{R}$. Let $\mathcal{T}$ be the value estimation operator that estimates the value of the next state $s'$. Theorem~\ref{th:sm2} shows that the Soft Mellowmax operator can reduce the severe overestimation bias in sparse models.
\begin{theorem}\label{th:sm2}
Let $B(\mathcal{T})=\mathbb{E}\left[\mathcal{T}\left(s^{\prime}\right)\right]-\max _{\boldsymbol{u}^{\prime}} Q_{\text {tot }}^*\left(s^{\prime}, \boldsymbol{u}^{\prime}\right)$ be the bias of value estimates of $\mathcal{T}$. For an arbitrary joint-action $Q$-function $\bar{Q}_{\text {tot}}$, if there exists some $V_{\text {tot }}^*\left(s^{\prime}\right)$ such that $V_{\text {tot }}^*\left(s^{\prime}\right)=Q_{\text {tot }}^*\left(s^{\prime}, \boldsymbol{u}^{\prime}\right)$ for different joint actions, $\sum_{\boldsymbol{u}^{\prime}}\left(\bar{Q}_{\text {tot}}\left(s^{\prime}, \boldsymbol{u}^{\prime}\right)-V_{\text {tot }}^*\left(s^{\prime}\right)\right)=0$, and $\frac{1}{|\boldsymbol{U}|} \sum_{\boldsymbol{u}^{\prime}}\left(\bar{Q}_{\text {tot}}\left(s^{\prime}, \boldsymbol{u}^{\prime}\right)-V_{\text {tot }}^*\left(s^{\prime}\right)\right)^2=C$ $(C>0)$, then $B\left(\mathcal{T}_\text {RigL-QMIX-SM}\right) \le B\left(\mathcal{T}_\text{RigL-QMIX}\right)$.

\begin{proof}
    Please refer to Appendix~\ref{app:sm2}.
\end{proof}
\end{theorem}

\begin{wrapfigure}{r}{.6\linewidth}
\centering
\subfigure[Win rates\label{subfig:trick_sm2_rate}]{\includegraphics[width=.45\linewidth]{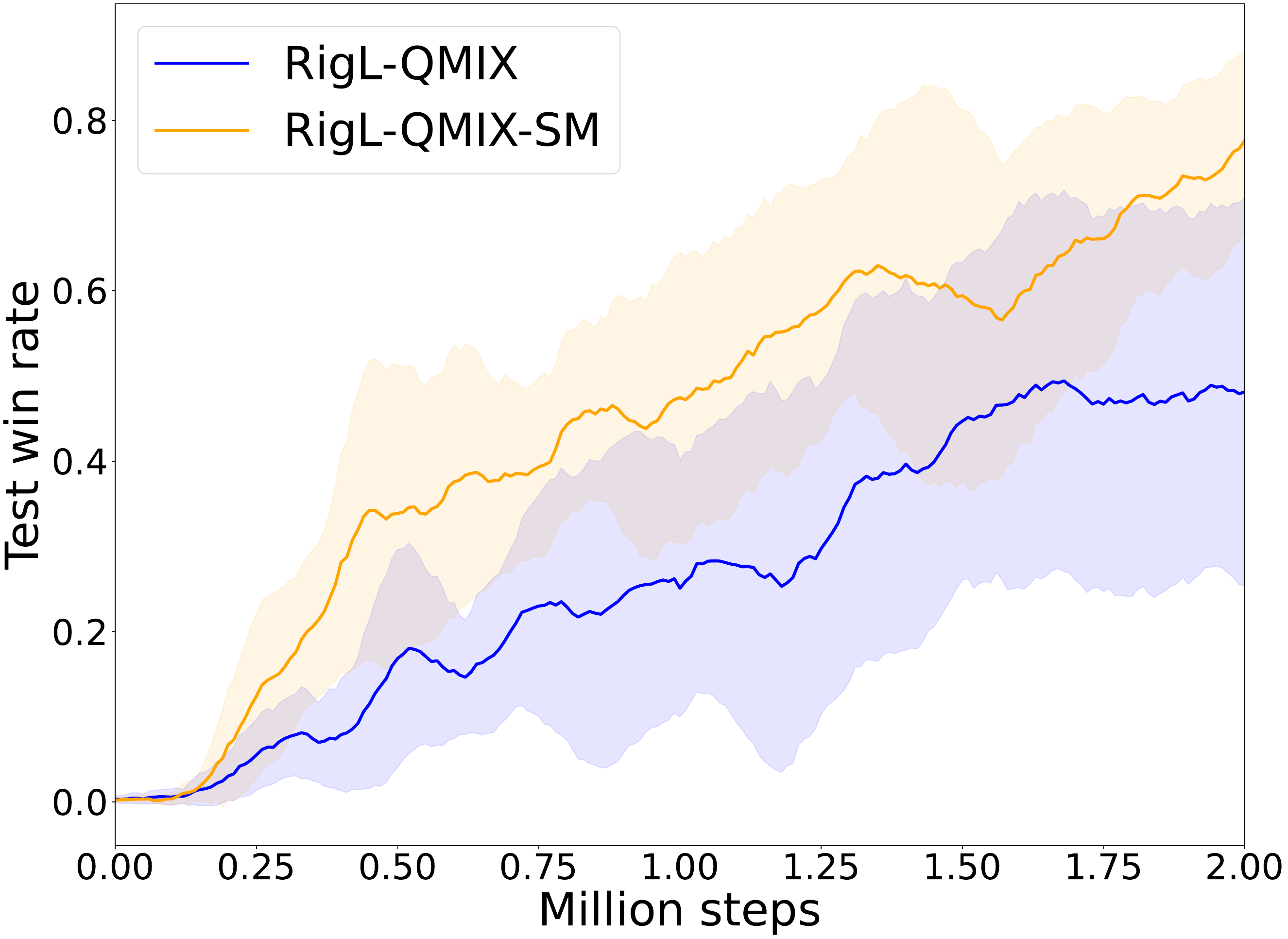}}
\subfigure[Estimated values\label{subfig:trick_sm2_q}]{\includegraphics[width=.43\linewidth]{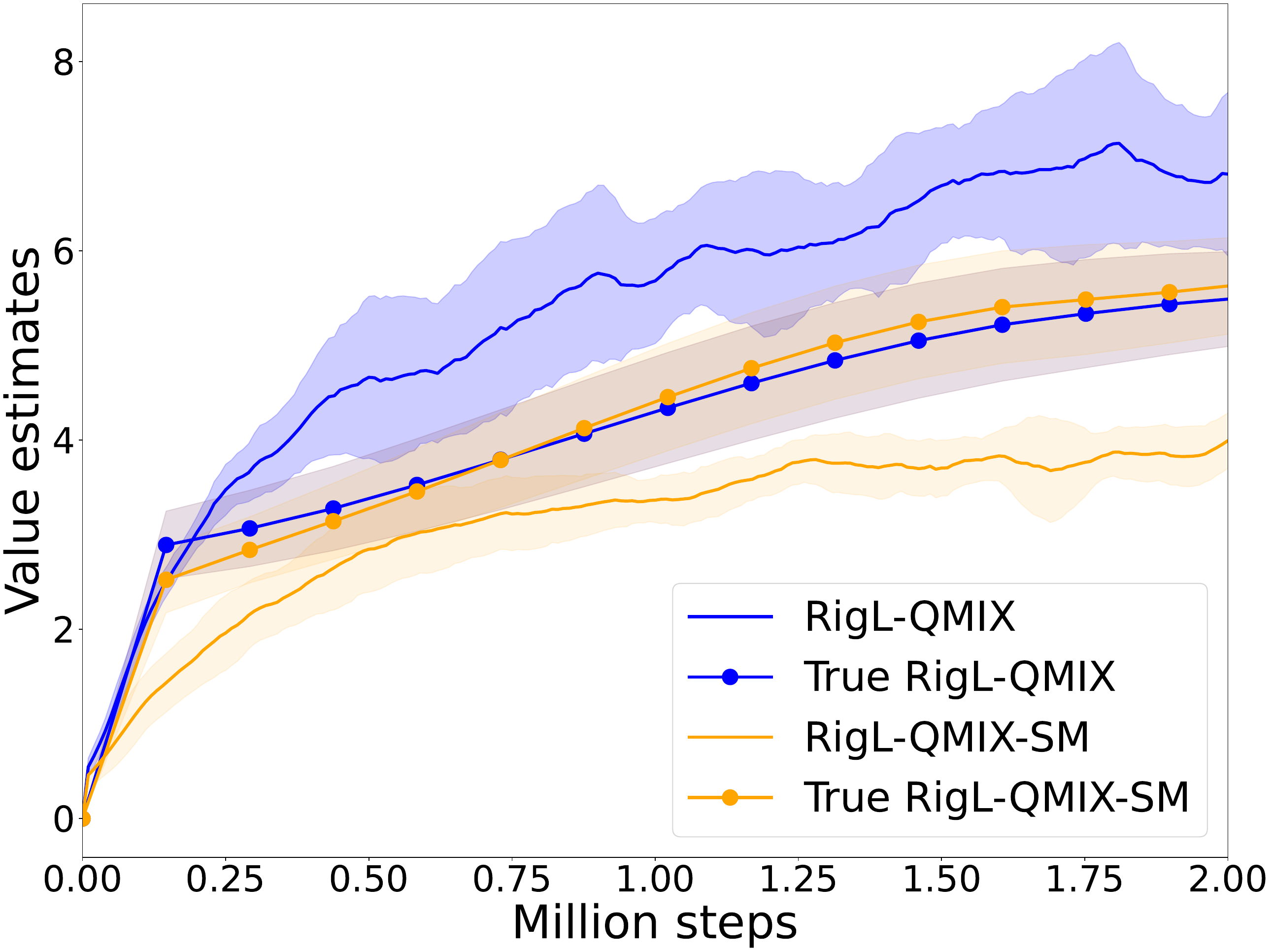}}
\caption{Effects of Soft Mellowmax operator.}
\label{fig:overestimation}
\end{wrapfigure}
Our empirical investigations reveal that overestimation remains a significant performance barrier in sparse models, resulting in substantial performance degradation. Figure~\ref{fig:overestimation} illustrates the win rates and estimated values of QMIX with and without our Soft Mellowmax operator on {\ttfamily 3s5z} in the SMAC. Figure~\ref{subfig:trick_sm2_rate} shows that the performance of RigL-QMIX-SM outperforms RigL-QMIX, while Figure~\ref{subfig:trick_sm2_q} demonstrates that the Soft Mellowmax operator effectively mitigates overestimation bias.
These findings highlight that QMIX still faces overestimation issues in sparse models and underscore the efficacy of the Soft Mellowmax operator in addressing this problem. Additionally, the Soft Mellowmax operator introduces negligible extra computational costs, as it averages the Q function over each agent's individual action spaces rather than the joint action spaces used in the Softmax operator \cite{pan2021regularized}, which grow exponentially with the number of agents.

\subsection{Improving the Rationality of Sample Distribution}
Although we have improved the reliability of the learning target's confidence as discussed above, the learning process can still suffer from instability due to improper sparsification. This suggests the need to enhance the distribution of training samples to stabilize the training process. Improper sparsification can lead to irreversible damage to the learning path \cite{igl2020transient} and exacerbate training instability as agents learn through bootstrapping. 

Generally, sparse models are more challenging to train compared to dense models due to the reduced hypothesis space \cite{evci2019difficulty}. Therefore, it is important to prioritize more recent training samples to estimate the true value function within the same training budget. 
Failing to do so can result in excessively large policy inconsistency errors, thereby damaging the learning process irreversibly. MAST introduces a dual buffer mechanism utilizing two First-in-First-Out (FIFO) replay buffers: $\mathcal{B}_1$ (with large capacity) and $\mathcal{B}_2$ (with small capacity), with some data overlap between them. While $\mathcal{B}_1$ adopts an off-policy style, $\mathcal{B}_2$ follows an on-policy approach. During each training step, MAST samples $b_1$ episodes from $\mathcal{B}_1$ and $b_2$ episodes from $\mathcal{B}_2$, conducting a gradient update using a combined batch size of $(b_1 + b_2)$. For instance, in our experiments, we set $|\mathcal{B}_1|:|\mathcal{B}_2| = 50:1$ and $b_1:b_2 = 3:1$. Generally, training with online data enhances learning stability, as the behavior policy closely matches the target policy. Conversely, training with offline data improves sample efficiency but can lead to instability.

\begin{wrapfigure}{r}{.3\linewidth}
\centering
\includegraphics[width=1\linewidth]{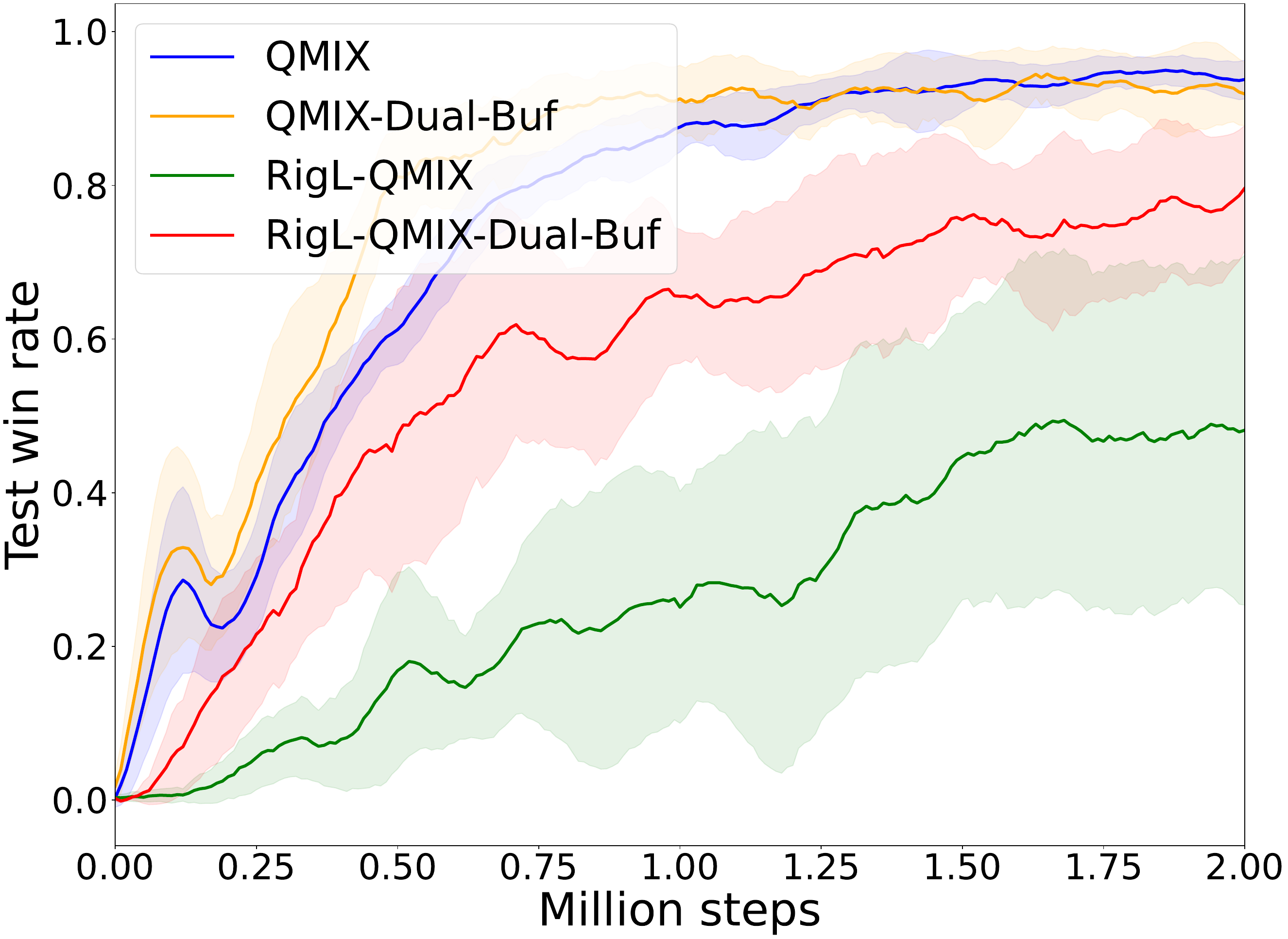}
\caption{Different buffers.}
\label{fig:buffer}
\end{wrapfigure}
Figure~\ref{fig:buffer} illustrates the training dynamics for RigL-QMIX in the SMAC's {\ttfamily 3s5z} task. The green curve with large variances highlights the training instability of QMIX in sparse models. However, with the integration of dual buffers, QMIX's training stability and efficiency are significantly improved under sparse conditions, leading to consistent policy enhancements and higher rewards. Notably, the dual buffer mechanism does not enhance dense training, suggesting that this approach is particularly effective in sparse scenarios where network parameters are crucial for ensuring stable policy improvements. Although prior works \cite{schaul2015prioritized, hou2017novel, banerjee2022improved} have explored prioritized or dynamic-capacity buffers, their applicability in this context may be limited due to the data being in episode form in value-based deep MARL algorithms, making it difficult to determine the priority of each training episode. Similarly, the dynamic buffer approach in \cite{tan2022rlx2} is also inapplicable, as the policy distance measure cannot be established for episode-form data. This further emphasizes the unique effectiveness of the dual buffer approach in enhancing training stability for sparse MARL models.

\begin{wrapfigure}{r}{.35\linewidth}
\centering
\includegraphics[width=1\linewidth]{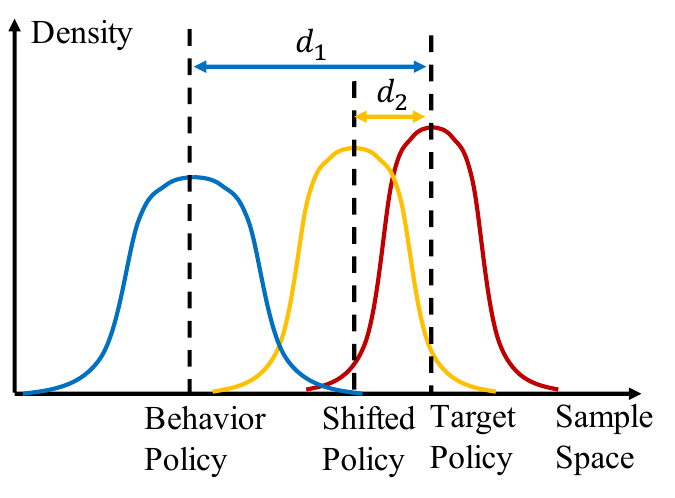}
\vspace{-.5cm}
\caption{Distribution Shift: $d_1$ and $d_2$ are distribution distances.}
\label{fig:db}
\vspace{-1cm}
\end{wrapfigure}

We highlight that the performance improvement observed with MAST is not due to providing more training data samples to agents. Instead, it results from a more balanced training data distribution enabled by the utilization of dual buffers. The primary objective of incorporating dual buffers is to shift the overall distribution of training data. Figure~\ref{fig:db} illustrates the distribution of samples induced by different policies, where behavior policy and target policy are defined in Theorem~\ref{th:tderror}.
Specifically, samples in the original single buffer are subject to the distribution of the behavior policy (blue), which introduces a policy inconsistency error $d_1$. However, by utilizing dual buffers, the distribution of training samples shifts towards the target policy, as there are more recent samples from the on-policy buffer. This reduces the policy inconsistency in our dual buffers, thereby bolstering the stability and effectiveness of the learning process under sparse models. Additionally, the extra on-policy buffer does not significantly reduce sample efficiency, as its capacity is small.

\section{Experiments}\label{sec:exp_2}
In this section, we conduct a comprehensive performance evaluation of MAST across various tasks in the StarCraft Multi-Agent Challenge (SMAC) \cite{samvelyan19smac} benchmark. Additional experiments on the multi-agent MuJoCo (MAMuJoCo) \cite{peng2021facmac} benchmark are provided in Appendix~\ref{app:comix}. MAST serves as a versatile sparse training framework specifically tailored for value decomposition-based MARL algorithms. In Section~\ref{subsec:ce_2}, we integrate MAST with state-of-the-art value-based deep MARL algorithms, including QMIX \cite{rashid2020monotonic}, WQMIX \cite{rashid2020weighted}, and RES \cite{pan2021regularized}. We also apply MAST to a hybrid value-based and policy-based algorithm, FACMAC \cite{peng2021facmac}. Subsequently, we assess the performance of sparse models generated by MAST in Section~\ref{subsec:mask}. Furthermore, a comprehensive ablation study of MAST components is detailed in Appendix~\ref{app:ablation}. Each reported result represents the average performance over eight independent runs, each utilizing distinct random seeds.

\subsection{Comparative Evaluation}\label{subsec:ce_2}
Table~\ref{tb:comp_eval_2} presents a comprehensive summary of our comparative evaluation in the SMAC benchmark, where MAST is benchmarked against the following baseline methods:
(i) Tiny: Utilizing tiny dense networks with a parameter count matching that of the sparse model during training.
(ii) SS: Employing static sparse networks with random initialization.
(iii) SET \cite{mocanu2018scalable}: Pruning connections based on their magnitude and randomly expanding connections.
(iv) RigL \cite{evci2020rigging}: This approach leverages dynamic sparse training, akin to MAST, by removing and adding connections based on magnitude and gradient criteria, respectively.
(v) RLx2 \cite{tan2022rlx2}: A specialized dynamic sparse training framework tailored for single-agent reinforcement learning.
\begin{table}[H]
    \caption{Comparisons of MAST with different sparse training baselines: "Sp." stands for "sparsity", "Total Size" means total model parameters, and the data is all normalized w.r.t. the dense model.}
    \label{tb:comp_eval_2}
    \centering
    \begin{tabular}{p{.6cm}|p{0.55cm}|p{0.5cm}p{0.75cm}p{1cm}p{1cm}|p{0.65cm}p{0.65cm}p{0.65cm}p{0.75cm}p{0.85cm}p{1cm}}
    \toprule
    Alg. & Env. & \makecell[c]{Sp.} & \makecell[c]{Total\\Size} & \makecell[c]{FLOPs\\(Train)} & \makecell[c]{FLOPs\\(Test)} & \makecell[c]{Tiny\\(\%)} & \makecell[c]{SS\\(\%)} & \makecell[c]{SET\\(\%)} & \makecell[c]{RigL\\(\%)} & \makecell[c]{RLx2\\(\%)} & \makecell[c]{MAST\\(\%)} \\
    \midrule
    \multirow{5}{*}{\makecell[c]{Q-\\MIX}} & 3m & 95\% & 0.066x & 0.051x & 0.050x & 98.3 & 91.6 & 96.0 & 95.3 & 12.1 & \textbf{100.9}\\
    & 2s3z & 95\% & 0.062x & 0.051x & 0.050x & 83.7 & 73.0 & 77.6 & 69.4 & 45.8 & \textbf{98.0} \\
    & 3s5z & 90\% & 0.109x & 0.101x & 0.100x & 68.2 & 34.0 & 52.3 & 45.2 & 50.1 & \textbf{99.0}\\
    & 64* & 90\% & 0.106x & 0.100x & 0.100x & 58.2 & 40.2 & 67.1 & 48.7 & 9.9 & \textbf{97.6}\\
    \cline{2-12}
    & Avg. & 92\% & 0.086x & 0.076x & 0.075x & 77.1 & 59.7 & 73.2 & 64.6 & 29.8 & \textbf{98.9}\\
    \midrule
    \multirow{5}{*}{\makecell[c]{WQ-\\MIX}} & 3m & 90\% & 0.108x & 0.100x & 0.100x & 98.3 &  96.9 &  97.8 & 97.8  & 98.0 & \textbf{98.6}\\
    & 2s3z & 90\% & 0.106x & 0.100x & 0.100x & 89.6 & 75.4 &  85.9 &  86.8 & 87.3 & \textbf{100.2}\\
    & 3s5z & 90\% & 0.105x & 0.100x & 0.100x & 70.7 &  62.5 &  56.0 &  50.4 & 60.7 & \textbf{96.1}\\
    & 64* & 90\% & 0.104x & 0.100x & 0.100x & 51.0 &  29.6 &  44.1 &  41.0 & 52.8 & \textbf{98.4}\\
    \cline{2-12}
    & Avg. & 90\% & 0.106x & 0.100x & 0.100x & 77.4 & 66.1 & 70.9 & 69.0 & 74.7 & \textbf{98.1}\\
    \midrule
    \multirow{5}{*}{RES} & 3m & 95\% & 0.066x & 0.055x & 0.050x & 97.8 &  95.6 &  97.3 &  91.1 & 97.9 & \textbf{99.8}\\
    & 2s3z & 90\% & 0.111x & 0.104x & 0.100x & 96.5 &  92.8 &  92.8 &  94.7 & 94.0 & \textbf{98.4}\\
    & 3s5z & 85\% & 0.158x & 0.154x & 0.150x & 95.1 &  89.0 &  90.3 &  92.8 & 86.2 & \textbf{99.4}\\
    & 64* & 85\% & 0.155x & 0.151x & 0.150x & 83.3 &  39.1 &  44.1 &  35.3 & 72.7 & \textbf{104.9}\\
    \cline{2-12}
    & Avg. & 89\% & 0.122x & 0.116x & 0.112x & 93.2 & 79.1 & 81.1 & 78.5 & 87.7 & \textbf{100.6}\\
    \bottomrule
    \end{tabular}
\end{table}
We set the same sparsity levels for both the joint Q function $Q_\text{tot}$, and each individual agent's Q function $Q_i$. For every algorithm and task, the sparsity level indicated in Table~\ref{tb:comp_eval_2} corresponds to the highest admissible sparsity threshold of MAST. Within this range, MAST's performance consistently remains within a $3\%$ margin compared to the dense counterpart, effectively representing the minimal sparse model size capable of achieving performance parity with the original dense model. All other baselines are evaluated under the same sparsity level as MAST. We assess the performance of each algorithm by computing the average win rate per episode over the final $20$ policy evaluations conducted during training, with policy evaluations taking place at $10000$-step intervals. Identical hyperparameters are employed across all scenarios.

\paragraph{Performance} Table~\ref{tb:comp_eval_2} unequivocally illustrates MAST's substantial performance superiority over all baseline methods in all four environments across the three algorithms. Notably, static sparse (SS) consistently exhibits the lowest performance on average, highlighting the difficulty of finding optimal sparse network topologies in the context of sparse MARL models. Dynamic sparse training methods, namely SET and RigL, slightly outperform SS, although their performance remains unsatisfactory. Sparse networks also, on average, underperform tiny dense networks. However, MAST significantly outpaces all other baselines, indicating the successful realization of accurate value estimation through our MAST method, which effectively guides gradient-based topology evolution. Notably, the single-agent method RLx2 consistently delivers subpar results in all experiments, potentially due to the sensitivity of the step length in the multi-step targets, and the failure of dynamic buffer for episode-form training samples.
\begin{wrapfigure}{r}{.35\linewidth}
\centering
\includegraphics[width=\linewidth]{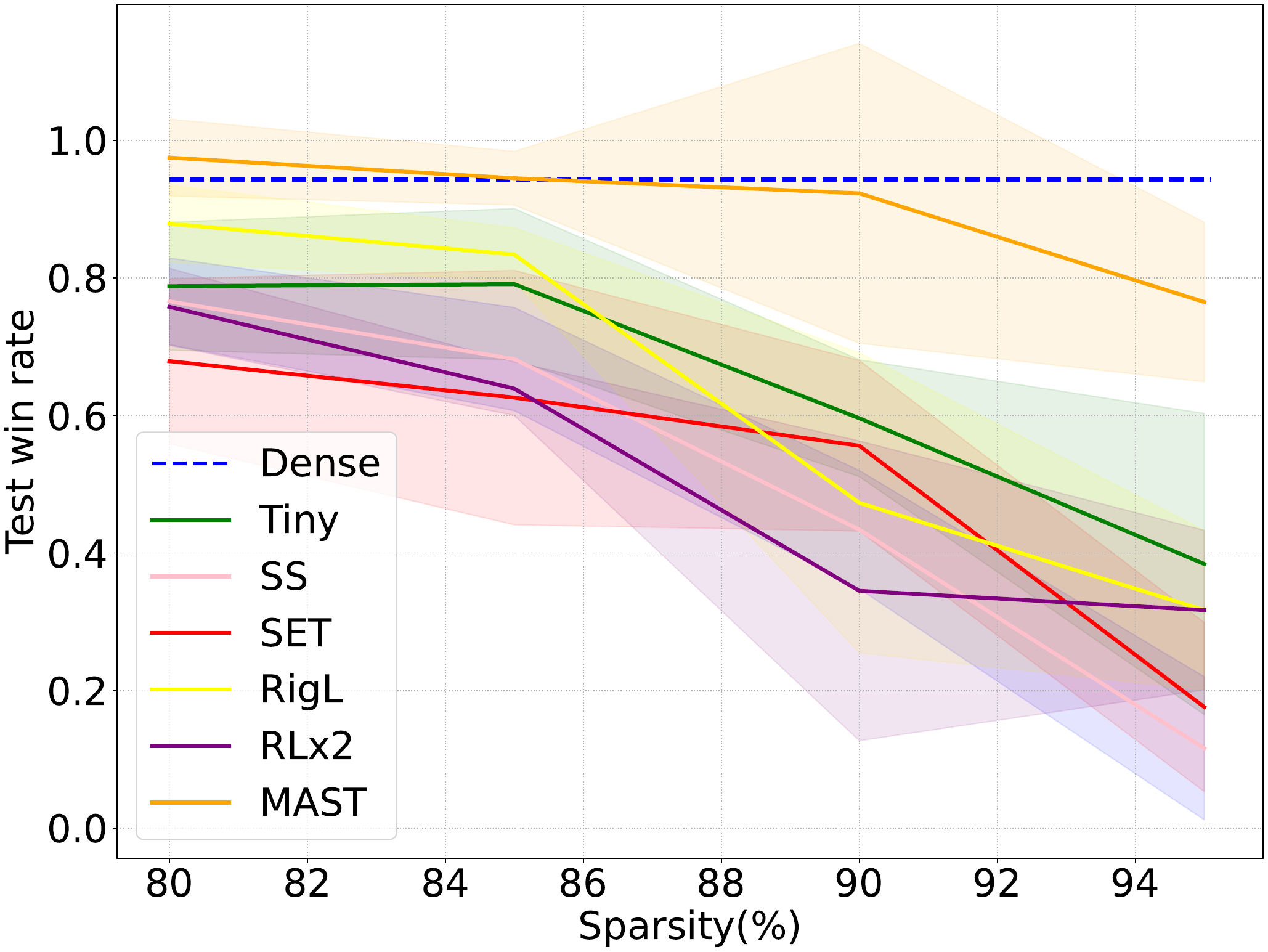}
\caption{Different sparsity.}
\label{fig:performance_2}
\vspace{-.7cm}
\end{wrapfigure}
To further substantiate the efficacy of MAST, we conduct performance comparisons across various sparsity levels in {\ttfamily 3s5z}, as depicted in Figure~\ref{fig:performance_2}. This reveals an intriguing observation: the performance of sparse models experiences a sharp decline beyond a critical sparsity threshold. Compared to conventional DST techniques, MAST significantly extends this critical sparsity threshold, enabling higher levels of sparsity while maintaining performance. Moreover, MAST achieves a higher critical sparsity threshold than the other two algorithms with existing baselines, e.g., SET and RigL, achieving a sparsity level of over $80\%$ on average. However, it is essential to note that the Softmax operator in RES averages the Q function over joint action spaces, which grow exponentially with the number of agents, resulting in significantly higher computational FLOPs and making it computationally incomparable to MAST. The detailed FLOPs calculation is deferred to Appendix~\ref{app:flops}.

\paragraph{FLOP Reduction and Model Compression} In contrast to knowledge distillation or behavior cloning methodologies, exemplified by works such as \cite{livne2020pops,vischer2021lottery}, MAST maintains a sparse network consistently throughout the entire training regimen. Consequently, MAST endows itself with a unique advantage, manifesting in a remarkable acceleration of training FLOPs. We observed  an acceleration  of up to $20\times$ in training and inference FLOPs for MAST-QMIX in the {\ttfamily 2s3z} task, with an average acceleration of $10\times$, $9\times$, and $8\times$ for QMIX, WQMIX, and RES-QMIX, respectively. Moreover, MAST showcases significant model compression ratios, achieving reductions in model size ranging from $5\times$ to $20\times$ for QMIX, WQMIX, and RES-QMIX, while incurring only minor performance degradation, all below $3\%$. 

\paragraph{Results on FACMAC} In addition to pure value-based deep MARL algorithms, we also evaluate MAST with a hybrid value-based and policy-based algorithm, FACMAC \cite{peng2021facmac}, in SMAC. The results are presented in Figure~\ref{fig:facmac}. From the figure, we observe that MAST consistently achieves a performance comparable with that of the dense models, and outperforms other methods in three environments, demonstrating its applicability across different algorithms.
\begin{figure}[H]
\centering
\subfigure[MMM]{\includegraphics[width=.3\linewidth]{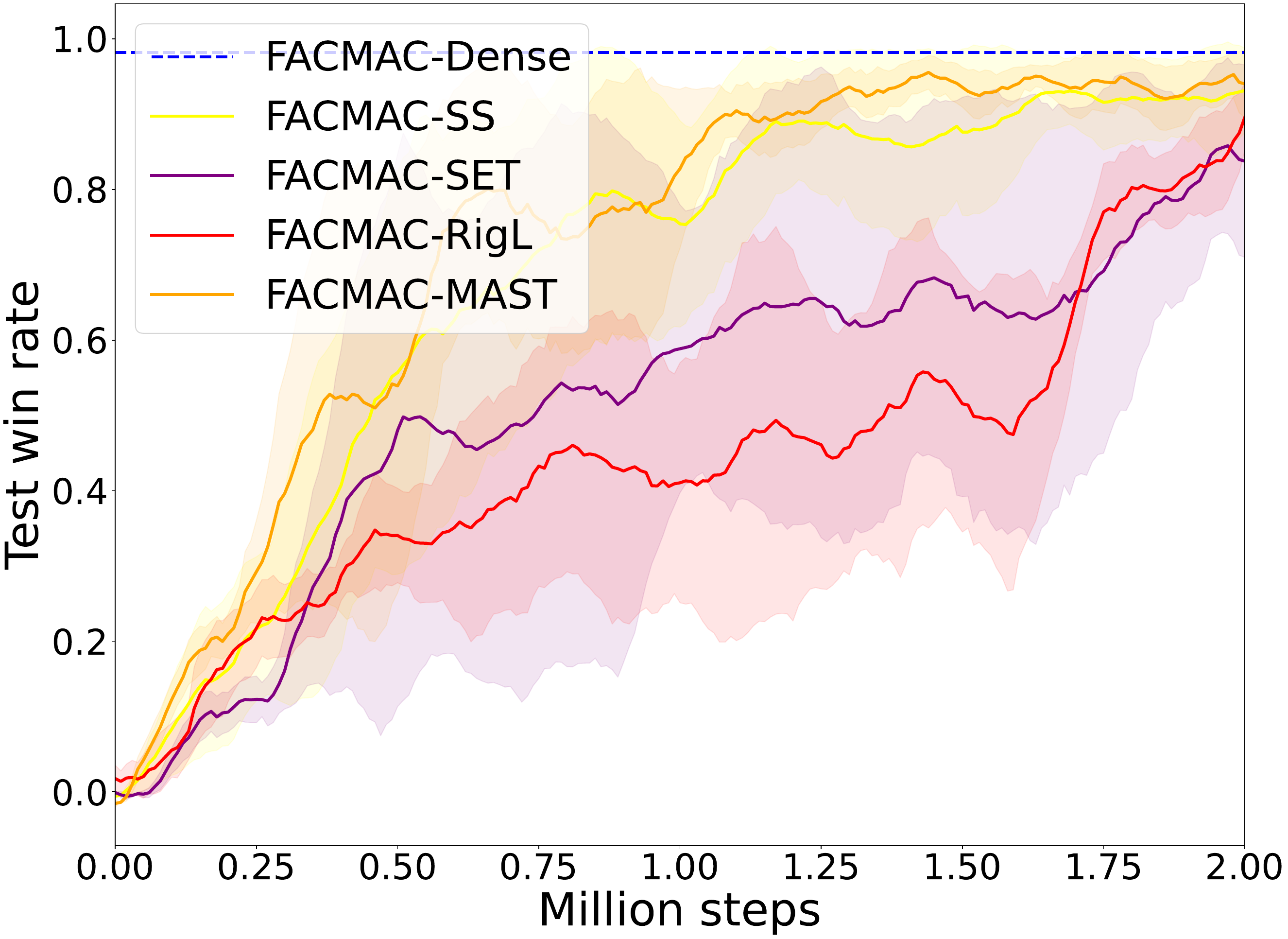}}
\subfigure[2c\_vs\_64zg]{\includegraphics[width=.3\linewidth]{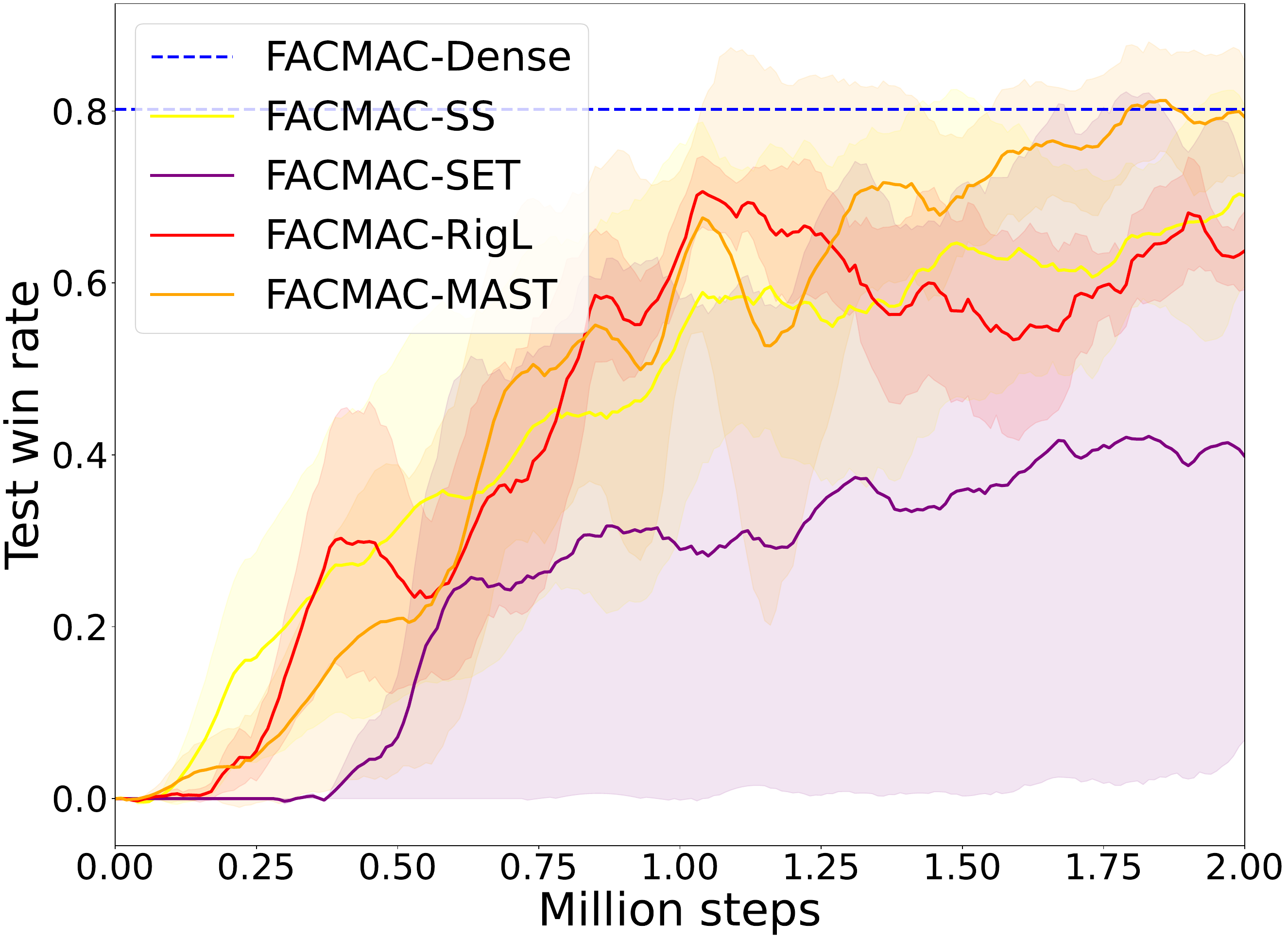}}
\subfigure[bane\_vs\_bane]{\includegraphics[width=.3\linewidth]{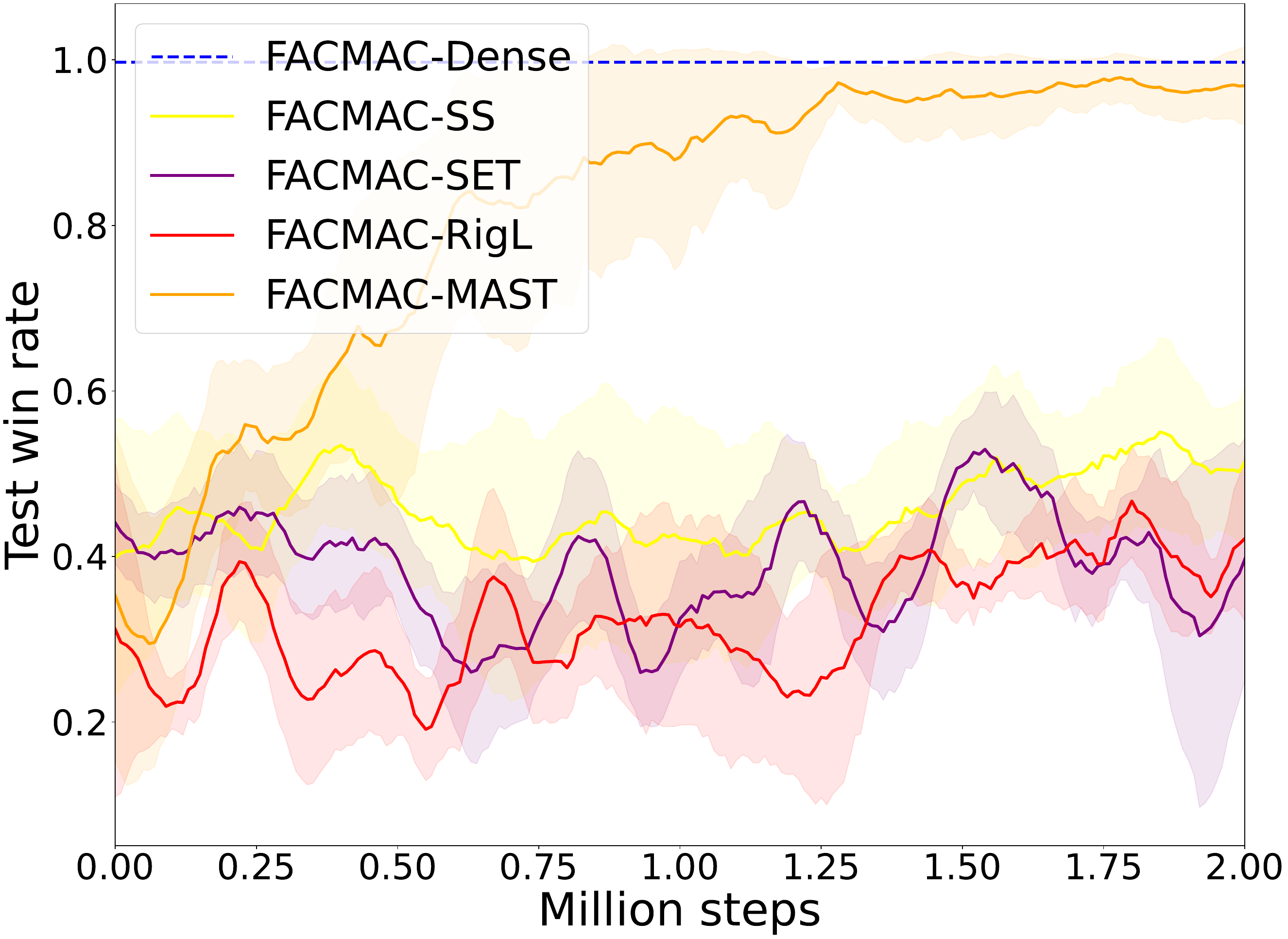}}
\caption{Mean episode win rates on different SMAC tasks with MAST-FACMAC.}
\label{fig:facmac}
\end{figure}

\subsection{Sparse Models Obtained by MAST}\label{subsec:mask}
\begin{wrapfigure}{r}{.3\linewidth}
\centering
\includegraphics[width=\linewidth]{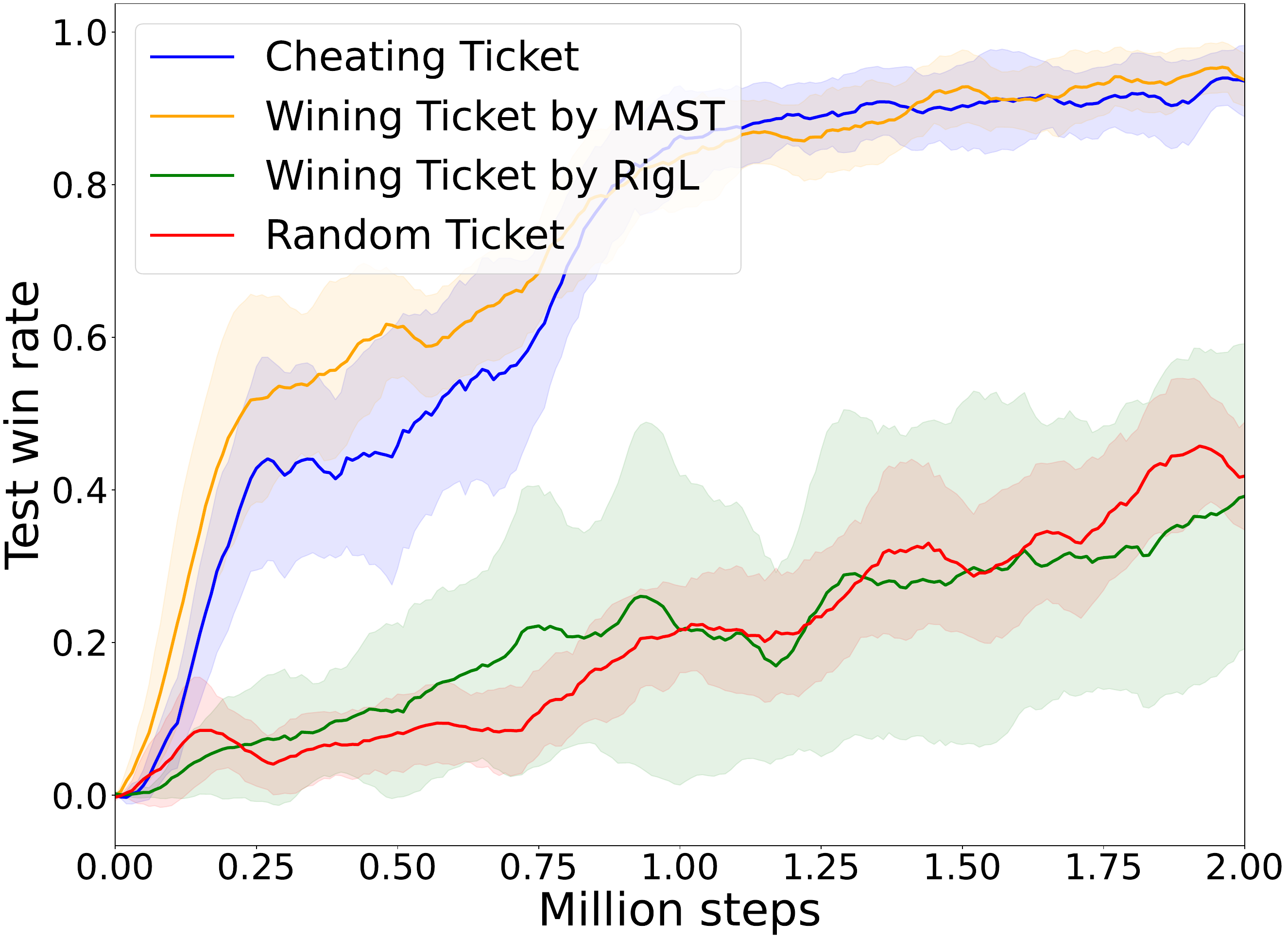}
\caption{Comparison of different sparse masks.}
\label{fig:mask}
\end{wrapfigure}
We conduct a comparative analysis of diverse sparse network architectures. With identical sparsity levels, distinct sparse architectures lead to different hypothesis spaces. As emphasized in \cite{frankle2018lottery}, specific architectures, such as the "winning ticket," outperform randomly generated counterparts. We compare three architectures: the "random ticket" (randomly sampled topology held constant during training), the "winning ticket" (topology from a MAST or RigL run and kept unchanged during training), and the "cheating ticket" (trained by MAST).

Figure~\ref{fig:mask} illustrates that both the "cheating ticket" and "winning ticket" by MAST achieve the highest performance, closely approaching the original dense model's performance. Importantly, using a fixed random topology during training fails to fully exploit the benefits of high sparsity, resulting in significant performance degradation. Furthermore, RigL's "winning ticket" fares poorly, akin to the "random ticket." These results underscore the advantages of our MAST approach, which automatically discovers effective sparse architectures through gradient-based topology evolution, without the need for pretraining methods, e.g., knowledge distillation \cite{schmitt2018kickstarting}. Crucially, our MAST method incorporates key elements: the hybrid TD($\lambda$) mechanism, Soft Mellowmax operator, and dual buffers. Compared to RigL, these components significantly improve value estimation and training stability in sparse models facilitating efficient topology evolution.

\begin{figure}[H]

\centering
\includegraphics[width=\linewidth]{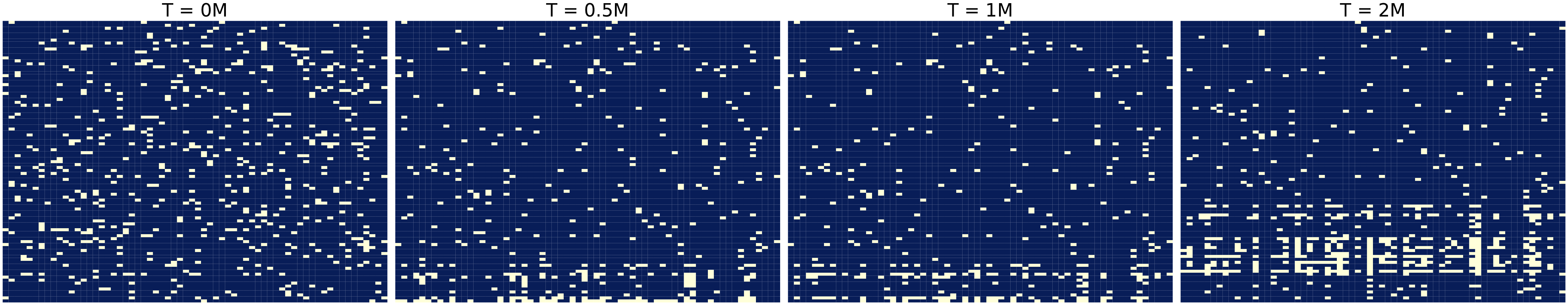}
\caption{Visualization of weight masks in the first hidden layer of agent $1$ by MAST-QMIX.}
\label{fig:vs1}
\end{figure}

Figure~\ref{fig:vs1} showcases the evolving sparse mask of a hidden layer during MAST-QMIX training in {\ttfamily 3s5z}, capturing snapshots at $0$, $5$, $10$, and $20$ million steps. In Figure~\ref{fig:vs1}, the light pixel in row $i$ and column $j$ indicates the existence of the connection for input dimension $j$ and output dimension $i$, while the dark pixel represents the empty connection. Notably, a pronounced shift in the mask is evident at the start of training, followed by a gradual convergence of connections within the layer onto a subset of input neurons. This convergence is discernible from the clustering of light pixels forming continuous rows in the lower segment of the final mask visualization, where several output dimensions exhibit minimal or no connections. This observation underscores the distinct roles played by various neurons in the representation process, showing the prevalent redundancy in dense models.

\begin{figure}[H]
\centering
\subfigure[Agent mask visualization.\label{subfig:mask_role}]{\includegraphics[width=.3\linewidth]{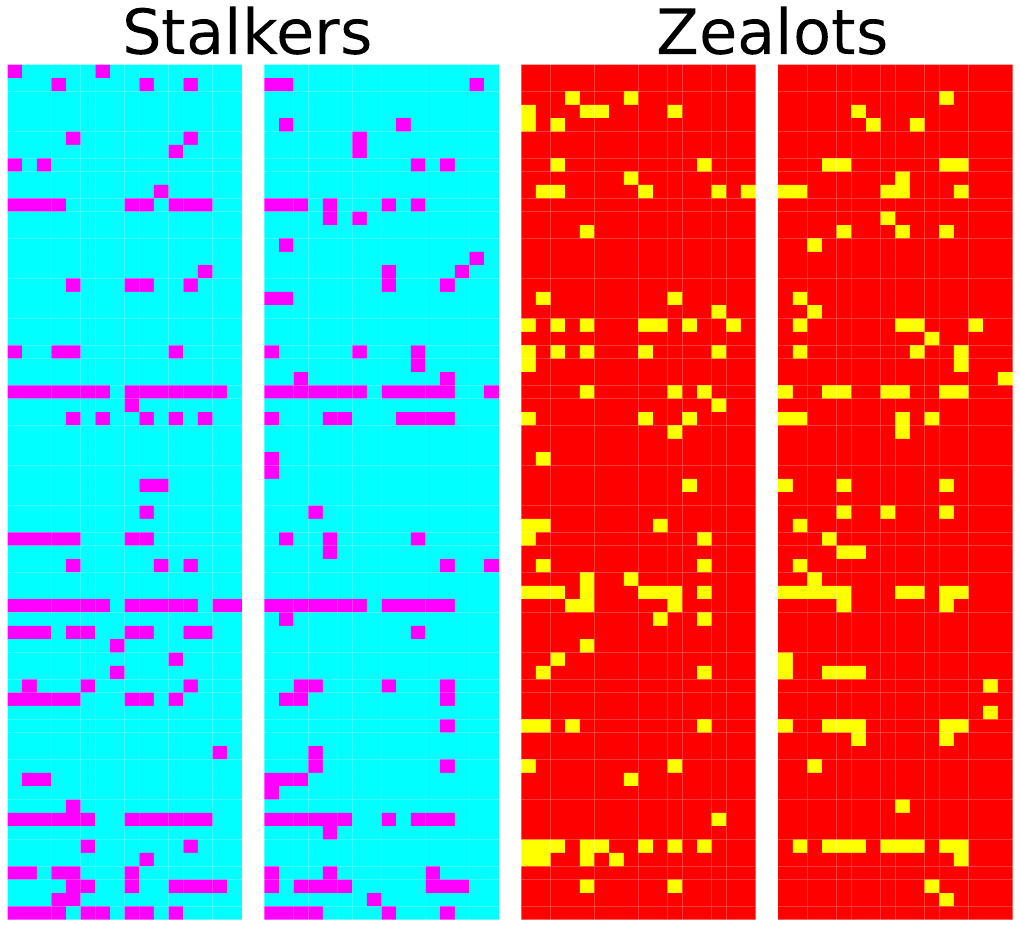}}
\hspace{2cm}
\subfigure[Different sparsity patterns.\label{subfig:role_curve}]{\includegraphics[width=.35\linewidth]{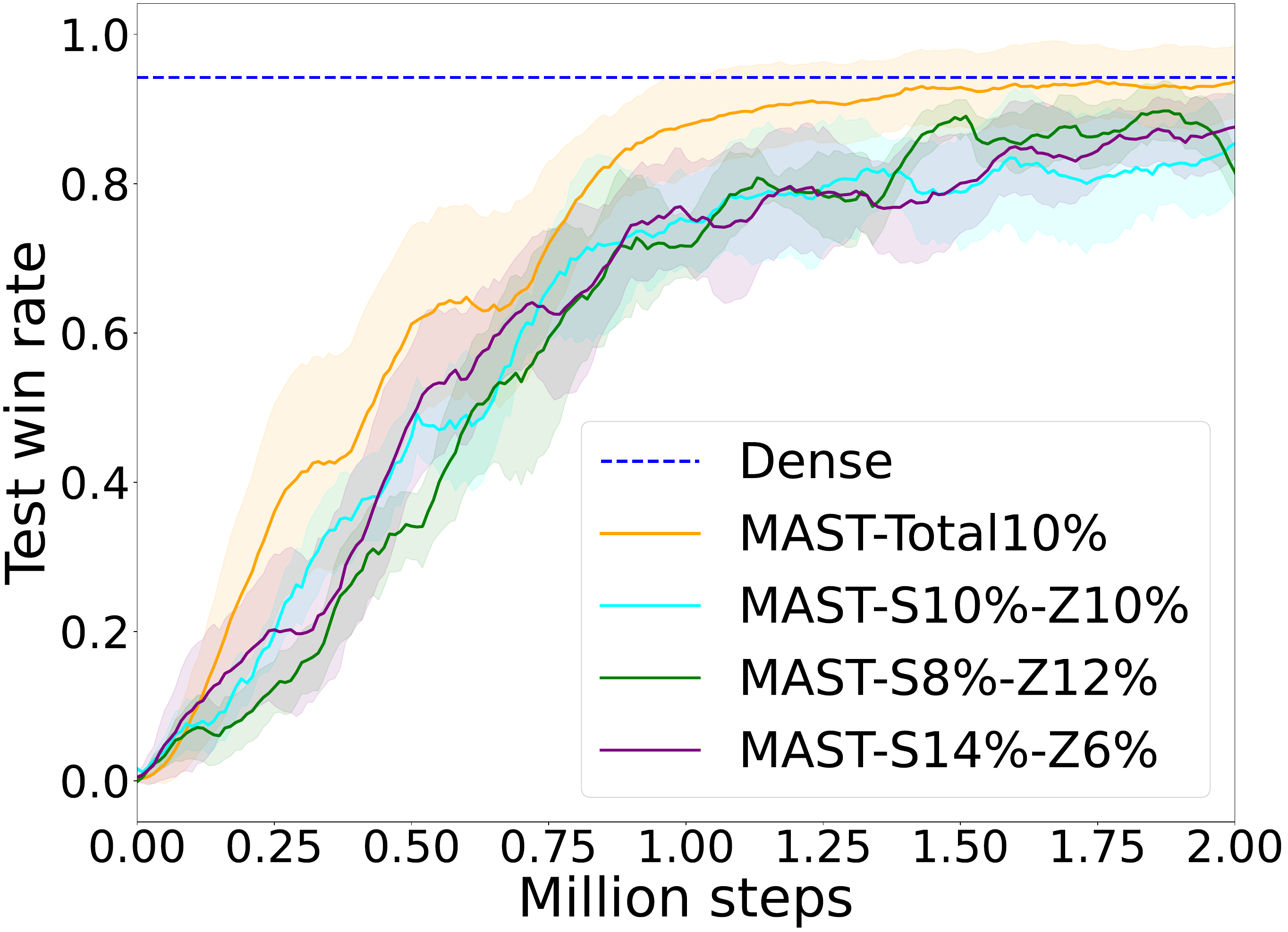}}
\caption{Agent roles.}
\label{fig:overestimation}
\end{figure}

When sparsifying agent networks, MAST only requires setting the total sparsity. The sparsity of different agents is determined automatically by MAST, by concatenating all the agent networks together in our implementation based on PyMARL \cite{samvelyan19smac} (detailed in Appendix~\ref{app:hp}), and treating these networks as a single network during topology evolution with only a total sparsity requirement. 
We visualize the trained masks of different agents in 3s5z in Figure~\ref{subfig:mask_role}, including the masks of two stalkers and two zealots, respectively.

Interestingly, we find that the network topology in the same type of agents looks very similar. However, stalkers have more connections than zealots, which aligns with the fact that stalkers play more critical roles due to their higher attack power and wider attack ranges. This observation highlights an advantage of the MAST framework, i.e., it can automatically  discover the proper sparsity levels for different agents to meet the total sparsity budget. To further validate this point, we compare the adaptive allocation scheme with fixed manually-set patterns in Figure~\ref{subfig:role_curve}. The manual patterns include (stalker-10\%, zealot-10\%), (stalker-8\%, zealot-12\%), and (stalker-14\%, zealot-6\%). The results also show that the adaptive sparsity allocation in MAST outperforms other manual sparsity patterns, demonstrating the superiority of MAST.

\section{Conclusion}
This paper introduces MAST, a novel sparse training framework for valued-based deep MARL. We identify and address value estimation errors and policy inconsistency caused by sparsification, two significant challenges in training sparse agents. MAST offers innovative solutions: a hybrid TD($\lambda$) target mechanism combined with the Soft Mellowmax operator for precise value estimation under extreme sparsity, and a dual buffer mechanism to reduce policy inconsistency and enhance training stability. Extensive experiments  validate MAST's effectiveness in sparse training, achieving model compression ratios of $5\times$ to $20\times$ with minimal performance degradation and up to a remarkable $20\times$ reduction in FLOPs for both training and inference.

\bibliographystyle{unsrt}  
\bibliography{references}  

\newpage
\appendix
\renewcommand{\appendixpagename}{\LARGE Supplementary Materials}
\appendixpage
\startcontents[section]
\printcontents[section]{l}{1}{\setcounter{tocdepth}{2}}

\section{Additional Details for MAST Framework}
\subsection{Comprehensive Related Work}\label{app:rw}
Sparse networks, initially proposed in deep supervised learning can train a 90\%-sparse network without performance degradation from scratch. However, for deep reinforcement learning, the learning target is not fixed but evolves in a bootstrap way \cite{tesauro1995temporal}, and the distribution of the training data can also be non-stationary \cite{desai2019evaluating}, which makes the sparse training more difficult. In the following, we list some representative works for training sparse models from supervised learning to reinforcement learning.

\paragraph{Sparse Models in Supervised Learning} Various techniques have been explored for creating sparse networks, ranging from pruning pre-trained dense networks \cite{han2015learning, han2015deep, srinivas2017training}, to employing methods like derivatives \cite{dong2017learning, molchanov2019importance}, regularization \cite{louizos2017learning}, dropout \cite{molchanov2017variational}, and weight reparameterization \cite{schwarz2021powerpropagation}. Another avenue of research revolves around the Lottery Ticket Hypothesis (LTH) \cite{frankle2018lottery}, which posits the feasibility of training sparse networks from scratch, provided a sparse ``winning ticket" initialization is identified. This hypothesis has garnered support in various deep learning models \cite{chen2020lottery, brix2020successfully}. Additionally, there is a body of work dedicated to training sparse neural networks from the outset, involving techniques that evolve the structures of sparse networks during training. Examples include Deep Rewiring (DeepR) \cite{bellec2017deep}, Sparse Evolutionary Training (SET) \cite{mocanu2018scalable}, Dynamic Sparse Reparameterization (DSR) \cite{mostafa2019parameter}, Sparse Networks from Scratch (SNFS) \cite{dettmers2019sparse}, and Rigged Lottery (RigL) \cite{evci2020rigging}. Furthermore, methods like Single-Shot Network Pruning (SNIP) \cite{lee2018snip} and Gradient Signal Preservation (GraSP) \cite{wang2020picking} are geared towards identifying static sparse networks prior to training.

\paragraph{Sparse Models in Single-Agent RL}
Existing research \cite{schmitt2018kickstarting, zhang2019accelerating} has employed knowledge distillation with static data to ensure training stability and generate small dense agents. Policy Pruning and Shrinking (PoPs) \cite{livne2020pops} generates sparse agents through iterative policy pruning, while the LTH in DRL is first indentified in \cite{yu2019playing}. Another line of investigation aims to train sparse DRL models from scratch, eliminating the necessity of pre-training a dense teacher. Specifically, \cite{sokar2021dynamic} introduces the Sparse Evolutionary Training (SET) approach, achieving a remarkable $50\%$ sparsity level through topology evolution in DRL. Additionally, \cite{graesser2022state} observes that pruning often yields superior results, with plain dynamic sparse training methods, including SET and RigL, significantly outperforming static sparse training approaches. More recently, RLx2 \cite{tan2022rlx2} has demonstrated the capacity to train DRL agents with highly sparse neural networks from scratch. Nevertheless, the application of RLx2 in MARL yields poor results, as demonstrated in Section~\ref{subsec:ce_2}.

\paragraph{Sparse Models in MARL}
Existing works have made attempts to train sparse MARL agents, such as \cite{yang2022learninggroup}, which prunes networks for multiple agents during training, employing weight grouping \cite{wang2019fully}. Another avenue of sparse MARL research seeks to enhance the scalability of MARL algorithms through sparse architectural modifications. For instance, \cite{sun2020scaling} proposes the use of a sparse communication graph with graph neural networks to reduce problem scale. \cite{kim2023parameter} adopts structured pruning for a deep neural network to extend the scalability. Yet another strand of sparse MARL focuses on parameter sharing between agents to reduce the number of trainable parameters, with representative works including \cite{gupta2017cooperative, li2021celebrating, christianos2021scaling}.
However, existing methods fail to maintain high sparsity throughout the training process, such that the FLOPs reduction during training is incomparable to the MAST framework outlined in our paper.

\subsection{Decentralized Partially Observable Markov Decision Process}\label{app:pomdp}
We model the MARL problem as a decentralized partially observable Markov decision process (Dec-POMDP) \cite{oliehoek2016concise}, represented by a tuple $\langle \mathcal{N}, \mathcal{S}, \mathcal{U}, P, r, \mathcal{Z}, O, \gamma\rangle$, where  $\mathcal{N}=\{1,\dots,N\}$ denotes the finite set of agents, $\mathcal{S}$ is the global state space, $\mathcal{U}$ is the action space for an agent, $P$ is the transition probability, $r$ is the reward function, $\mathcal{Z}$ is the observation space for an agent, $O$ is the observation function, and and $\gamma \in[0,1)$ is the discount factor. At each timestep $t$, each agent $i\in\mathcal{N}$ receives an observation $z \in \mathcal{Z}$ from the observation function $O(s,i):\mathcal{S}\times\mathcal{N}\mapsto\mathcal{Z}$ due to partial observability, and chooses an action $u_i \in \mathcal{U}$, which forms a joint action $\boldsymbol{u} \in \boldsymbol{\mathcal{U}} \equiv \mathcal{U}^n$.
The joint action $\boldsymbol{u}$ taken by all agents leads to a transition to the next state $s^{\prime}$ according to transition probability $P(s^{\prime} \mid s, \boldsymbol{u}):\mathcal{S}\times\boldsymbol{\mathcal{U}}\times\mathcal{S}\mapsto[0,1]$ and a joint reward $r(s, \boldsymbol{u}):\mathcal{S}\times\mathcal{U}\mapsto\mathbb{R}$. As the time goes by, each agent $i\in\mathcal{N}$ has an action-observation history $\tau_i \in\boldsymbol{ \mathcal{T}} \equiv(\mathcal{Z} \times \mathcal{U})^*$, where $\boldsymbol{\mathcal{T}}$ is the history space. Based on $\tau_i$, each agent $i$ outputs an action $u_i$ according to its constructed policy $\pi_i(u_i \mid \tau_i):\boldsymbol{\mathcal{T}}\times\mathcal{U}\mapsto[0,1]$. 
The goal of agents is to find an optimal joint policy $\boldsymbol{\pi}=\langle\pi_1, \ldots, \pi_N\rangle$, which maximize the joint cumulative rewards
$J(s_0;\boldsymbol{\pi})=\mathbb{E}_{\boldsymbol{u}_t\sim\boldsymbol{\pi}(\cdot|s_t),s_{t+1}\sim P(\cdot|s_t,\boldsymbol{u}_t)}\left[\sum_{t=0}^{\infty} \gamma^i r(s_t,\boldsymbol{u}_t)\right]$,
where $s_0$ is the initial state. The joint action-value function associated with policy $\boldsymbol{\pi}$ is defined as 
$Q^{\boldsymbol{\pi}}(s_t, \boldsymbol{u}_t)=\mathbb{E}_{\boldsymbol{u}_{t+i}\sim\boldsymbol{\pi}(\cdot|s_{t+i}),s_{t+i+1}\sim P(\cdot|s_{t+i},\boldsymbol{u}_{t+i})}\left[\sum_{i=0}^{\infty} \gamma^i r(s_{t+i},\boldsymbol{u}_{t+i})\right]$.

\subsection{Proof of Theorem~\ref{th:tderror}}\label{app:pf} 
\begin{proof}
By the definitions of multi-step targets, we have
\begin{align}\label{eq:th1_eq1}
&\mathbb{E}_{\rho}[\mathcal{T}_n(s_t,\boldsymbol{u}_t)]\notag\\
=&\mathbb{E}_{\rho}[\sum_{k=0}^{n-1} \gamma^{k}r_{t+k}+\gamma^n\max _{\boldsymbol{u}}Q_\text{tot}\left(s_{t+n}, \boldsymbol{u};\theta\right)]\notag\\
=&\mathbb{E}_{\rho}[\sum_{k=0}^{n-1} \gamma^{k}r_{t+k}+\gamma^nQ_\text{tot}\left(s_{t+n},\pi(s_{t+n});\theta\right)]\notag\\
=&\mathbb{E}_{\rho}[\sum_{k=0}^{n-1} \gamma^{k}r_{t+k}+\gamma^nQ_\text{tot}\left(s_{t+n},\rho(s_{t+n});\phi\right)]\notag\\
&+\gamma^n\mathbb{E}_{\rho}[Q_\text{tot}\left(s_{t+n},\pi(s_{t+n});\theta\right)-Q_\text{tot}\left(s_{t+n},\rho(s_{t+n});\phi\right)]\notag\\
=&\mathbb{E}_{\rho}[\sum_{k=0}^{n-1} \gamma^{k}r_{t+k}+\gamma^nQ_\text{tot}^{\rho}\left(s_{t+n},\rho(s_{t+n})\right)]\notag\\
&+\gamma^n\mathbb{E}_{\rho}[Q_\text{tot}\left(s_{t+n},\rho(s_{t+n});\phi\right)-Q_\text{tot}^{\rho}\left(s_{t+n},\rho(s_{t+n})\right)]\notag\\
&+\gamma^n\mathbb{E}_{\rho}[Q_\text{tot}\left(s_{t+n},\pi(s_{t+n});\theta\right)-Q_\text{tot}\left(s_{t+n},\rho(s_{t+n});\phi\right)]\notag\\
=&Q_\text{tot}^{\rho}\left(s_{t},\boldsymbol{u}_t\right)+\gamma^n\mathbb{E}_{\rho}[\epsilon(s_{t+n},\rho(s_{t+n}))]+\gamma^n\mathbb{E}_{\rho}[Q_\text{tot}\left(s_{t+n},\pi(s_{t+n});\theta\right)-Q_\text{tot}\left(s_{t+n},\rho(s_{t+n});\phi\right)].
\end{align}
Besides, we also have
\begin{align}\label{eq:th1_eq2}
&\mathbb{E}_{\rho}[Q_\text{tot}\left(s_{t+n},\pi(s_{t+n});\theta\right)-Q_\text{tot}\left(s_{t+n},\rho(s_{t+n});\phi\right)]\notag\\
=&\mathbb{E}_{\rho}[Q_\text{tot}\left(s_{t+n},\pi(s_{t+n});\theta\right)-Q_\text{tot}^{\pi}\left(s_{t+n},\pi(s_{t+n})\right)]\notag\\
&+\mathbb{E}_{\rho}[Q_\text{tot}^{\pi}\left(s_{t+n},\pi(s_{t+n})\right)-Q_\text{tot}^{\rho}\left(s_{t+n},\rho(s_{t+n})\right)]\notag\\
&+\mathbb{E}_{\rho}[Q_\text{tot}^{\rho}\left(s_{t+n},\rho(s_{t+n})\right)-Q_\text{tot}\left(s_{t+n},\rho(s_{t+n});\phi\right)]\notag\\
=&\mathbb{E}_{\rho}[\epsilon(s_{t+n},\pi(s_{t+n}))]+\mathbb{E}_{\rho}[Q_\text{tot}^{\pi}\left(s_{t+n},\pi(s_{t+n})\right)-Q_\text{tot}^{\rho}\left(s_{t+n},\rho(s_{t+n})\right)]+\mathbb{E}_{\rho}[\epsilon(s_{t+n},\rho(s_{t+n}))].
\end{align}
Thus, combining Eq.~(\ref{eq:th1_eq1}) and Eq.~(\ref{eq:th1_eq2}) gives
\begin{align*}
&\mathbb{E}_{\rho}[\mathcal{T}_n(s_t,\boldsymbol{u}_t)]-Q_\text{tot}^{\pi}\left(s_t,\boldsymbol{u}_t\right)=\gamma^n\mathbb{E}_{\rho}[\underbrace{2\epsilon(s_{t+n},\rho(s_{t+n}))+\epsilon(s_{t+n},\pi(s_{t+n}))]}_\text{Network fitting error}\\
&+\underbrace{Q_\text{tot}^{\rho}\left(s_{t},\boldsymbol{u}_t\right)-Q_\text{tot}^{\pi}\left(s_t,\boldsymbol{u}_t\right)}_\text{Policy inconsistency error}+\underbrace{\gamma^n\mathbb{E}_{\rho}[Q_\text{tot}^{\pi}\left(s_{t+n},\pi(s_{t+n})\right)-Q_\text{tot}^{\rho}\left(s_{t+n},\rho(s_{t+n})\right)]}_\text{Discounted policy inconsistency error}.
\end{align*}
\end{proof}

\subsection{Proof of Theorem~\ref{th:sm2}}\label{app:sm2}
Relace the Softmax operator in the proof of Theorem~3 in \cite{pan2021regularized} with Eq.~(\ref{eq:sm}) gives the result directly.

\subsection{MAST with Different Algorithms}\label{app:alg}
In this section, we present the pseudocode implementations of MAST for QMIX \cite{rashid2020monotonic} and WQMIX \cite{rashid2020weighted} in Algorithm~\ref{alg:qmix} and Algorithm~\ref{alg:wqmix}, respectively. It is noteworthy that RES \cite{pan2021regularized} exclusively modifies the training target without any alterations to the learning protocol or network structure. Consequently, the implementation of MAST with RES mirrors that of QMIX. 

Crucially, MAST stands as a versatile sparse training framework, applicable to a range of value decomposition-based MARL algorithms, extending well beyond QMIX, WQMIX\footnote{Note that WQMIX encompasses two distinct instantiations, namely Optimistically-Weighted (OW) QMIX and Centrally-Weighted (CW) QMIX. In this paper, we specifically focus on OWQMIX.}, and RES. Furthermore, MAST's three innovative components—hybrid TD($\lambda$), Soft Mellowmax operator, and dual buffer—can be employed independently, depending on the specific algorithm's requirements. This flexible framework empowers the training of sparse networks from the ground up, accommodating a wide array of MARL algorithms.

In the following, we delineate the essential steps of implementing MAST with QMIX (Algorithm~\ref{alg:qmix}). The steps for WQMIX are nearly identical, with the exception of unrestricted agent networks and the unrestricted mixing network's inclusion.
Also, note that we follow the symbol definitions from \cite{colom2021empirical} in Algorithm~\ref{alg:qmix} and \ref{alg:wqmix}.
\paragraph{Gradient-based Topology Evolution:} The process of topology evolution is executed within Lines~\ref{line:te1}-\ref{line:te2} in Algorithm~\ref{alg:qmix}. Specifically, the topology evolution update occurs at intervals of $\Delta_m$ timesteps. For a comprehensive understanding of additional hyperparameters pertaining to topology evolution, please refer to the definitions provided in Algorithm~\ref{alg:updatemask}.
\paragraph{TD Targets:} Hybrid TD($\lambda$) with Soft Mellowmax operator is computed in the Line \ref{line:td} in Algorithm~\ref{alg:qmix}, which modify the TD target $y$ as follows:
\begin{equation}\label{eq:td}
y_\text{S}=\begin{cases}
G_t^{(1)},&\text{ if }t<T_0.\\
(1-\lambda) \sum_{n=1}^{\infty} \lambda^{n-1} \mathcal{T}_{t}^{(n)},&\text{Otherwise.}
\end{cases}
\end{equation}
\vspace{-.4cm}
\begin{algorithm}[H]
\caption{MAST-QMIX}\label{alg:qmix}
\begin{algorithmic}[1]
    \STATE Initialize sparse agent networks, mixing network and hypernetwork with random parameters $\theta$ and random masks $M_\theta$ with determined sparsity $S$ .
    \STATE $\hat{\theta} \leftarrow \theta \odot M_{\theta}$ \emph{~//~Start with a random sparse network}
    \STATE Initialize target networks $\hat{\theta}^- \leftarrow \hat{\theta}$
	\STATE Set the learning rate to $\alpha$
    \STATE Initialize the replay buffer $\mathcal{B}_1\leftarrow\{\}$ with large capacity $C_1$ and $\mathcal{B}_2\leftarrow\{\}$ with small capacity $C_2$
	\STATE Initialize training $\text{step} \leftarrow 0$
	\WHILE{$\text{step} < T_{max}$}
	\STATE $t \leftarrow 0$
    \STATE $s_0\leftarrow\text{initial state}$
        \WHILE{$s_t \neq$ terminal and $t \textless $ episode limit}
        \FOR{each agent a}
        \STATE $\tau^a_t\leftarrow\tau^a_{t-1}\cup\{(o_t,u_{t-1})\}$
        \STATE $\epsilon \leftarrow $ epsilon-schedule(step)
        \STATE $u_t^a\leftarrow \begin{cases}\operatorname{argmax}_{u_t^a} Q\left(\tau_t^a, u_t^a\right) & \text { with probability } 1-\epsilon \\ \operatorname{randint}(1,|U|) & \text { with probability } \epsilon\end{cases}$ \emph{~//~$\epsilon$-greedy exploration}
        \ENDFOR
        \STATE Get reward $r_t$ and next state $s_{t+1}$
        \STATE $\mathcal{B}_1\leftarrow\mathcal{B}_1 \cup\left\{\left(s_t, \mathbf{u}_t, r_t, s_{t+1}\right)\right\}$ \emph{~//~Data in the buffer is of episodes form.}
        \STATE $\mathcal{B}_2\leftarrow\mathcal{B}_2 \cup\left\{\left(s_t, \mathbf{u}_t, r_t, s_{t+1}\right)\right\}$
        \STATE $t\leftarrow t+1$,$\text{step} \leftarrow\text{step}+1$
        \ENDWHILE
        \IF{$|\mathcal{B}_1|>\text { batch-size }$}\label{line:db1}
        \STATE $b\leftarrow \text { random batch of episodes from } \mathcal{B}_1 \text{ and } \mathcal{B}_2 $ \emph{~//~Sample from dual buffers.}
        \FOR{each timestep $t$ in each episode in batch $b$} 
        \STATE
        $$
        Q_{tot}\leftarrow \text{Mixing-network}\left((Q_1(\tau^1_t,u^1_t),\cdots,Q_n(\tau^n_t,u^n_t));\text{Hypernetwork}(s_t;\hat{\theta})\right)
        $$
        \STATE Compute TD target $y$ according to Eq.~(\ref{eq:td}).\label{line:td} \emph{~//~TD($\lambda$) targets with Soft Mellowmax operator.}
        \ENDFOR
        \STATE $\Delta Q_{tot}\leftarrow y - Q_{tot}$
        \STATE $\Delta \hat{\theta}\leftarrow\nabla_{\hat{\theta}}\frac{1}{b}\sum(\Delta Q_{tot })^2 $
        \STATE $\hat{\theta}\leftarrow\hat{\theta}-\alpha \Delta \hat{\theta}$
        \ENDIF\label{line:db2}
        \IF{$\text{step}\mod\Delta_m=0$}\label{line:te1}
        \STATE Topology\_Evolution($\text{networks}_{\hat{\theta}}$ by Algorithm~\ref{alg:updatemask}.
        \ENDIF\label{line:te2}
        \IF{$\text{step}\mod I=0$, where is the target network update interval}
        \STATE $\hat{\theta}^- \leftarrow \hat{\theta}$ \emph{~//~Update target network.}
        \STATE $\hat{\theta}^- \leftarrow \hat{\theta}^- \odot M_{\hat{\theta}} $
        \ENDIF
	\ENDWHILE
\end{algorithmic}
\end{algorithm}
\begin{algorithm}[H]
\caption{MAST-(OW)QMIX}\label{alg:wqmix}
	\begin{algorithmic}[1]
        \STATE Initialize sparse agent networks, mixing network and hypernetwork with random parameters $\theta$ and random masks $M_{\theta}$ with determined sparsity $S$.
        \STATE Initialize unrestricted agent networks and unrestricted mixing network with random parameters $\phi$ and random masks $M_{\phi}$ with determined sparsity $S$.
        \STATE $\hat{\theta} \leftarrow \theta \odot M_{\theta}$ , $\hat{\phi} \leftarrow \phi \odot M_{\phi}$ \emph{~//~Start with a random sparse network}
        \STATE Initialize target networks $\hat{\theta}^- \leftarrow \hat{\theta}$,$\hat{\phi}^- \leftarrow \hat{\phi}$
	\STATE Set the learning to rate $\alpha$ 
        \STATE Initialize the replay buffer $\mathcal{B}_1\leftarrow\{\}$ with large capacity $C_1$ and $\mathcal{B}_2\leftarrow\{\}$ with small capacity $C_2$
	\STATE Initialize training $\text{step} \leftarrow 0$
	\WHILE{$\text{step} < T_{max}$}
	\STATE $t\leftarrow $ 0, 
        \STATE $s_0\leftarrow $ initial state
        \WHILE{$s_t \neq$ terminal and $t \textless $ episode limit}
        \FOR{each agent a}
        \STATE $\tau^a_t\leftarrow\tau^a_{t-1}\cup\{(o_t,u_{t-1})\}$
        \STATE $\epsilon \leftarrow $ epsilon-schedule(step)
        \STATE $u_t^a\leftarrow \begin{cases}\operatorname{argmax}_{u_t^a} Q(\tau_t^a, u_t^a; \hat{\theta}) & \text { with probability } 1-\epsilon \\ \operatorname{randint}(1,|U|) & \text { with probability } \epsilon\end{cases}$ \emph{~//~$\epsilon$-greedy exploration}
        \ENDFOR
        \STATE Get reward $r_t$ and next state $s_{t+1}$
        \STATE $\mathcal{B}_1\leftarrow\mathcal{B}_1 \cup\left\{\left(s_t, \mathbf{u}_t, r_t, s_{t+1}\right)\right\}$ \emph{~//~Data in the buffer is of episodes form.}
        \STATE $\mathcal{B}_2\leftarrow\mathcal{B}_2 \cup\left\{\left(s_t, \mathbf{u}_t, r_t, s_{t+1}\right)\right\}$
        \STATE $t\leftarrow t+1$, $\text{step} \leftarrow \text{step} +1$
        \ENDWHILE
        \IF{$|\mathcal{B}_1|>\text { batch-size }$}
        \STATE $b \leftarrow \text { random batch of episodes from } \mathcal{B}_1 \text{ and } \mathcal{B}_2 $  \emph{~//~Sample from dual buffers.}
        \FOR{each timestep $t$ in each episode in batch $b$} 
        \STATE
        $$
        Q_{tot}\leftarrow\text{Mixing-network}\left((Q_1(\tau^1_t,u^1_t;\hat{\theta}),...,Q_n(\tau^n_t,u^n_t;\hat{\theta})); \text{Hypernetwork}(s_t;\hat{\theta})\right)
        $$
        \STATE $$\hat{Q}^*\leftarrow\text{Unrestricted-Mixing-network}\left(Q_1(\tau^1_t,u^1_t;\hat{\phi}),...,Q_n(\tau^n_t,u^n_t;\hat{\phi}),s_t\right)$$
        \STATE Compute TD target $y$ with target Unrestricted-Mixing network according to Eq.~(\ref{eq:td}). \emph{~//~TD($\lambda$) targets with Soft Mellowmax operator.}
        \STATE $\omega(s_t, \mathbf{u_t})\leftarrow
         \begin{cases} 1, & Q_{tot} < y \\ \alpha, & \text { otherwise. } \end{cases}$
      
        \ENDFOR
        \STATE $\Delta Q_{tot}\leftarrow y - Q_{tot}$
        \STATE $\Delta \hat{\theta} \leftarrow\nabla_{\hat{\theta}} \frac{1}{b}\sum\omega(s,\mathbf{u})(\Delta Q_{tot })^2 $
        \STATE $\hat{\theta}\leftarrow\hat{\theta}-\alpha \Delta \hat{\theta}$
        \STATE $\Delta \hat{Q}^* \leftarrow y - \hat{Q}^*$
        \STATE $\Delta \hat{\phi} \leftarrow\nabla_{\hat{\phi}} \frac{1}{b}\sum(\Delta \hat{Q}^*)^2 $
        \STATE $\hat{\phi}\leftarrow\hat{\phi}-\alpha \Delta \hat{\phi}$
        \ENDIF
        \IF{$\text{step}\mod\Delta_m=0$}
        \STATE Topology\_Evolution($\text{networks}_{\hat{\theta}}$) and Topology\_Evolution($\text{networks}_{\hat{\phi}}$) by Algorithm~\ref{alg:updatemask}.
        \ENDIF
        \IF{$\text{step}\mod I=0$, where is the target network update interval}
        \STATE $\hat{\theta}^- \leftarrow \hat{\theta}, \hat{\phi}^- \leftarrow \hat{\phi}$
        \STATE $\hat{\theta}^- \leftarrow \hat{\theta}^- \odot M_{\hat{\theta}}, \hat{\phi}^- \leftarrow \hat{\phi}^- \odot M_{\hat{\phi}} $
        \ENDIF
	\ENDWHILE
\end{algorithmic}
\end{algorithm}
Here, $\lambda\in[0,1]$ is a hyperparameter, and $\mathcal{T}_{t}^{(n)}=\sum_{i=t}^{t+n} \gamma^{i-t} r_i+\gamma^{n+1}f_s\left(\operatorname{sm}_\omega(\bar{Q}_1(\tau_1,\cdot), \ldots, \operatorname{sm}_\omega(\bar{Q}_N(\tau_N,\cdot)\right)$, where $f_s$ denotes the mixing network and $\bar{Q}_i$ is the target network of $Q_i$. The loss function of MAST, $\mathcal{L}_{\mathrm{S}}(\theta)$, is defined as:
\begin{equation}\label{eq:loss}
    \mathcal{L}_\text{S}(\theta)=\mathbb{E}_{\left(s, \boldsymbol{u}, r, s^{\prime}\right) \sim \mathcal{B}_1\cup \mathcal{B}_2}\left[\left(y_\text{S}-Q_{t o t}(s, \boldsymbol{u})\right)^2\right]
\end{equation}

\paragraph{Dual Buffers:} With the creation of two buffers $\mathcal{B}_1$ and $\mathcal{B}_2$, the gradient update with data sampled from dual buffers is performed in Lines~\ref{line:db1}-\ref{line:db2} in Algorithm~\ref{alg:qmix}.

\subsection{Limitations of MAST}\label{app:limit}
This paper introduces MAST, a novel framework for sparse training in deep MARL, leveraging gradient-based topology evolution to explore network configurations efficiently. However, understanding its limitations is crucial for guiding future research efforts.

\paragraph{Hyperparameters:} MAST relies on multiple hyperparameters for its key components: topology evolution, TD($\lambda$) targets with Soft Mellowmax Operator, and dual buffers. Future work could explore methods to automatically determine these hyperparameters or streamline the sparse training process with fewer tunable settings.

\paragraph{Implementation:} While MAST achieves efficient MARL agent training with minimal performance trade-offs using ultra-sparse networks surpassing 90\% sparsity, its current use of unstructured sparsity poses challenges for running acceleration. The theoretical reduction in FLOPs might not directly translate to reduced running time. Future research should aim to implement MAST in a structured sparsity pattern to bridge this gap between theoretical efficiency and practical implementation.

\section{Experimental Details}\label{app:exp}
In this section, we offer comprehensive experimental insights, encompassing hardware configurations, environment specifications, hyperparameter settings, model size computations, FLOPs calculations, and supplementary experimental findings.

\subsection{Hardware Setup}\label{app:hardware}
Our experiments are implemented with PyTorch 2.0.0 \cite{paszke2017automatic} and run on $4\times$ NVIDIA GTX Titan X (Pascal) GPUs. Each run needs about $12\sim24$ hours for QMIX or WQMIX, and about $24\sim72$ hours for RES for two million steps. depends on the environment types.

\subsection{Environment}\label{app:environment}
\vspace{-.2cm}
We assess the performance of our MAST framework using the SMAC benchmark \cite{samvelyan19smac}, a dedicated platform for collaborative multi-agent reinforcement learning research based on Blizzard's StarCraft II real-time strategy game, specifically version 4.10. It is important to note that performance may vary across different versions. Our experimental evaluation encompasses four distinct maps, each of which is described in detail below.
\vspace{-.2cm}
\begin{itemize}
    \item {\ttfamily \!3m}: An easy map, where the agents are $3$ Marines, and the enemiesa are $3$ Marines.
    \item {\ttfamily \!2s3z}: An easy map, where the agents are $2$ Stalkers and $3$ Zealots, and the enemies are $2$ Stalkers and $3$ Zealots.
    \item {\ttfamily \!3s5z}: An easy map, where the agents are $3$ Stalkers and $5$ Zealots, and the enemies are $3$ Stalkers and $5$ Zealots.
    \item {\ttfamily \!2c\_vs\_64zg}: A hard map, where the agents are $2$ Colossi, and the enemies are $64$ Zerglings.
\end{itemize}

We also evaluate MAST on the Multi-Agent MuJoCo (MAMuJoCo) benchmark from \cite{peng2021facmac}, which is an environment designed for evaluating continuous MARL algorithms, focusing on cooperative robotic control tasks. It extends the single-agent MuJoCo framework included with OpenAI Gym \cite{gym}. Inspired by modular robotics, MAMuJoCo includes scenarios with a large number of agents, aiming to stimulate progress in continuous MARL by providing diverse and challenging tasks for decentralized coordination. We test MAST-COMIX in the Humanoid, Humanoid Standup, and ManyAgent Swimmer scenarios in MAMuJoCo. The environments are tested using their default configurations, with other settings following FACMAC \cite{peng2021facmac}. Specifically, we set the maximum observation distance to $k=0$. In the ManyAgent Swimmer scenario, we configure 10 agents, each controlling a consecutive segment of length 2. The agent network architecture of MAST-COMIX uses an MLP with two hidden layers of 400 dimensions each, following the settings in FACMAC. All other hyperparameters are also based on FACMAC.

\subsection{Hyperparameter Settings}\label{app:hp}
\vspace{-.2cm}
Table~\ref{tb:hyper} provides a comprehensive overview of the hyperparameters employed in our experiments for MAST-QMIX, MAST-WQMIX, and MAST-RES. It includes detailed specifications for network parameters, RL parameters, and topology evolution parameters, allowing for a thorough understanding of our configurations. Besides, MAST is implemented based on the PyMARL \cite{samvelyan19smac} framework with the same network structures and hyperparameters as given in Table~\ref{tb:hyper}.
We also provide a hyperparameter recommendation for three key components, i.e. gradient-based topology evolution, Soft Mellowmax enabled hybrid TD($\lambda$) targets and dual buffers, in Table~\ref{tb:key} for deployment MAST framework in other problems.
\begin{table}[H]
\centering
\caption{Recommendation for Key Hyperparameters in MAST.}
\label{tb:key}
\centering
\begin{tabular}{c|c|c}
\toprule
Category &\textbf{Hyperparameter} &\textbf{Value} \\
\midrule
\makecell[c]{Topology\\Evolution}
& Initial mask update fraction $\zeta_0$ & 0.5 \\\cline{2-3}
& Mask update interval $\Delta_m$ & $200$ episodes \\
\midrule
\makecell[c]{TD Targets}
& Burn-in time $T_0$ & $3/8$ of total training steps\\\cline{2-3}
& $\lambda$ value in TD($\lambda$) & 0.6 or 0.8\\\cline{2-3}
& $\alpha$ in soft mellow-max operator & 1 \\\cline{2-3}
& $\omega$ in soft mellow-max operator & 10 \\
\midrule 
\makecell[c]{Dual\\Buffer}
& Offline buffer size $C_1$ & $5\times 10^3$ episodes\\\cline{2-3}
& Online buffer size $C_2$ & $128$ episodes\\\cline{2-3}
& \makecell[c]{Sample partition\\of online and offline buffer} & 2:6\\
\bottomrule
\end{tabular}
\end{table}
\begin{table}[htbp]
\caption{Hyperparameters of MAST-QMIX, MAST-WQMIX and MAST-RES.}
\label{tb:hyper}
\centering
\begin{tabular}{c|c|l}
\toprule
Category &\textbf{Hyperparameter} &\textbf{Value} \\
\midrule
\multirow{20}{*}{\makecell[c]{Shared\\Hyperparameters}} &	Optimizer & RMSProp \\\cline{2-3}
&Learning rate $\alpha$ & $5\times 10^{-4}$ \\\cline{2-3}
&Discount factor $\gamma$ & 0.99 \\\cline{2-3}
&\makecell[c]{Number of hidden units \\ per layer of agent network} & 64\\ \cline{2-3}
& \makecell[c]{Hidden dimensions in the \\ GRU layer of agent network} & 64\\ \cline{2-3}
& \makecell[c]{Embedded dimensions \\ of mixing network} & 32\\\cline{2-3}
& \makecell[c]{Hypernet layers \\ of mixing network} & 2 \\
& \makecell[c]{ Embedded dimensions \\ of hypernetwork} & 64  \\\cline{2-3}
& Activation Function & ReLU \\ \cline{2-3}
& Batch size $B$ & 32 episodes \\\cline{2-3}
& Warmup steps & 50000\\\cline{2-3}
& Initial $\epsilon$ & 1.0\\\cline{2-3}
& Final $\epsilon$ & 0.05\\\cline{2-3}
& Double DQN update & True\\\cline{2-3}
& Target network update interval $I$ & $200$ episodes\\\cline{2-3}
& Initial mask update fraction $\zeta_0$ & 0.5 \\\cline{2-3}
& Mask update interval $\Delta_m$ & timesteps of $200$ episodes \\\cline{2-3}
& Offline buffer size $C_1$ & $5\times 10^3$ episodes\\\cline{2-3}
& Online buffer size $C_2$ & $128$ episodes\\\cline{2-3}
& Burn-in time $T_0$ & $7.5\times 10^5$\\\cline{2-3}
& $\alpha$ in soft mellow-max operator & 1 \\\cline{2-3}
& $\omega$ in soft mellow-max operator & 10 \\\cline{2-3}
& \makecell[c]{Number of episodes \\ in a sampled batch \\ of offline buffer $S_1$}  & 20\\\cline{2-3}
& \makecell[c]{ Number of episodes \\ in a sampled batch \\ of online buffer $S_2$} & 12\\
\midrule
\makecell[c]{Hyperparameters\\for MAST-QMIX} 
&Linearly annealing steps for $\epsilon$ & 50k\\\cline{2-3}
& $\lambda$ value in TD($\lambda$) & 0.8\\
\midrule 
\makecell[c]{Hyperparameters\\for MAST-WQMIX}
&Linearly annealing steps for $\epsilon$ & 100k\\\cline{2-3}
& $\lambda$ value in TD($\lambda$) & 0.6\\\cline{2-3}
&Coefficient of $Q_{tot}$ loss & 1 \\\cline{2-3}
&Coefficient of $\hat{Q}^*$ loss & 1 \\\cline{2-3}
& \makecell[c]{Embedded dimensions of \\ unrestricted mixing network} & 256\\\cline{2-3}
& \makecell[c]{Embedded number of actions \\ of unrestricted agent network} & 1 \\\cline{2-3}
& $\alpha$ in weighting function & 0.1 \\
\midrule 
\makecell[c]{Hyperparameters\\for MAST-RES}
&Linearly annealing steps for $\epsilon$ & 50k\\\cline{2-3}
& $\lambda$ value in TD($\lambda$) & 0.8\\\cline{2-3}
& $\lambda$ value in Softmax operator & 0.05\\\cline{2-3}
&Inverse temperature $\beta$ & 5.0\\
\bottomrule
\end{tabular}
\end{table}

Besides, to extend the existing QMIX to continuous action spaces, we utilized COMIX from FACMAC \cite{peng2021facmac} for our experiments, which employs the Cross-Entropy Method (CEM) for approximate greedy action selection. The hyperparameter configuration of CEM also follows FACMAC settings.

\subsection{Calculation of Model Sizes and FLOPs}\label{app:flops_ms}
\vspace{-.2cm}
\subsubsection{Model Size}
\vspace{-.2cm}
First, we delineate the calculation of model sizes, which refers to the total number of parameters within the model.
\vspace{-.2cm}
\begin{itemize}
    \item For a sparse network with $L$ fully-connected layers, the model size, as expressed in prior works \cite{evci2020rigging, tan2022rlx2}, can be computed using the equation:
\begin{equation}\label{eq:linear_ms}
    M_\text{linear} = \sum_{l=1}^L(1-S_l) I_l O_l,
\end{equation}
where $S_l$ represents the sparsity, $I_l$ is the input dimensionality, and $O_l$ is the output dimensionality of the $l$-th layer. 

\item For a sparse network with $L$ GRU layers, considering the presence of $3$ gates in a single layer, the model size can be determined using the equation:
\begin{equation}\label{eq:gru_ms}
    M_\text{GRU} = \sum_{l=1}^L(1-S_l)\times 3\times h_l \times (h_l+I_l),
\end{equation}
where $h_l$ represents the hidden state dimensionality.
\end{itemize}
\vspace{-.2cm}
Specifically, the "Total Size" column in Table~\ref{tb:comp_eval_2} within the manuscript encompasses the model size, including both agent and mixing networks during training. For QMIX, WQMIX, and RES, target networks are employed as target agent networks and target mixing networks. We denote the model sizes of the agent network, mixing network, unrestricted agent network, and unrestricted mixing network as $M_{\text{Agent}}$, $M_{\text{Mix}}$, $M_{\text{Unrestricted-Agent}}$, and $M_{\text{Unrestricted-Mix}}$, respectively. Detailed calculations of these model sizes are provided in the second column of Table~\ref{tb:flops}.
\begin{table}[H]
\caption{FLOPs and model size for MAST-QMIX , MAST-WQMIX and MAST-RES. } 
\label{tb:flops}
\centering
\begin{tabular}{l|c|c|c}
\toprule
Algorithm & Model size & \makecell[c]{Training \\ FLOPs} & \makecell[c]{Inference \\ FLOPs} \\
\midrule
MAST-QMIX & $2M_{\text{Agent}}+2M_{\text{Mix}}$ & $4B(\text{FLOPs}_{\text{Agent}}+\text{FLOPs}_{\text{Mix}})$ & $\text{FLOPs}_{\text{Agent}}$ \\\cline{2-4}
MAST-WQMIX & \makecell[c]{$M_{\text{Agent}}+M_{\text{Mix}}+$\\$2M_{\text{Unrestricted-Agent}}+$ \\ $2M_{\text{Unrestricted-Mix}}$} & \makecell[c]{$3B(\text{FLOPs}_{\text{Agent}}+\text{FLOPs}_{\text{Mix}})+$\\$4B\cdot\text{FLOPs}_{\text{Unrestricted-Agent}}+$\\$4B\cdot\text{FLOPs}_{\text{Unrestricted-Mix}}$} & $\text{FLOPs}_{\text{Agent}}$ \\\cline{2-4}
MAST-RES & $2M_{\text{Agent}}+2M_{\text{Mix}}$ & \makecell[c]{$4B\cdot\text{FLOPs}_{\text{Agent}}+$\\
$(5+nm)B\cdot\text{FLOPs}_{\text{Mix}}$} & $\text{FLOPs}_{\text{Agent}}$ \\
\bottomrule
\end{tabular}
\end{table}
\vspace{-.6cm}
\subsubsection{FLOPs Calculation}\label{app:flops}
\vspace{-.2cm}
Initially, for a sparse network with $L$ fully-connected layers, the required FLOPs for a forward pass are computed as follows (also adopted in \cite{evci2020rigging} and \cite{tan2022rlx2}): 
\begin{equation}
\begin{aligned}
\text{FLOPs} = \sum_{l=1}^L(1-S_l) (2I_l-1)O_l,
\end{aligned}
\end{equation}
where $S_l$ is the sparsity, $I_l$ is the input dimensionality, and $O_l$ is the output dimensionality of the $l$-th layer. Similarly, for a sparse network with $L$ GRU \cite{chung2014empirical} layers, considering the presence of $3$ gates in a single layer, the required FLOPs for a forward pass are:
\begin{equation}
\begin{aligned}
\text{FLOPs} = \sum_{l=1}^L(1-S_l)\times3\times h_l\times [2(h_l+I_l)-1],
\end{aligned}
\end{equation}
where $h_l$ is the hidden state dimensionality.

We denote $B$ as the batch size employed in the training process, and $\text{FLOPs}_{\text{Agent}}$ and $\text{FLOPs}_{\text{Mix}}$ as the FLOPs required for a forward pass in the agent and mixing networks, respectively. The inference FLOPs correspond exactly to $\text{FLOPs}_{\text{Agent}}$, as detailed in the last column of Table~\ref{tb:flops}. When it comes to training FLOPs, the calculation encompasses multiple forward and backward passes across various networks, which will be thoroughly elucidated later. Specifically, we compute the FLOPs necessary for each training iteration. Additionally, we omit the FLOPs associated with the following processes, as they exert minimal influence on the ultimate result:
\vspace{-.2cm}
\begin{itemize}
    \item \textbf{Interaction with the environment:} This operation, where agents decide actions for interaction with the environment, incurs FLOPs equivalent to $\text{FLOPs}_{\text{Agent}}$. Notably, this value is considerably smaller than the FLOPs required for network updates, as evident in Table~\ref{tb:flops}, given that $B\gg1$.
    \item \textbf{Updating target networks:} Each parameter in the networks is updated as $\theta' \leftarrow \theta$. Consequently, the number of FLOPs in this step mirrors the model size, and is thus negligible.
    \item \textbf{Topology evolution:} This element is executed every $200$ gradient updates. To be precise, the average FLOPs involved in topology evolution are computed as $B \times \frac{2\text{FLOPs}_{\text{Agent}}}{(1-S^{(a)})\Delta_m}$ for the agent, and $B \times \frac{2\text{FLOPs}_{\text{Mix}}}{(1-S^{(m)})\Delta_m}$ for the mixer. Given that $\Delta_m=200$, the FLOPs incurred by topology evolution are negligible.
\end{itemize}

Therefore, our primary focus shifts to the FLOPs related to updating the agent and mixer. We will first delve into the details for QMIX, with similar considerations for WQMIX and RES.

\subsubsection{Training FLOPs Calculation in QMIX}
\vspace{-.2cm}
Recall the way to update networks in QMIX is given by
\begin{equation}
    \theta\leftarrow\theta-\alpha \nabla_{\theta} \frac{1}{B}\sum(y_t-Q_{tot}(s_i,a_i;\theta))^2,
\end{equation}
where $B$ is the batch size. Subsequently, we can compute the FLOPs of training as:
\begin{equation}
\text{FLOPs}_{\text{train}}=\text{FLOPs}_{\text{TD\_target}}+\text{FLOPs}_{\text{compute\_loss}}+\text{FLOPs}_{\text{backward\_pass}},\label{eq:critic_flops}
\end{equation}
where $\text{FLOPs}_{\text{TD\_target}}$, $\text{FLOPs}_{\text{compute\_loss}}$, and $\text{FLOPs}_{\text{backward\_pass}}$ refer to the numbers of FLOPs in computing the TD targets in forward pass, loss function  in forward pass, and gradients in backward pass (backward-propagation), respectively. 
By Eq.~(\ref{eq:td}) and (\ref{eq:loss}), we have:
\begin{equation}
\begin{aligned}
\text{FLOPs}_{\text{TD\_target}}=&\,B\times(\text{FLOPs}_{\text{Agent}}+\text{FLOPs}_{\text{Mix}}),
\\
\text{FLOPs}_{\text{compute\_loss}}=&\,B\times(\text{FLOPs}_{\text{Agent}}+\text{FLOPs}_{\text{Mix}}).\label{eq:app_critic_flops1}
\end{aligned}
\end{equation}
For the FLOPs of gradients backward propagation, $\text{FLOPs}_{\text{backward\_pass}}$, we compute it as two times the computational expense of the forward pass, which is adopted in existing literature \cite{evci2020rigging}, i.e.,
\begin{equation}
\text{FLOPs}_{\text{backward\_pass}}=B\times 2\times(\text{FLOPs}_{\text{Agent}}+\text{FLOPs}_{\text{Mix}}),\label{eq:app_critic_flops2}
\end{equation}

Combining Eq.~(\ref{eq:critic_flops}), Eq.~(\ref{eq:app_critic_flops1}), and Eq.~(\ref{eq:app_critic_flops2}), the FLOPs of training in QMIX is:
\begin{equation}
\text{FLOPs}_{\text{train}}=B\times4\times(\text{FLOPs}_{\text{Agent}}+\text{FLOPs}_{\text{Mix}}).
\end{equation}

\subsubsection{Training FLOPs Calculation in WQMIX}
\vspace{-.2cm}
The way to update the networks in WQMIX is different from that in QMIX. Specifically, denote the parameters of the original network and unrestricted network as $\theta$ and $\phi$, respectively, which are updated according to
\begin{equation}
    \begin{aligned}
    \theta\leftarrow&\theta-\alpha \nabla_{\theta} \frac{1}{B}\sum_i \omega(s_i,a_i)(\mathcal{T}_\lambda-Q_{tot}(s_i,a_i;\theta))^2\\
\phi\leftarrow&\phi-\alpha \nabla_{\phi} \frac{1}{B}\sum_i(\mathcal{T}_\lambda-\hat{Q^*}(s_i,a_i;\phi))^2
\end{aligned},
\end{equation}
where $B$ is the batch size, $\omega$ is the weighting function, $\hat{Q^*}$ is the unrestricted joint action value function.
As shown in Algorithm~\ref{alg:wqmix}, the way to compute TD target in WQMIX is different from that in QMIX. Thus, we have
\begin{equation}
\text{FLOPs}_{\text{TD\_target}}=B\times(\text{FLOPs}_{\text{Unrestricted-Agent}}+\text{FLOPs}_{\text{Unrestricted-Mix}}).
\end{equation}
In this paper, we take an experiment on one of two instantiations of QMIX. i.e., OW-QMIX \cite{rashid2020weighted}. Thus, the number of FLOPs in computing loss is
\begin{equation}
\text{FLOPs}_{\text{compute\_loss}}=B\times(\text{FLOPs}_{\text{Agent}}+\text{FLOPs}_{\text{Mix}}+\text{FLOPs}_{\text{Unrestricted-Agent}}+\text{FLOPs}_{\text{Unrestricted-Mix}}).
\end{equation}
where unrestricted-agent and unrestricted-mix have similar network architectures as  $Q_{tot}$ and $Q_{tot}$ to, respevtively. The FLOPs of gradients backward propagation can be given as
\begin{equation}
\text{FLOPs}_{\text{backward\_pass}}=B\times 2\times (\text{FLOPs}_{\text{Agent}}+\text{FLOPs}_{\text{Mix}}+\text{FLOPs}_{\text{Unrestricted-Agent}}+\text{FLOPs}_{\text{Unrestricted-Mix}}).
\end{equation}
Thus, the FLOPs of training in WQMIX can be computed by 
\begin{equation}
\text{FLOPs}_{\text{train}}=B\times(3\text{FLOPs}_{\text{Agent}}+3\text{FLOPs}_{\text{Mix}}+4\text{FLOPs}_{\text{Unrestricted-Agent}}+4\text{FLOPs}_{\text{Unrestricted-Mix}}).
\end{equation}
\vspace{-.8cm}
\subsubsection{Training FLOPs Calculation in RES}\label{app:softmax}
\vspace{-.2cm}
Calculations of FLOPs for RES are similar to those in QMIX. The way to update the network parameter in RES is:
\begin{equation}
    \theta\leftarrow\theta-\alpha \nabla_{\theta} \frac{1}{B}\sum_i(\mathcal{T}_\lambda-Q_{tot}(s_i,a_i;\theta))^2,
\end{equation}
where $B$ is the batch size. Meanwhile, note that the way to compute TD target in RES \cite{pan2021regularized} includes computing the \emph{approximate Softmax operator}, we have:
\begin{equation}
    \text{FLOPs}_{\text{TD\_target}}=B\times (\text{FLOPs}_{\text{Agent}}+n\times m\times(2\text{FLOPs}_{\text{Mix}})),
\end{equation}
where $n$ is the number of agents, $m$ is the maximum number of actions an agent can take in a scenario.
Other terms for updating networks are the same as QMIX. Thus, the FLOPs of
training in RES can be computed by
\begin{equation}
    \text{FLOPs}_{\text{train}}=B\times(4\text{FLOPs}_{\text{Agent}}+(3+2\times n\times m)\text{FLOPs}_{\text{Mix}}).
\end{equation}
\vspace{-1cm}
\subsection{Training Curves of Comparative Evaluation in Section~\ref{subsec:ce_2}}
\begin{figure}[H]
\centering
\subfigure[MAST-QMIX $95\%$ on {\ttfamily 3m}]{\includegraphics[width=.38\linewidth]{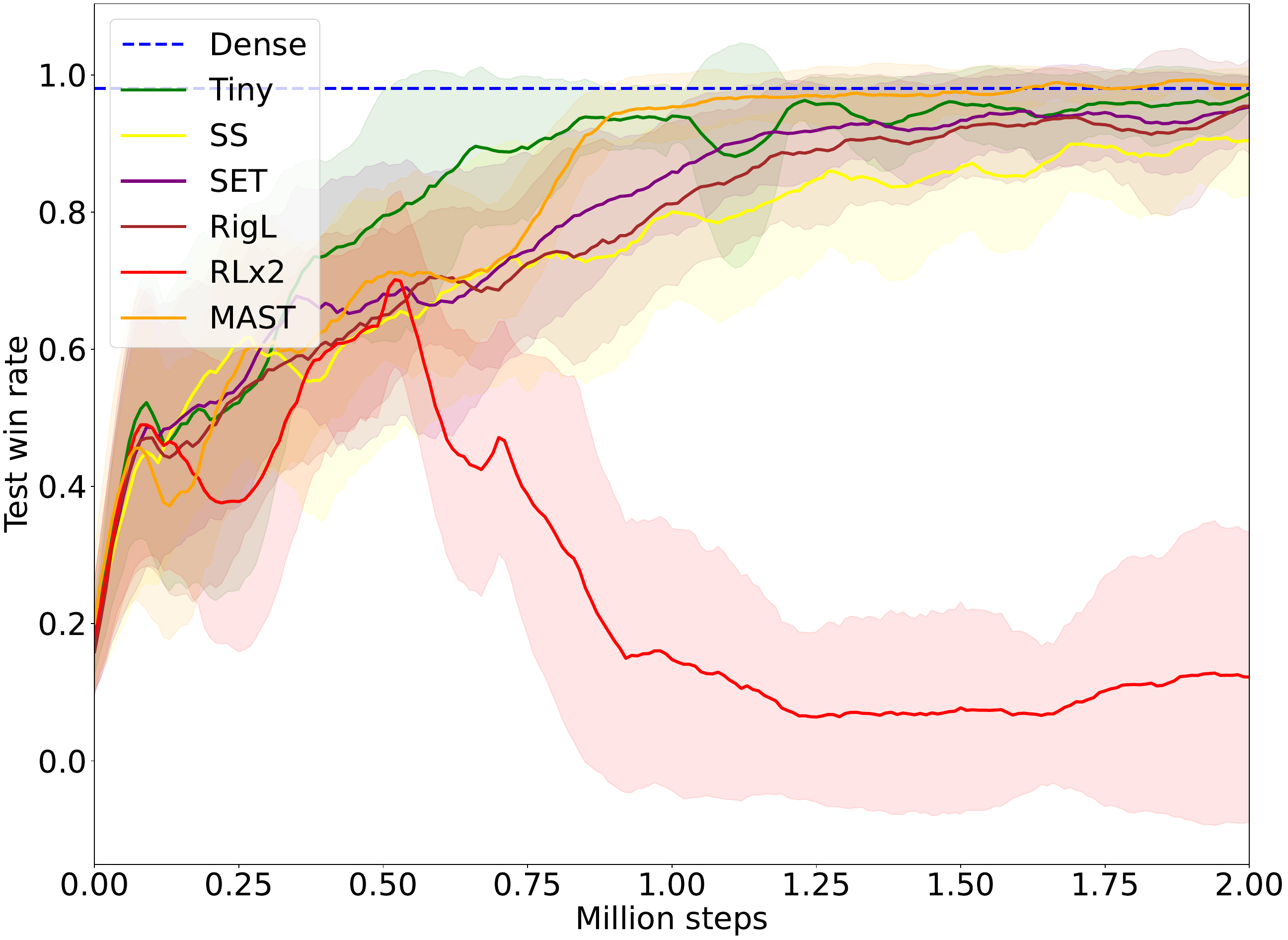}}
\subfigure[MAST-QMIX $95\%$ on {\ttfamily 2s3z}]{\includegraphics[width=.38\linewidth]{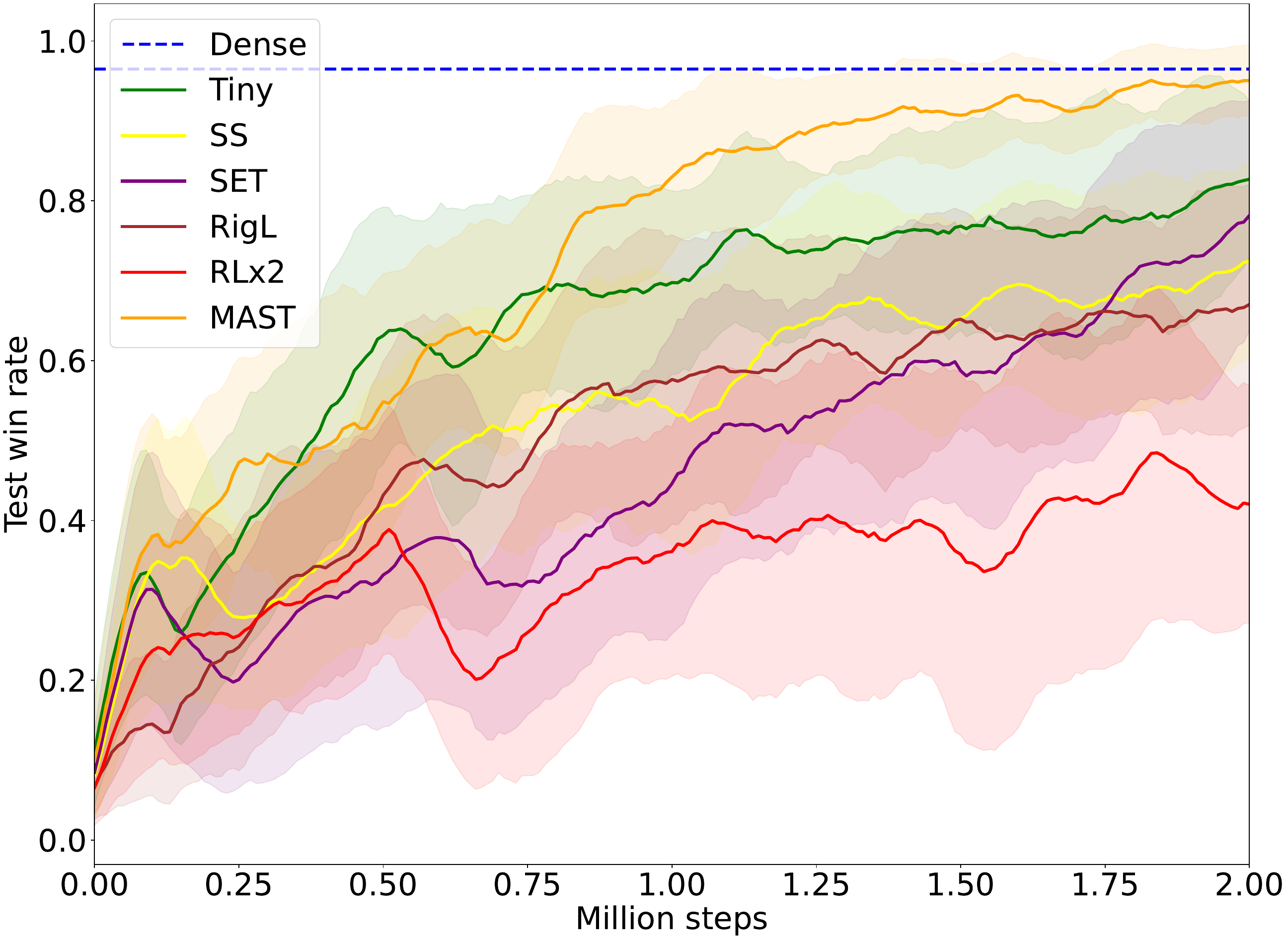}}
\subfigure[MAST-QMIX $90\%$ on {\ttfamily 3s5z}]{\includegraphics[width=.38\linewidth]{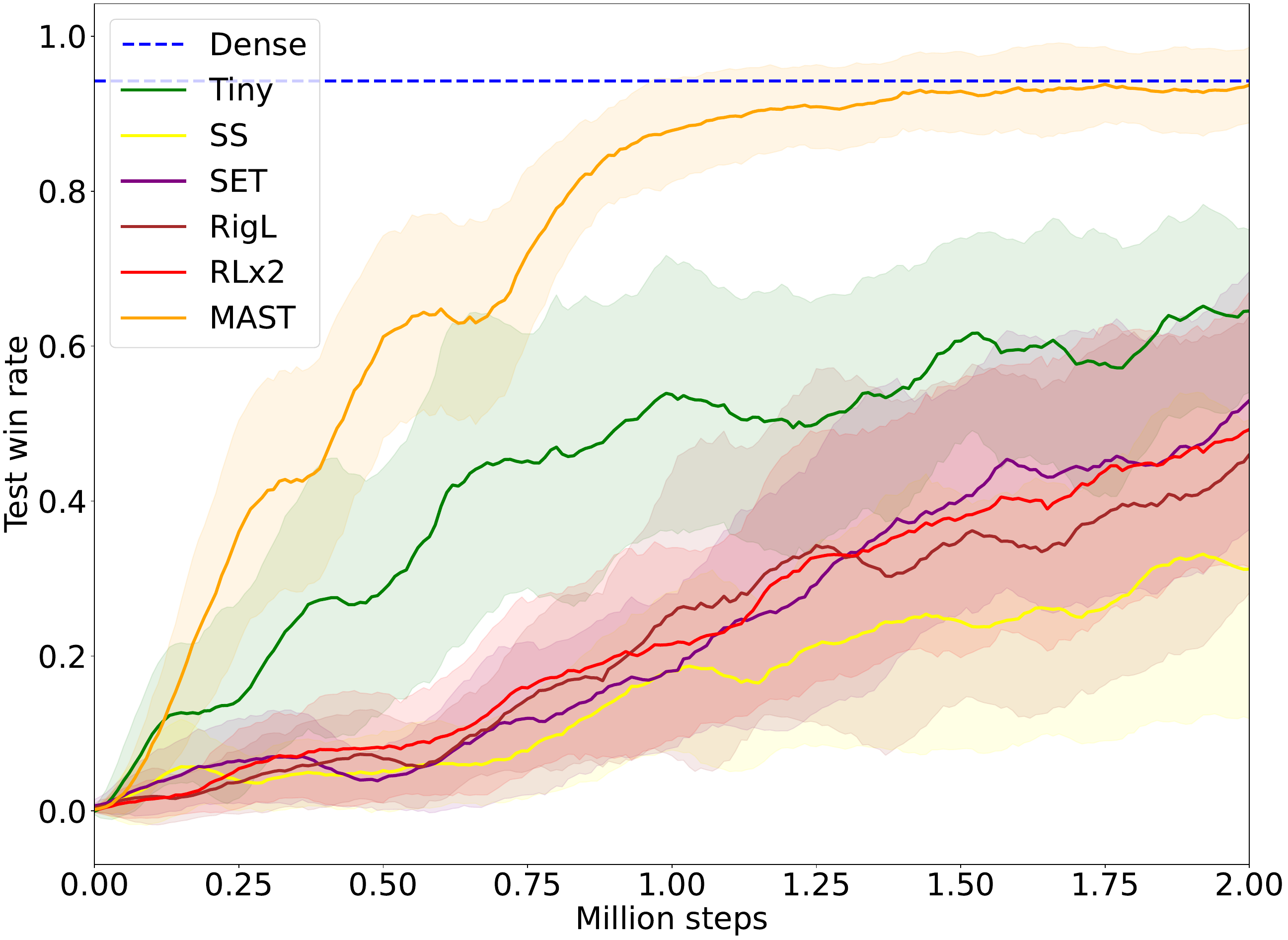}}
\subfigure[MAST-QMIX $90\%$ on {\ttfamily 64zg}]{\includegraphics[width=.38\linewidth]{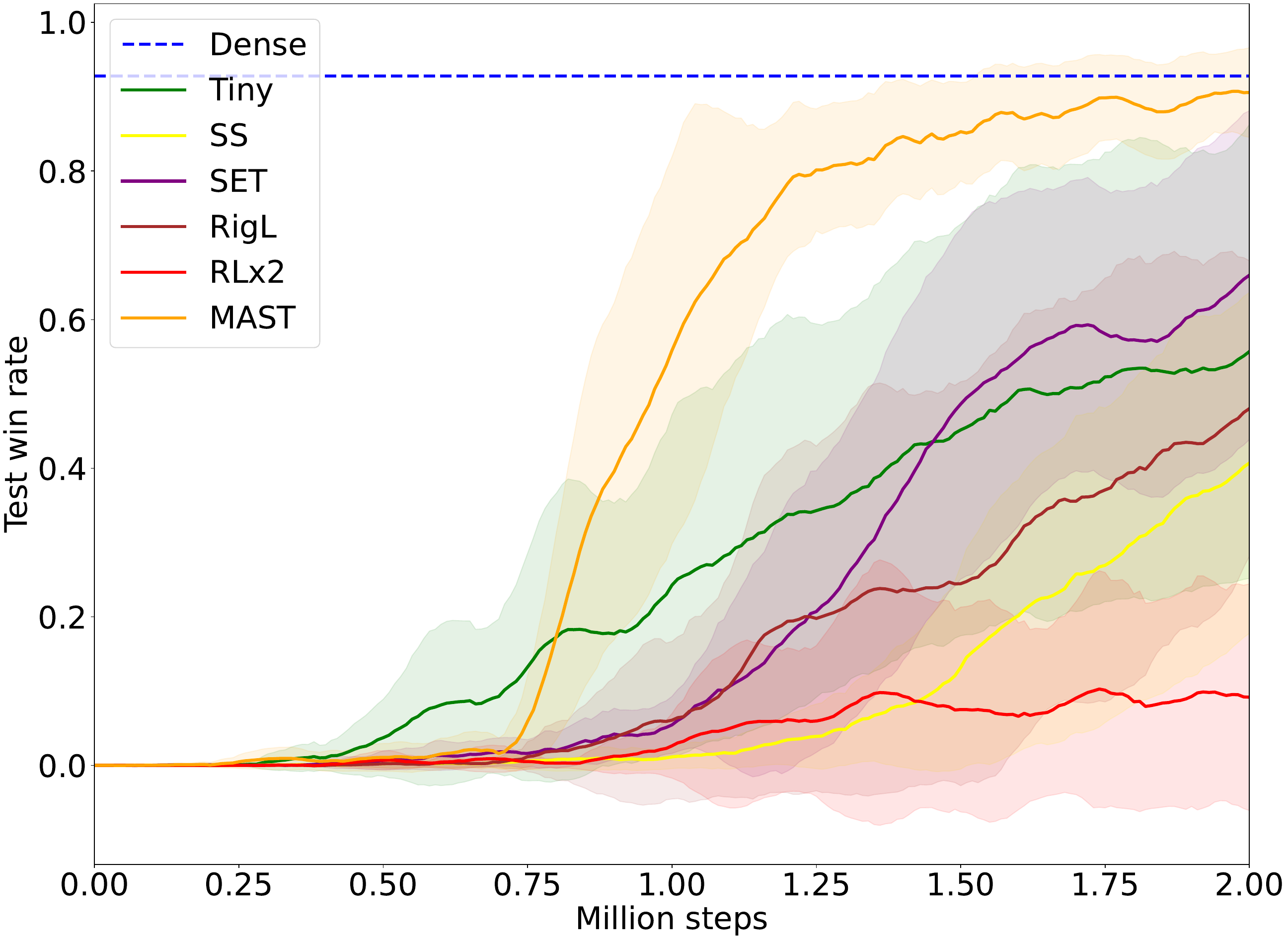}}
\vspace{-.4cm}
\caption{Training processes of MAST-QMIX on four SAMC benchmarks.}
\label{fig:performance_qmix}
\end{figure}

\begin{figure}[H]
\centering
\subfigure[MAST-WQMIX $90\%$ on {\ttfamily 3m}]{\includegraphics[width=.38\linewidth]{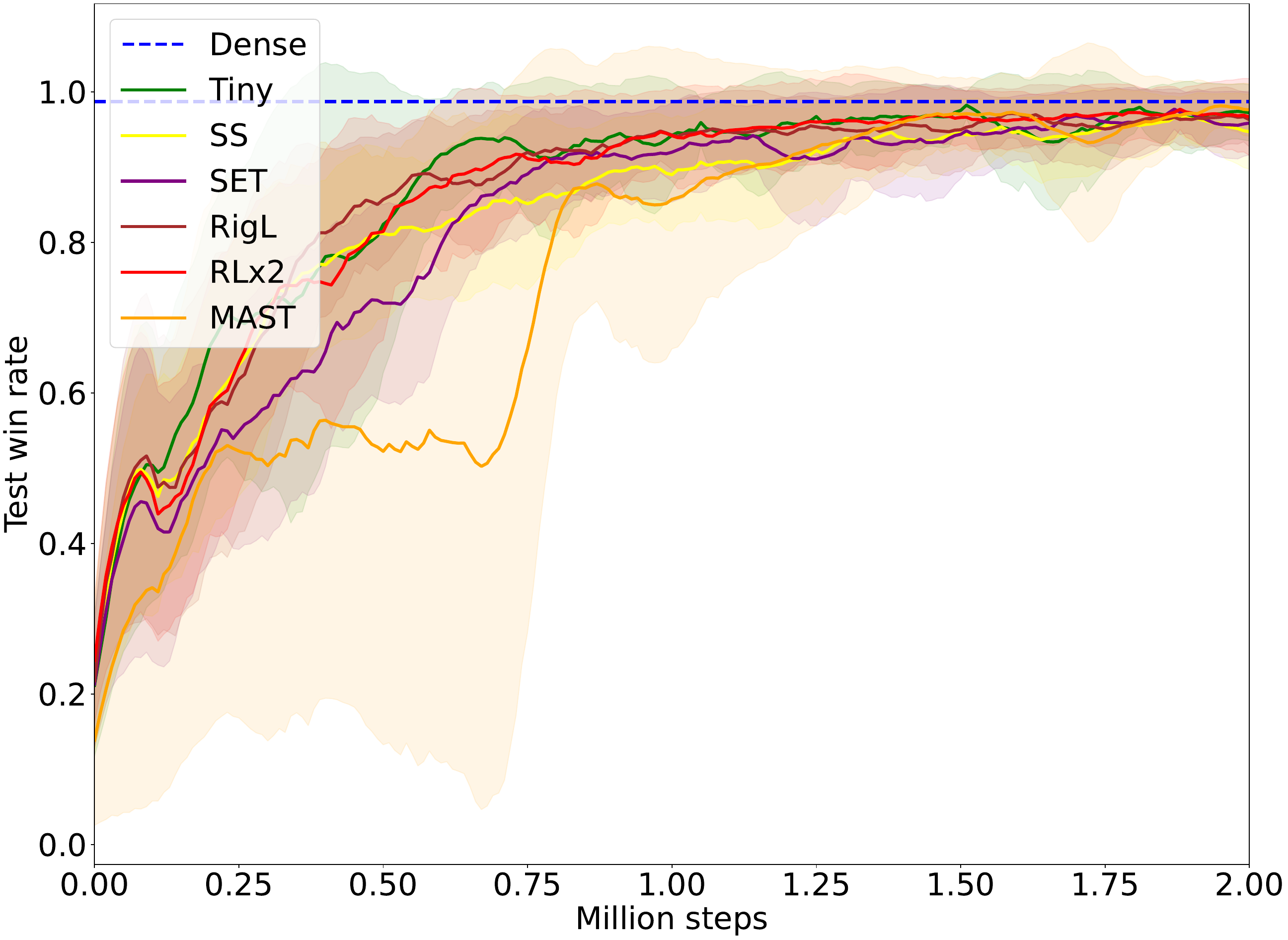}}
\subfigure[MAST-WQMIX $90\%$ on {\ttfamily 2s3z}]{\includegraphics[width=.38\linewidth]{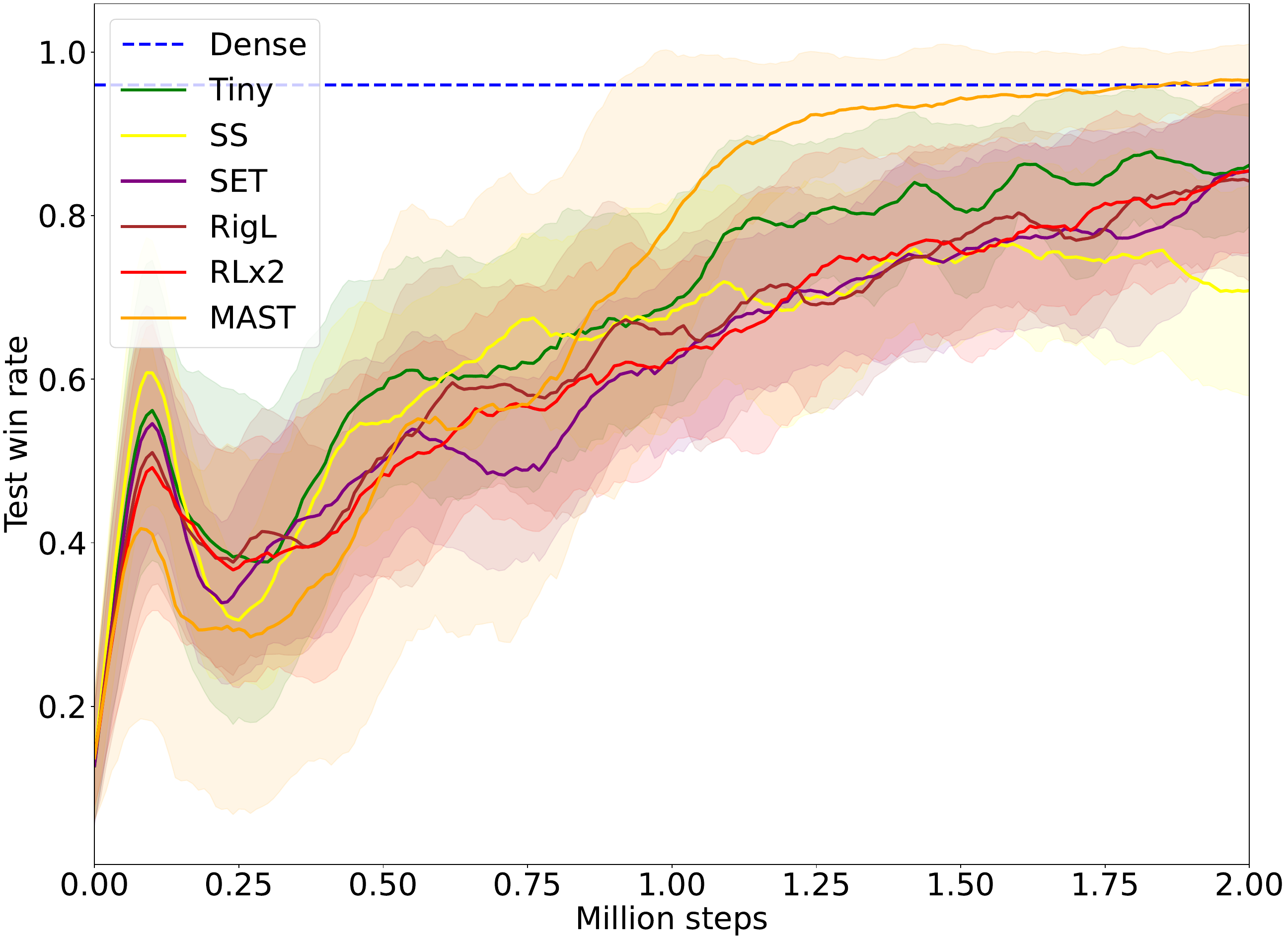}}
\subfigure[MAST-WQMIX $90\%$ on {\ttfamily 3s5z}]{\includegraphics[width=.38\linewidth]{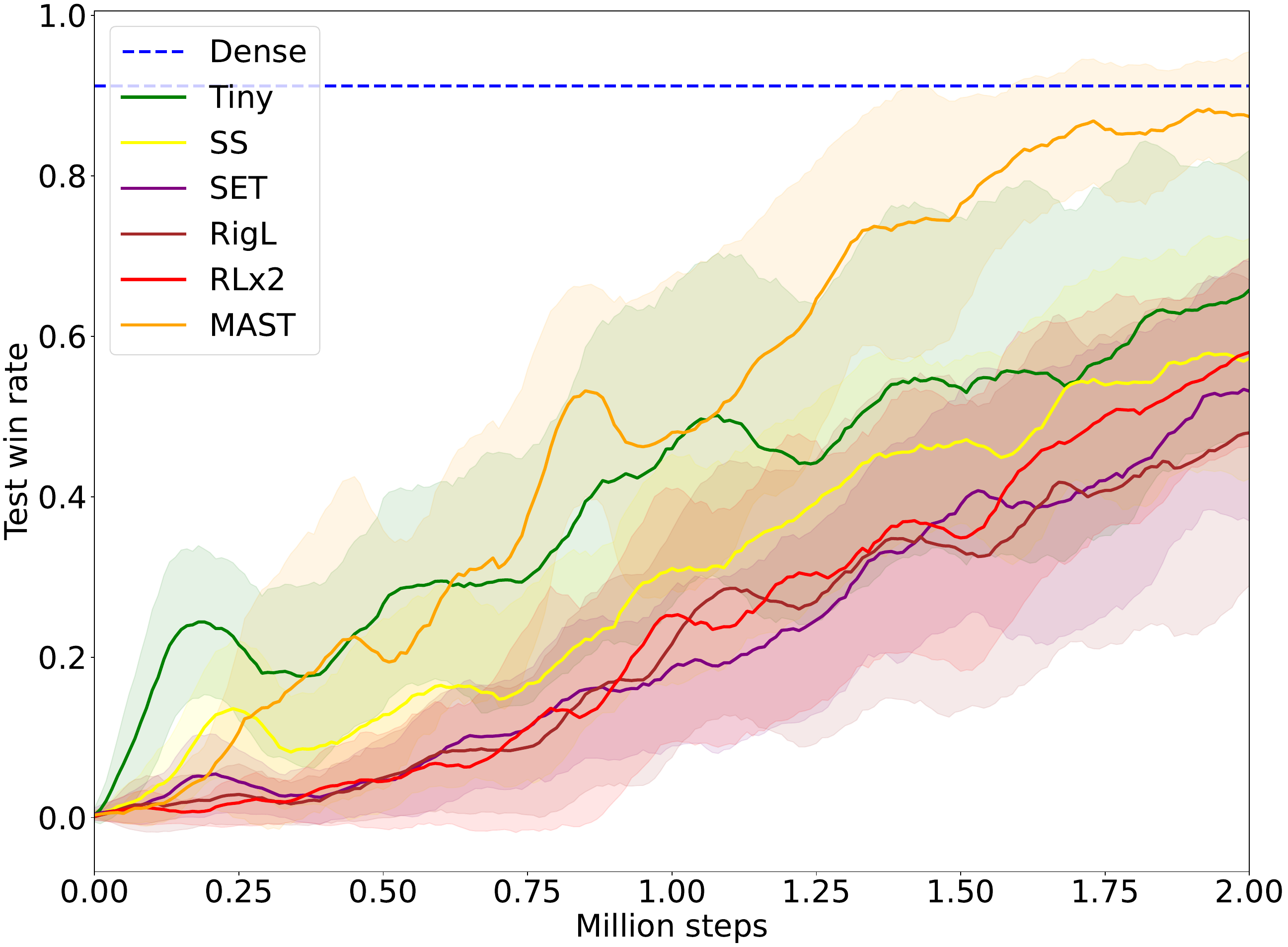}}
\subfigure[MAST-WQMIX $90\%$ on {\ttfamily 64zg}]{\includegraphics[width=.38\linewidth]{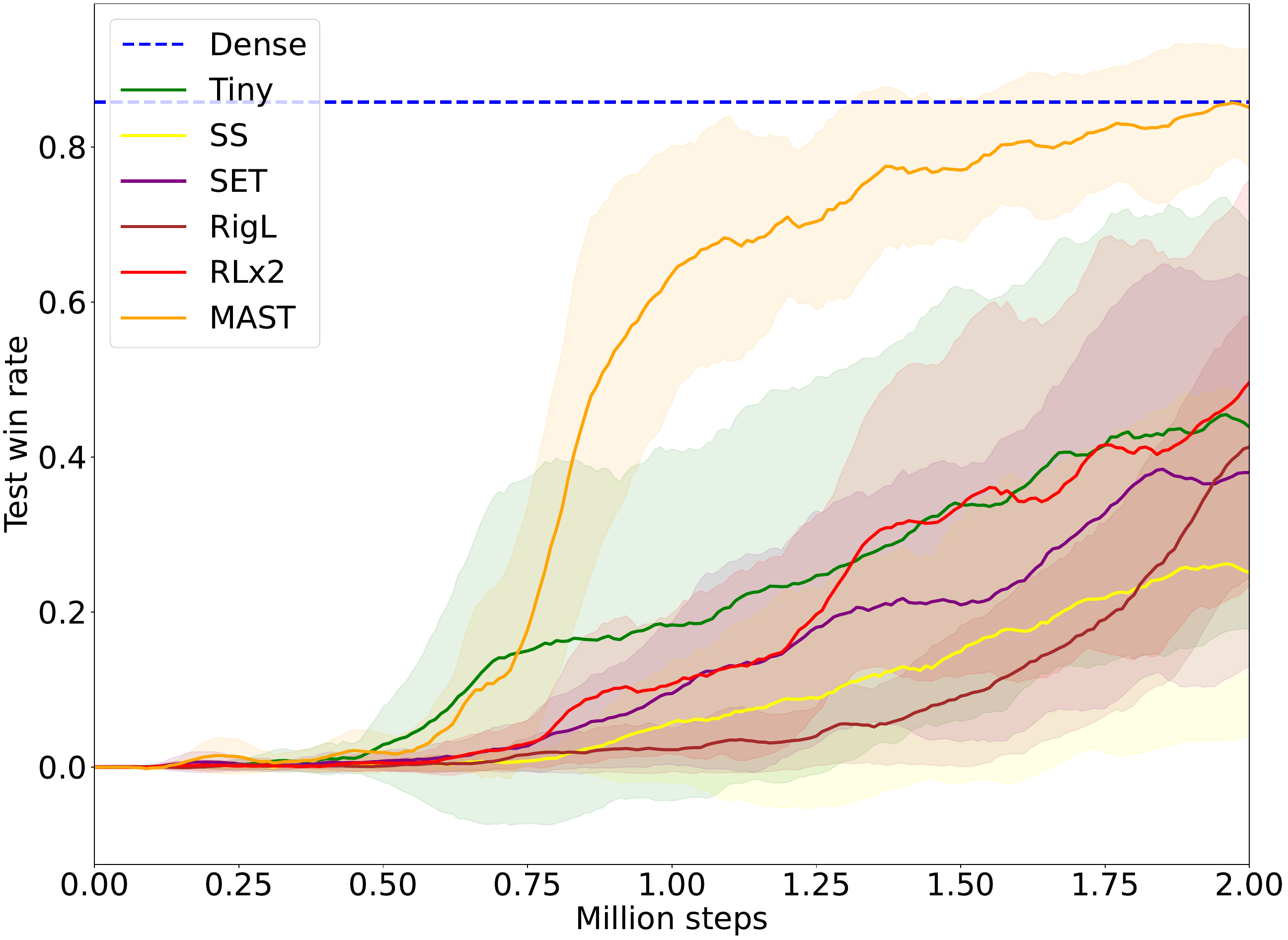}}
\vspace{-.4cm}
\caption{Training processes of MAST-WQMIX on four SAMC benchmarks.}
\label{fig:performance_wqmix}
\end{figure}

\begin{figure}[H]
\centering
\subfigure[MAST-RES $95\%$ on {\ttfamily 3m}]{\includegraphics[width=.38\linewidth]{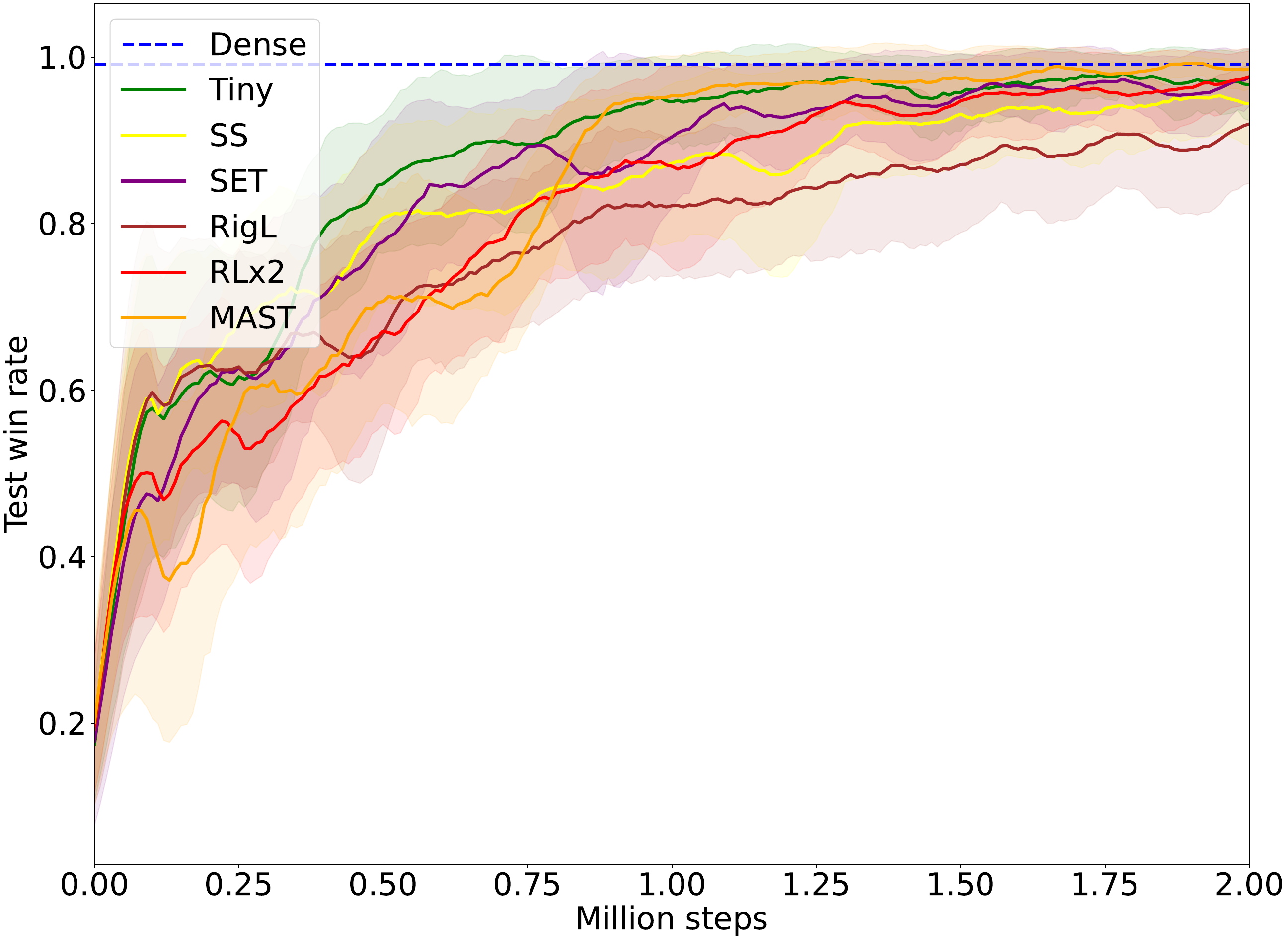}}
\subfigure[MAST-RES $90\%$ on {\ttfamily 2s3z}]{\includegraphics[width=.38\linewidth]{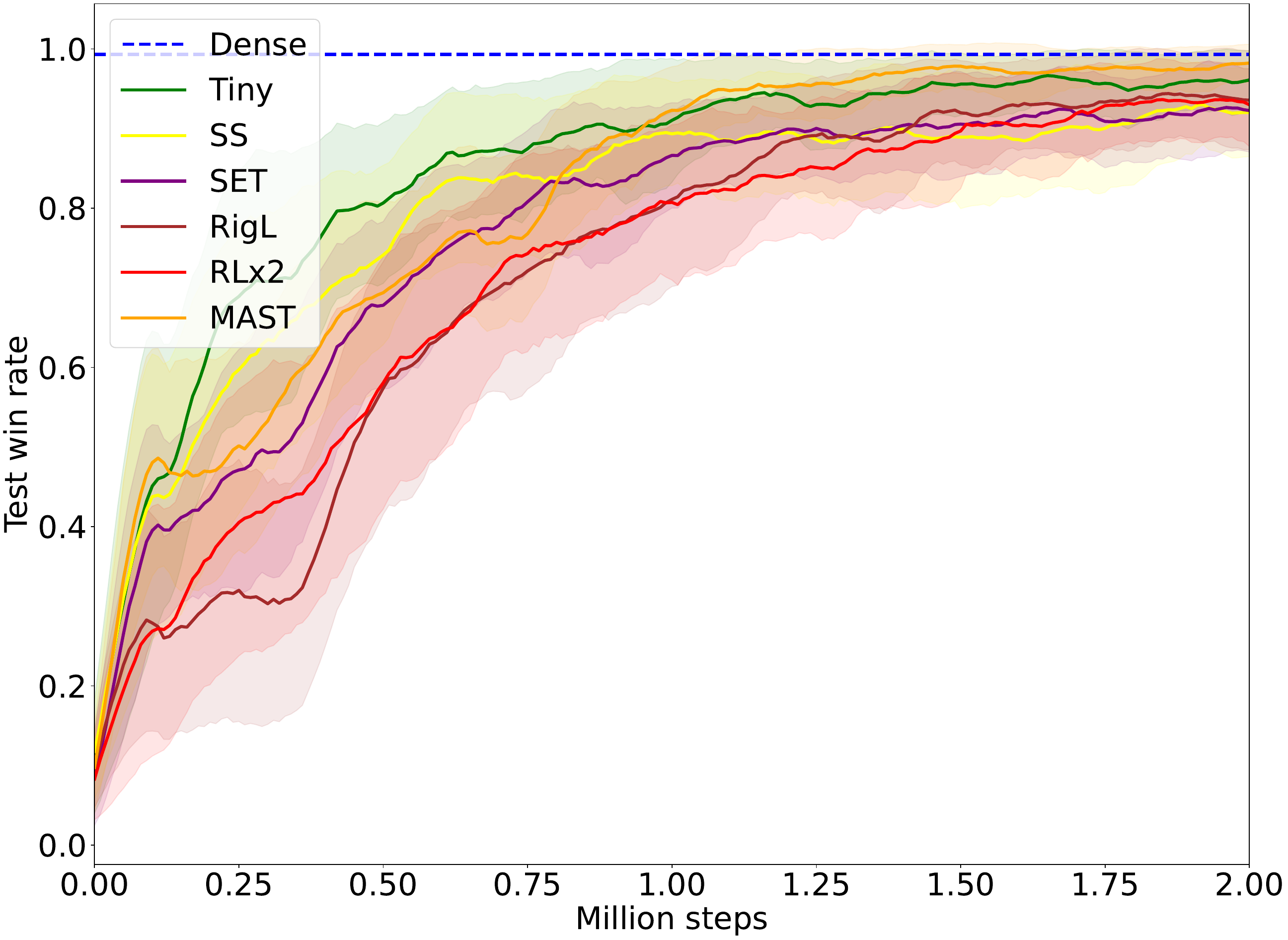}}
\subfigure[MAST-RES $85\%$ on {\ttfamily 3s5z}]{\includegraphics[width=.38\linewidth]{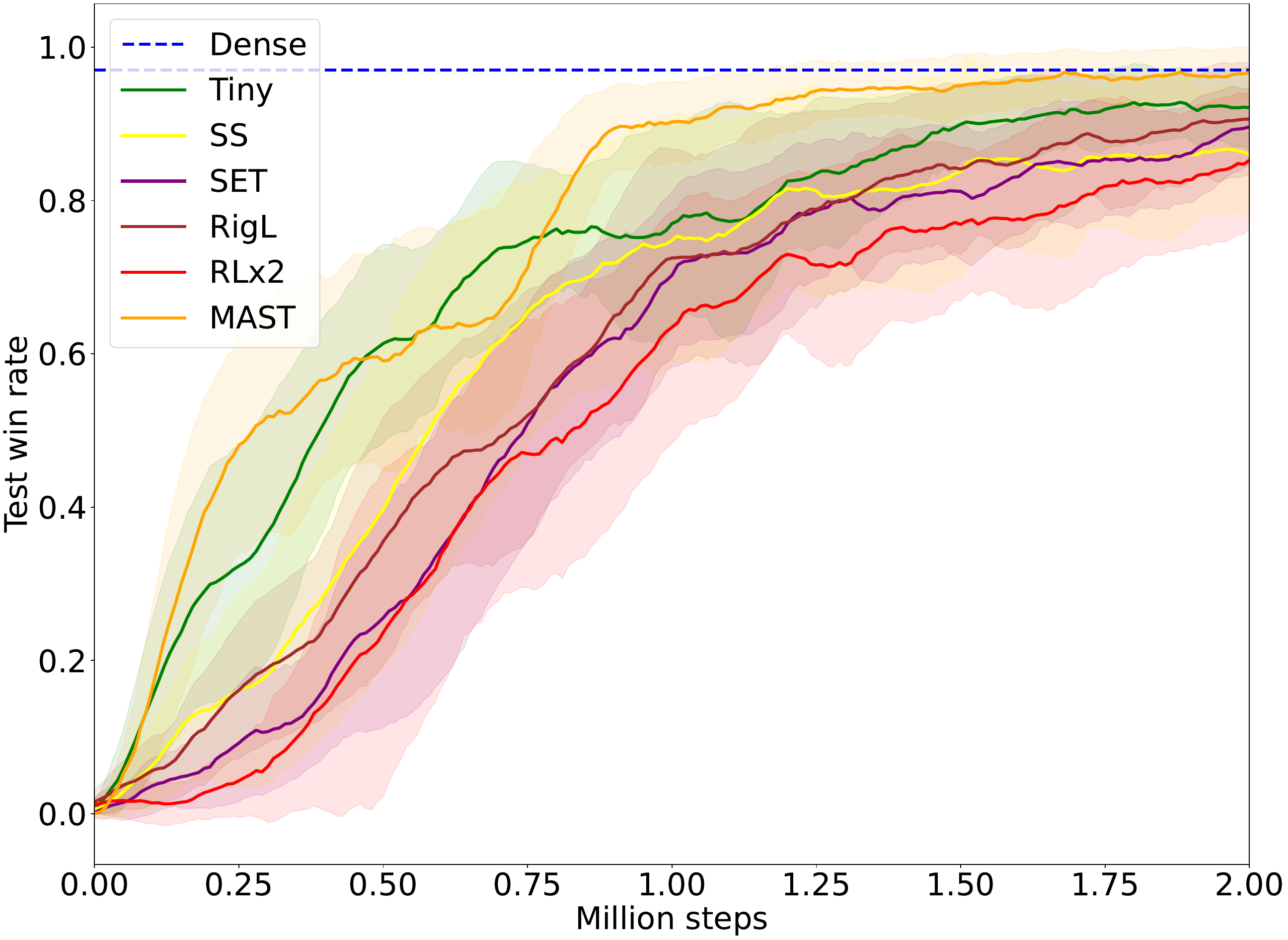}}
\subfigure[MAST-RES $85\%$ on {\ttfamily 64zg}]{\includegraphics[width=.38\linewidth]{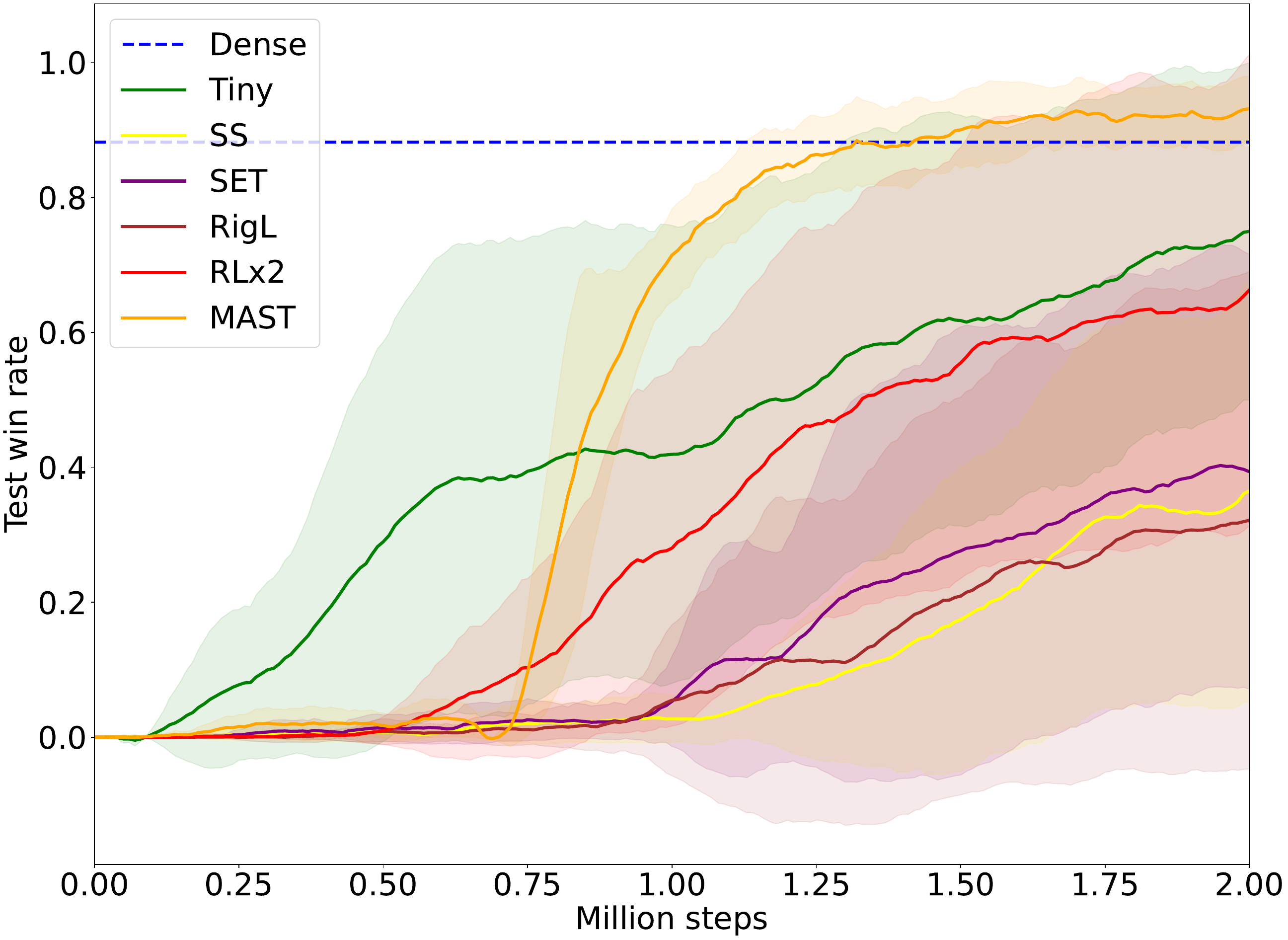}}
\caption{Training processes of MAST-RES on four SAMC benchmarks.}
\label{fig:performance_res}
\end{figure}

\vspace{-.2cm}
Figure~\ref{fig:performance_qmix}, Figure~\ref{fig:performance_wqmix}, and Figure~\ref{fig:performance_res} show the training curves of different algorithms in four SMAC environments. The performance is calculated as the average win rate per episode over the last 20 evaluations of the training. MAST outperforms baseline algorithms on all four environments in all three algorithms.
We smooth the training curve by a 1-D filter by {\ttfamily scipy.signal.savgol\_filter} in Python with {\ttfamily window\_length=21} and {\ttfamily polyorder=2}.

These figures unequivocally illustrate MAST's substantial performance superiority over all baseline methods in all four environments across the three algorithms. Notably, static sparse ({\ttfamily SS}) consistently exhibit the lowest performance on average, highlighting the difficulty of finding optimal sparse network topologies in the context of sparse MARL models. Dynamic sparse training methods, namely {\ttfamily SET} and {\ttfamily RigL}, slightly outperform ({\ttfamily SS}), although their performance remains unsatisfactory. Sparse networks also, on average, underperform tiny dense networks. However, MAST significantly outpaces all other baselines, indicating the successful realization of accurate value estimation through our MAST method, which effectively guides gradient-based topology evolution. Notably, the single-agent method {\ttfamily RLx2} consistently delivers subpar results in all experiments, potentially due to the sensitivity of the step length in the multi-step targets, and the failure of dynamic buffer for episode-form training samples.

\vspace{-.6cm}
\vspace{-1.5cm}
\vspace{-.4cm}
\vspace{-.4cm}
\subsection{Standard Deviations of Results in Table~\ref{tb:comp_eval_2}}\label{app:std}
\vspace{-.2cm}
Table~\ref{tb:comp_eval_std} showcases algorithm performance across four SMAC environments along with their corresponding standard deviations. It's important to note that the data in Table~\ref{tb:comp_eval_std} is not normalized concerning the dense model. Notably, MAST's utilization of topology evolution doesn't yield increased variance in results, demonstrating consistent performance across multiple random seeds.
\vspace{-1cm}
\begin{table}[H]
\caption{Results in Table~\ref{tb:comp_eval_2} with standard deviations
}
\label{tb:comp_eval_std}
\centering
\begin{tabular}{l|l|cccccc}
\toprule
Alg. & \multicolumn{1}{c|}{Env.} & \makecell[c]{Tiny(\%)} & \makecell[c]{SS(\%)} & \makecell[c]{SET(\%)} & \makecell[c]{RigL(\%)} &   \makecell[c]{RLx2(\%)} & \makecell[c]{Ours(\%)}\\
\midrule
\multirow{5}{*}{\makecell[c]{Q-\\MIX}} & 3m &  \makecell[c]{96.3$\pm$4.3} & \makecell[c]{89.8$\pm$7.9} & \makecell[c]{94.1$\pm$5.9} & \makecell[c]{93.4$\pm$9.5} & \makecell[c]{11.9$\pm$20.5} & \makecell[c]{\textbf{98.9}$\pm$2.0}\\
& 2s3z &  \makecell[c]{80.8$\pm$12.9} & \makecell[c]{70.4$\pm$13.1} & \makecell[c]{74.9$\pm$16.4} & \makecell[c]{67.0$\pm$14.0} & \makecell[c]{44.2$\pm$17.0}& \makecell[c]{\textbf{94.6}$\pm$4.6} \\
& 3s5z & \makecell[c]{64.2$\pm$11.8} & \makecell[c]{32.0$\pm$20.3} & \makecell[c]{49.3$\pm$16.6} & \makecell[c]{42.6$\pm$19.2} & \makecell[c]{47.2$\pm$16.2}& \makecell[c]{\textbf{93.3}$\pm$5.1}\\
& 64* & \makecell[c]{54.0$\pm$29.9} & \makecell[c]{37.3$\pm$23.4} & \makecell[c]{62.3$\pm$21.9} & \makecell[c]{45.2$\pm$23.4} & \makecell[c]{9.2$\pm$15.0}& \makecell[c]{\textbf{90.6}$\pm$7.8}\\
\cline{2-8}
& Avg. & \makecell[c]{73.8$\pm$14.7} & \makecell[c]{57.4$\pm$16.2} & \makecell[c]{70.1$\pm$15.2} & \makecell[c]{62.0$\pm$16.5} & \makecell[c]{28.1$\pm$17.2}& \makecell[c]{\textbf{94.3}$\pm$4.9}\\
\midrule
\multirow{5}{*}{\makecell[c]{WQ-\\MIX}} & 3m &  \makecell[c]{97.0$\pm$4.0}  & \makecell[c]{95.6$\pm$4.0} &  \makecell[c]{96.5$\pm$3.6} & \makecell[c]{96.5$\pm$3.6} & \makecell[c]{96.7$\pm$4.3} & \makecell[c]{\textbf{97.3}$\pm$4.0}\\
& 2s3z & \makecell[c]{86.0$\pm$7.9} & \makecell[c]{72.4$\pm$12.4} & \makecell[c]{82.5$\pm$10.9} & \makecell[c]{83.3$\pm$10.3} & \makecell[c]{83.8$\pm$9.9} & \makecell[c]{\textbf{96.2}$\pm$4.2}\\
& 3s5z & \makecell[c]{64.5$\pm$17.9} & \makecell[c]{57.0$\pm$14.5} & \makecell[c]{51.1$\pm$15.0} & \makecell[c]{46.0$\pm$20.5} & \makecell[c]{55.4$\pm$11.3} & \makecell[c]{\textbf{87.6}$\pm$6.9}\\
& 64* & \makecell[c]{43.8$\pm$27.4} & \makecell[c]{25.4$\pm$22.0} & \makecell[c]{37.8$\pm$26.2} & \makecell[c]{35.2$\pm$16.7} & \makecell[c]{45.3$\pm$24.7} & \makecell[c]{\textbf{84.4}$\pm$8.4}\\
\cline{2-8}
& Avg.& \makecell[c]{68.5$\pm$13.5} & \makecell[c]{62.2$\pm$13.0} & \makecell[c]{64.0$\pm$13.5} & \makecell[c]{65.8$\pm$11.4} & \makecell[c]{70.3$\pm$12.5}& \makecell[c]{\textbf{91.4}$\pm$5.9}\\
\midrule
\multirow{5}{*}{\makecell[c]{RES}} & 3m &  \makecell[c]{96.9$\pm$4.1}  & \makecell[c]{94.7$\pm$4.8} &  \makecell[c]{96.4$\pm$4.3} & \makecell[c]{90.3$\pm$7.4} & \makecell[c]{97.0$\pm$3.8}& \makecell[c]{\textbf{102.2}$\pm$3.2}\\
& 2s3z & \makecell[c]{95.8$\pm$3.8} & \makecell[c]{92.2$\pm$5.9} & \makecell[c]{92.2$\pm$5.5} & \makecell[c]{94.0$\pm$5.7} &\makecell[c]{93.4$\pm$5.5} & \makecell[c]{\textbf{97.7}$\pm$2.6}\\
& 3s5z & \makecell[c]{92.2$\pm$4.8} & \makecell[c]{86.3$\pm$8.8} & \makecell[c]{87.6$\pm$5.9} & \makecell[c]{90.0$\pm$7.3} &\makecell[c]{83.6$\pm$9.2} & \makecell[c]{\textbf{96.4}$\pm$3.4}\\
& 64* & \makecell[c]{73.5$\pm$25.8} & \makecell[c]{34.5$\pm$29.6} & \makecell[c]{38.9$\pm$32.3} & \makecell[c]{31.1$\pm$36.2} &\makecell[c]{64.1$\pm$33.8} & \makecell[c]{\textbf{92.5}$\pm$4.9}\\
\cline{2-8}
& Avg.& \makecell[c]{89.6$\pm$9.6} & \makecell[c]{76.9$\pm$12.3} & \makecell[c]{78.8$\pm$12.0} & \makecell[c]{76.3$\pm$14.1} &\makecell[c]{84.5$\pm$13.1} & \makecell[c]{\textbf{97.2}$\pm$3.5}\\
\midrule
\multicolumn{2}{c|}{Avg.} & \makecell[c]{78.7$\pm$12.8} & \makecell[c]{65.6$\pm$13.9} & \makecell[c]{72.0$\pm$13.7} & \makecell[c]{67.8$\pm$14.5} & \makecell[c]{61.0$\pm$14.3}& \makecell[c]{\textbf{94.2}$\pm$4.6} \\
\bottomrule
\end{tabular}
\end{table}
\subsection{Ablation Study}\label{app:ablation}
We conduct a comprehensive ablation study on three critical elements of MAST: hybrid TD($\lambda$) targets, the Soft Mellowmax operator, and dual buffers, specifically evaluating their effects on QMIX and WQMIX. Notably, since MAST-QMIX shares similarities with MAST-RES, our experiments focus on QMIX and WQMIX within the {\ttfamily 3s5z} task. This meticulous analysis seeks to elucidate the influence of each component on MAST and their robustness in the face of hyperparameter variations. The reported results are expressed as percentages and are normalized to dense models.

\paragraph{Hybrid TD($\lambda$)}
We commence our analysis by evaluating different burn-in time $T_0$ in hybrid TD($\lambda$) in Table~\ref{tb:ablation_td_1}. Additionally, we explore the impact of different $\lambda$ values within hybrid TD($\lambda$) in Table~\ref{tb:ablation_td_2}. These results reveal hybrid TD($\lambda$) targets achieve optimal performance with a burn-in time of $T_0=0.75\text{M}$ and $\lambda=0.6$. It is noteworthy that hybrid TD($\lambda$) targets lead to significant performance improvements in WQMIX, while their impact on QMIX is relatively modest.
\begin{table}[H]
    \caption{Ablation study on burn-in time $T_0$ in hybrid TD($\lambda$).}
    \label{tb:ablation_td_1}
    \centering
    \begin{tabular}{l|cccccc}
    \toprule
    Alg. & $0$ & $0.45$M & $0.75\text{M}$ & $1$M & $1.5\text{M}$ & $2\text{M}$ \\
    \midrule
    QMIX / RES &  93.6 & 94.3 & \textbf{97.9} & 92.0 & 92.5 & 91.5 \\
    WQMIX & 83.5 & 88.4 & 98.0 & \textbf{98.7} & 76.9 & 70.3   \\
    \midrule
    Avg. & 88.5 & 91.3 & \textbf{97.9} & 95.4 & 84.7 & 80.9   \\
    \bottomrule
    \end{tabular}
\end{table}

\begin{table}[H]
    \caption{Ablation study on $\lambda$ in hybrid TD($\lambda$).}
    \label{tb:ablation_td_2}
    \centering
    \begin{tabular}{l|cccccc}
    \toprule
    Alg. & $0$ & $0.2$ & $0.4$ & $0.6$ & $0.8$ & $1$ \\
    \midrule
    QMIX / RES & 91.5 & 94.7 & 96.8 & 96.8 & \textbf{97.9} & 89.4 \\
    WQMIX & 83.5 & 83.5 & 74.7 & \textbf{98.0} & 96.1 & 87.9  \\
    \midrule
    Avg. & 87.5 & 89.1 & 85.7 & \textbf{97.4} & 97.0 & 88.6  \\
    \bottomrule
    \end{tabular}
\end{table}

\begin{wrapfigure}{r}{.35\linewidth}
\centering
\includegraphics[width=\linewidth]{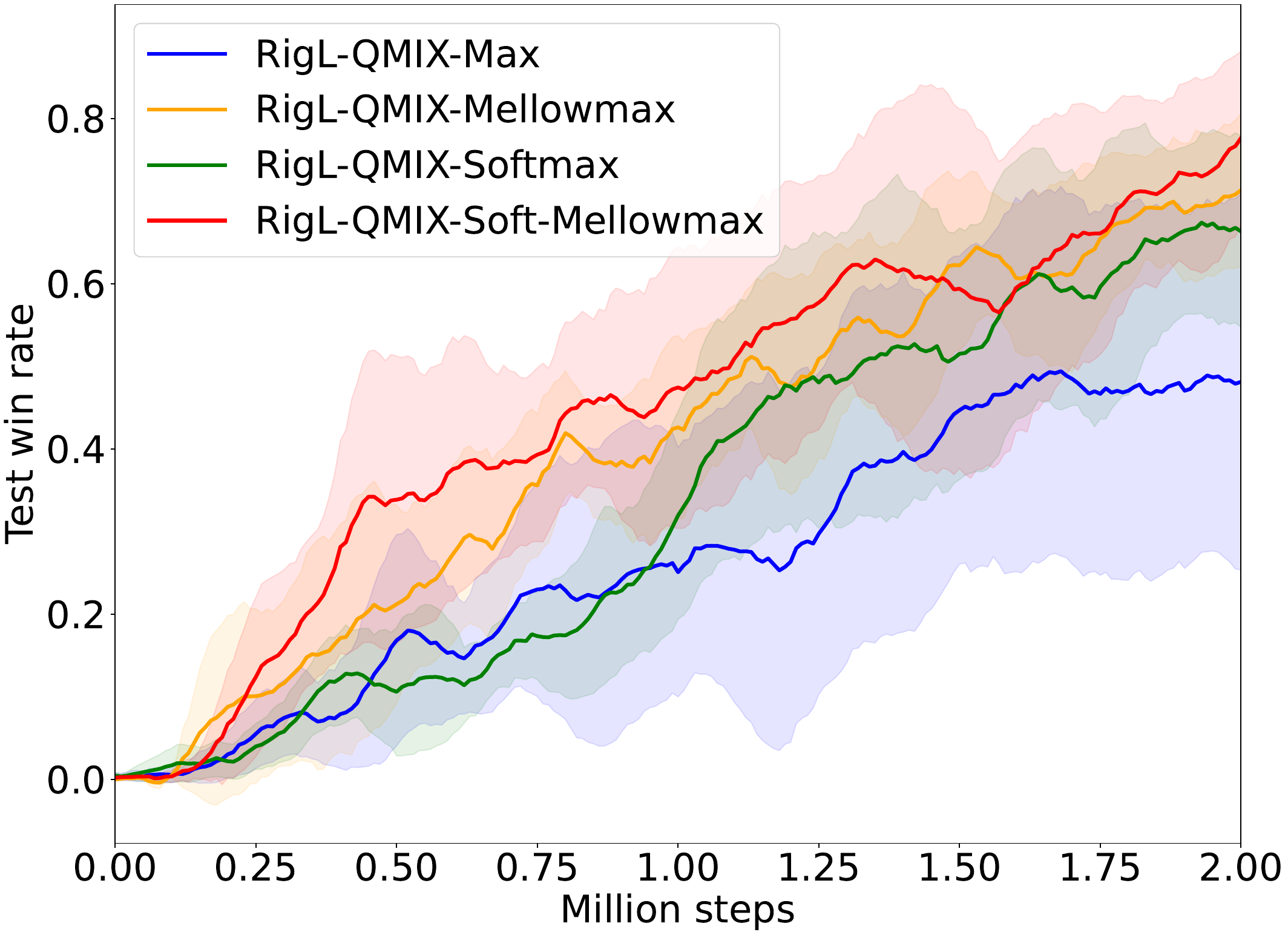}
\vspace{-.7cm}
\caption{Comparison of different operators.}
\vspace{-1cm}
\label{fig:op}
\end{wrapfigure}
\paragraph{Soft Mellowmax Operator} The Soft Mellowmax operator in Eq.(\ref{eq:sm}) introduces two hyperparameters, $\alpha$ and $\omega$. A comprehensive examination of various parameter configurations is presented in Table~\ref{tb:ablation_sm}, showing that the performance of MAST exhibits robustness to changes in the two hyperparameters associated with the Soft Mellowmax operator.

Additionally, it is worth noting that the Softmax operator is also employed in \cite{pan2021regularized} to mitigate overestimation in multi-agent Q learning. To examine the effectiveness of various operators, including max, Softmax, Mellowmax, and Soft Mellowmax, we conduct a comparative analysis in Figure~\ref{fig:op}. Our findings indicate that the Soft Mellowmax operator surpasses all other baselines in alleviating overestimation. Although the Softmax operator demonstrates similar performance to the Soft Mellowmax operator, it should be noted that the Softmax operator entails higher computational costs.
\begin{table}[H]
    \caption{Ablation study on the Soft Mellowmax operator.}
    \label{tb:ablation_sm}
    \centering
    \begin{tabular}{l|ccccc}
    \toprule
    Alg. & \makecell[c]{$\alpha=1$\\$\omega=10$} & \makecell[c]{$\alpha=5$\\$\omega=5$} & \makecell[c]{$\alpha=5$\\$\omega=10$} & \makecell[c]{$\alpha=10$\\$\omega=5$} & \makecell[c]{$\alpha=10$\\$\omega=10$} \\
    \midrule
    QMIX / RES & 97.9 & \textbf{100.0} & 98.9 & 96.8 & 97.9 \\
    WQMIX & \textbf{98.0} & 92.3 & 87.9 & 92.3 & 85.7 \\
    \midrule
    Avg. & \textbf{97.9} & 96.1 & 93.4 & 94.5 & 91.8 \\
    \bottomrule
    \end{tabular}
\end{table}
\paragraph{Dual Buffers} In each training step, we concurrently sample two batches from the two buffers, $\mathcal{B}_1$ and $\mathcal{B}_2$. We maintain a fixed total batch size of $32$ while varying the sample partitions $b_1:b_2$ within MAST. The results, detailed in Table~\ref{tb:ablation_db}, reveal that employing two buffers with a partition ratio of $6:2$ yields the best performance. Additionally, we observed a significant degradation in MAST's performance when using data solely from a single buffer, whether it be the online or offline buffer. This underscores the vital role of dual buffers in sparse MARL.
\begin{table}[H]
    \caption{Ablation study on dual buffers.}
    \label{tb:ablation_db}
    \centering
    \begin{tabular}{l|cccccc}
    \toprule
    Alg. & $8:0$ & $6:2$ & $5:3$ & $3:5$ & $2:6$ & $0:8$ \\
    \midrule
    QMIX / RES & 93.6 & \textbf{99.4} & 97.9 & 97.8 & 89.6 & 85.1 \\
    WQMIX & 64.8 & \textbf{101.2} & 98.0 & 86.8 & 95.1 & 70.3 \\
    \midrule
    Avg. & 79.2 & \textbf{100.3} & 97.9 & 92.3 & 92.4 & 77.7 \\
    \bottomrule
    \end{tabular}
\end{table}

\subsection{Sensitivity Analysis for Hyperparameters}\label{app:sen}
\vspace{-.2cm}
Table~\ref{tb:sensitivity-mask-update-interval} shows the performance with different mask update intervals (denoted as $\Delta_m$) in different environments, which reveals several key observations:
\begin{itemize}
    \item Findings indicate that a small $\Delta_m$ negatively impacts performance, as frequent mask adjustments may prematurely drop critical connections before their weights are adequately updated by the optimizer.
    \item Overall, A moderate $\Delta_m=200$ episodes performs well in different algorithms.
\end{itemize}
\begin{table}[H]
\centering
\caption{Sensitivity analysis on mask update interval.}
\label{tb:sensitivity-mask-update-interval}
\begin{tabular}{lccccccccc}
\toprule
Alg. & \makecell[c]{$\Delta_m=20$\\episodes} & \makecell[c]{$\Delta_m=100$\\episodes} & \makecell[c]{$\Delta_m=200$\\episodes} & \makecell[c]{$\Delta_m=1000$\\episodes} & \makecell[c]{$\Delta_m=2000$\\episodes} \\
\midrule
QMIX/RES &  99.4\% & 97.7\% & 99.0\% & \textbf{100.6}\% & \textbf{100.6}\%\\
WQMIX &  83.2\% & 91.9\% & \textbf{96.1}\% & 68.1\% & 71.5\%\\
\midrule
Average & 91.3\% & 94.8\% & \textbf{97.5}\% & 84.3\% & 86.0\% \\
\bottomrule
\end{tabular}
\end{table}
\vspace{-.6cm}

\subsection{Experiments on QMIX in Multi-Agent MuJoCo}\label{app:comix}
\begin{figure}[H]
\vspace{-.4cm}
\centering
\subfigure[2-Agent Humanoid]{\includegraphics[width=.3\linewidth]{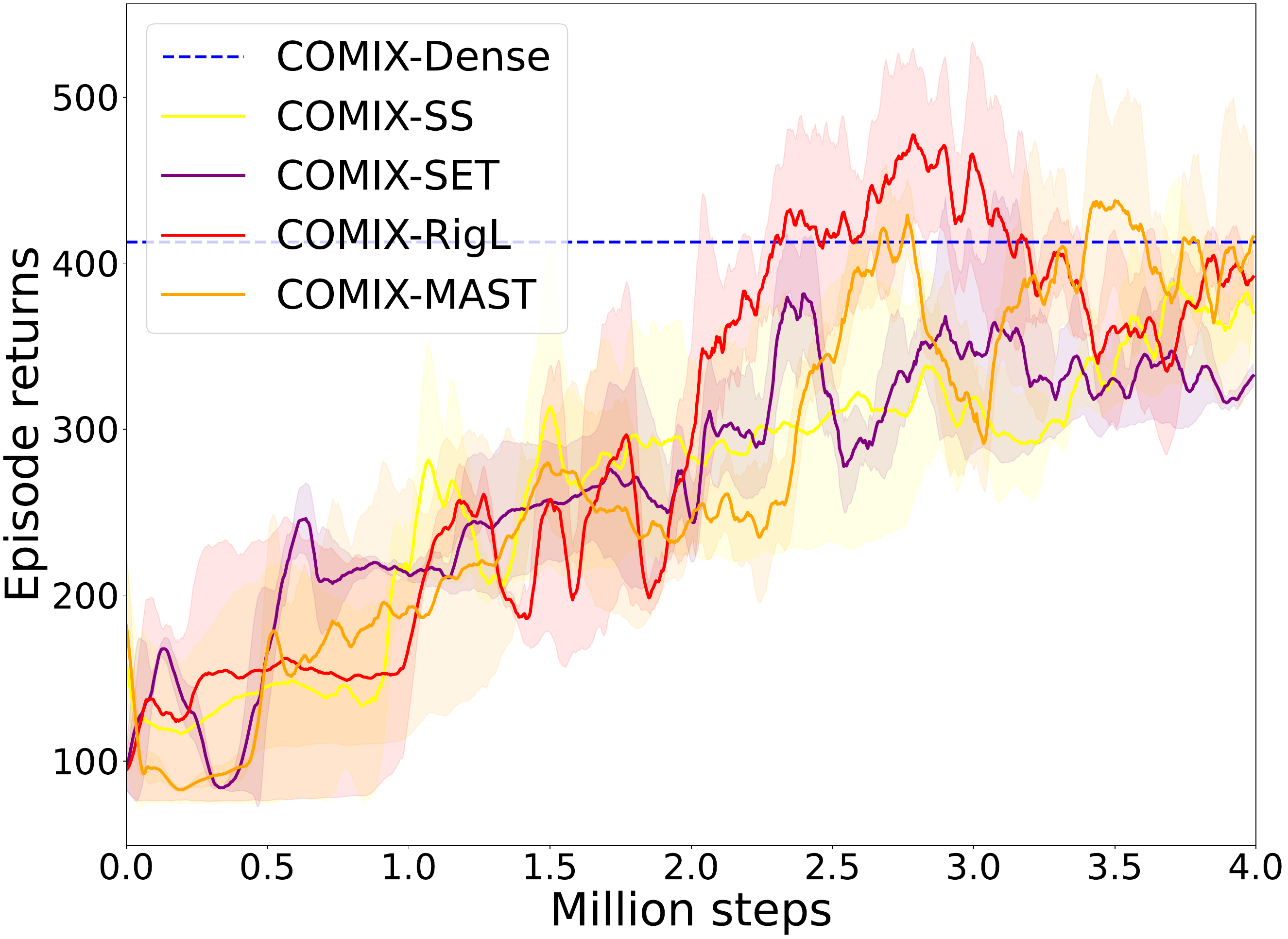}}
\subfigure[2-Agent HumanoidStandup]{\includegraphics[width=.315\linewidth]{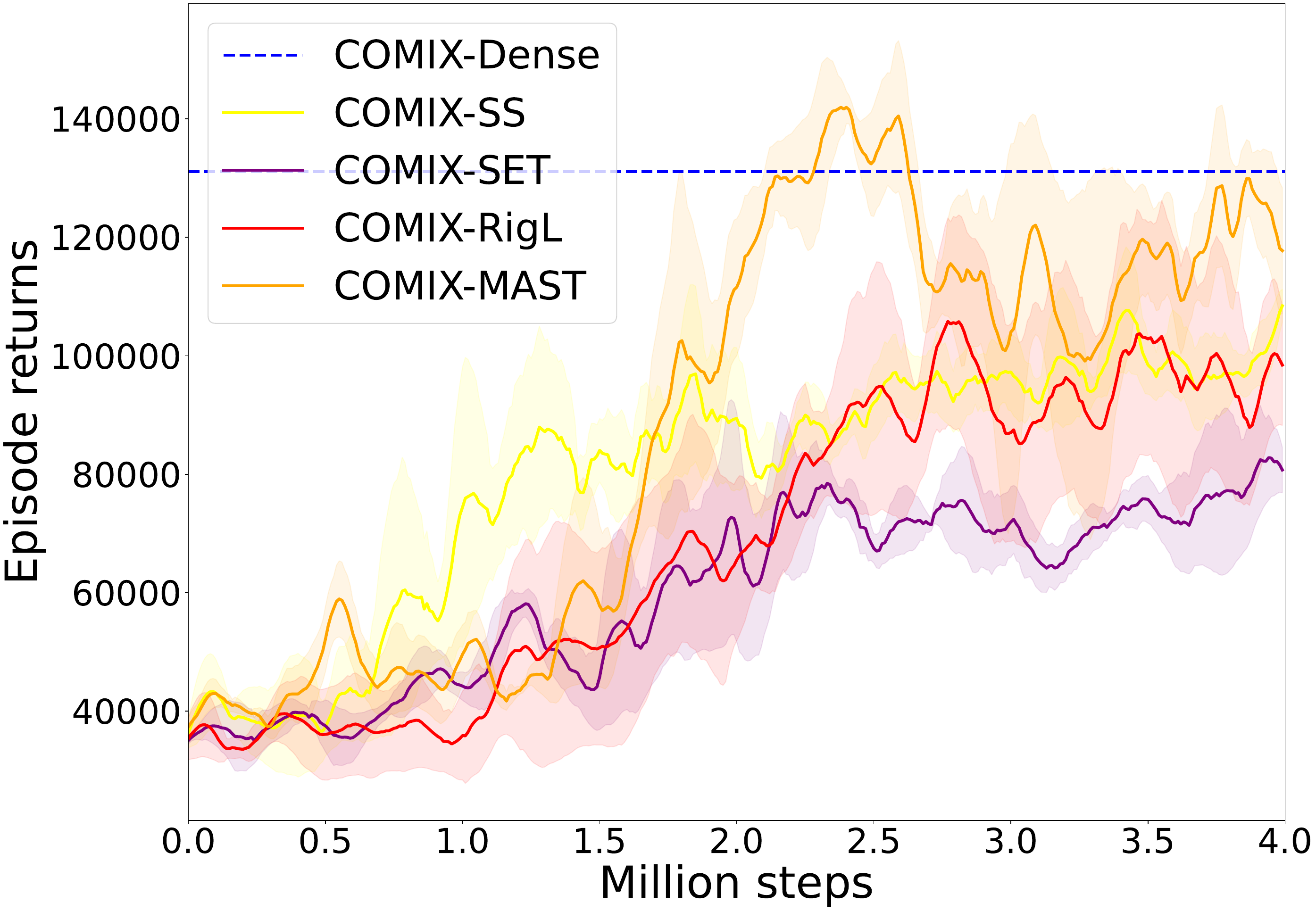}}
\subfigure[ManyAgent Swimmer]{\includegraphics[width=.3\linewidth]{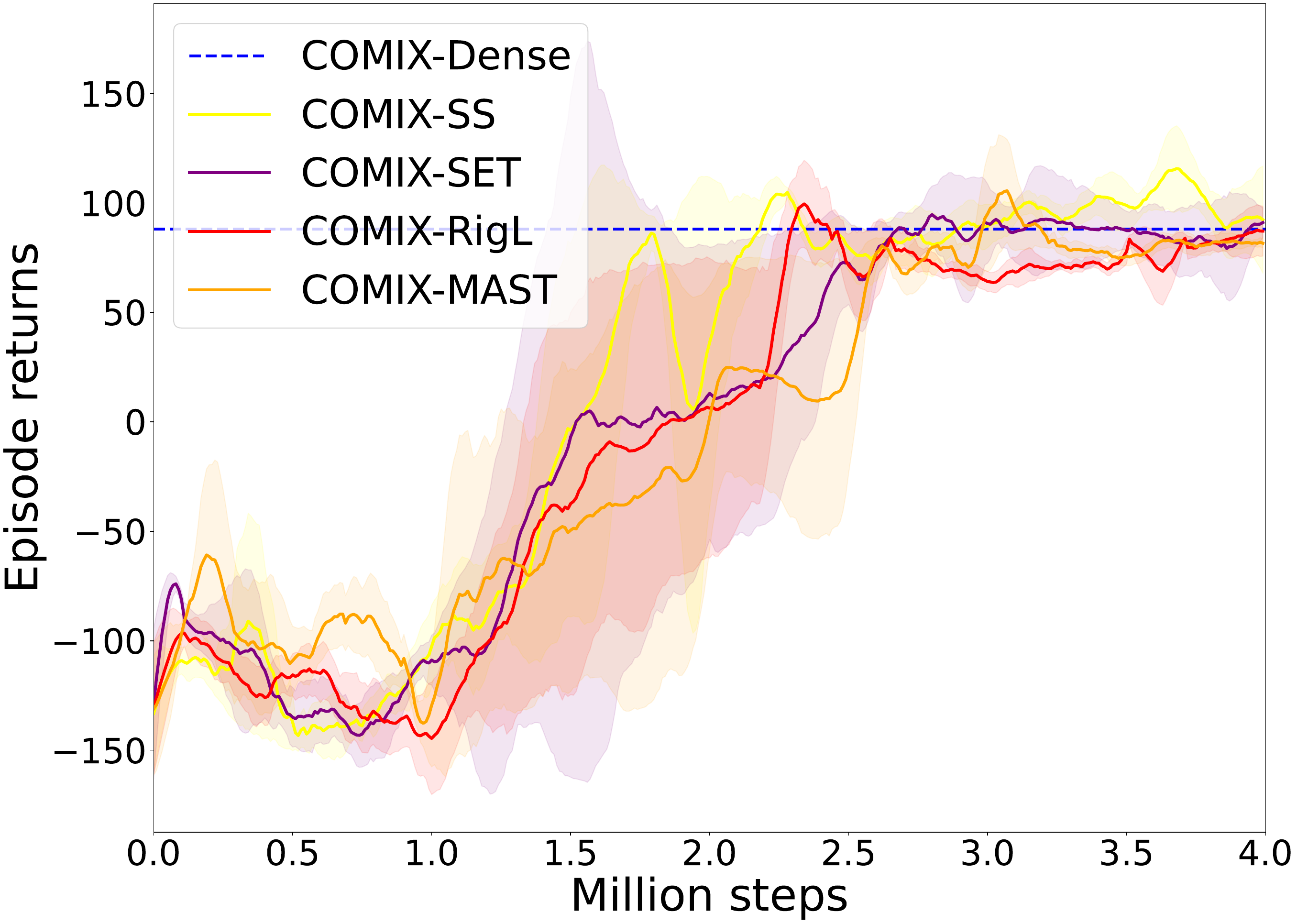}}
\vspace{-.3cm}
\caption{Mean episode return on different MAMuJoCo tasks with MAST-QMIX.}
\label{fig:mamujoco}
\vspace{-.6cm}
\end{figure}

We compare MAST-QMIX with other baselines on the MAMuJoCo benchmark, and the results are presented in Figure~\ref{fig:mamujoco}. From the figure, we observe that MAST consistently achieves dense performance and outperforms other methods in three environments, except for ManyAgent Swimmer, where the static sparse network performs similarly to MAST. This similarity may be attributed to the simplicity of the ManyAgent Swimmer environment, which allows a random static sparse network to achieve good performance. This comparison demonstrates the applicability of MAST across different environments.

\subsection{Experiments on FACMAC in SMAC}\label{app:facmac}
In addition to pure value-based deep MARL algorithms, we also evaluate MAST with a hybrid value-based and policy-based algorithm, FACMAC \cite{peng2021facmac}, in SMAC. The results are presented in Figure~\ref{fig:facmac}. From the figure, we observe that MAST consistently achieves dense performance and outperforms other methods in three environments, demonstrating its applicability across different algorithms.
\begin{figure}[H]
\vspace{-.4cm}
\centering
\subfigure[MMM]{\includegraphics[width=.3\linewidth]{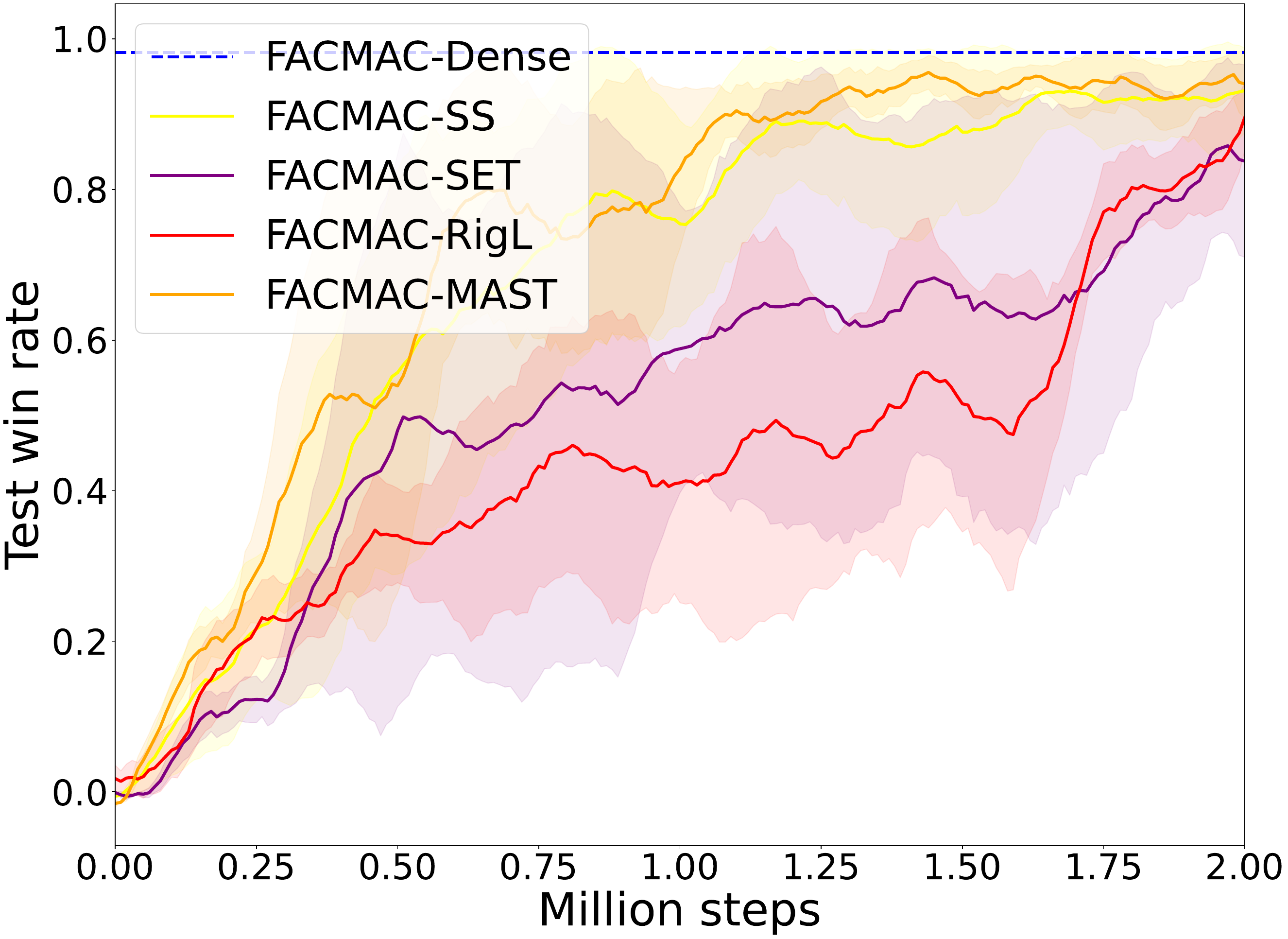}}
\subfigure[2c\_vs\_64zg]{\includegraphics[width=.3\linewidth]{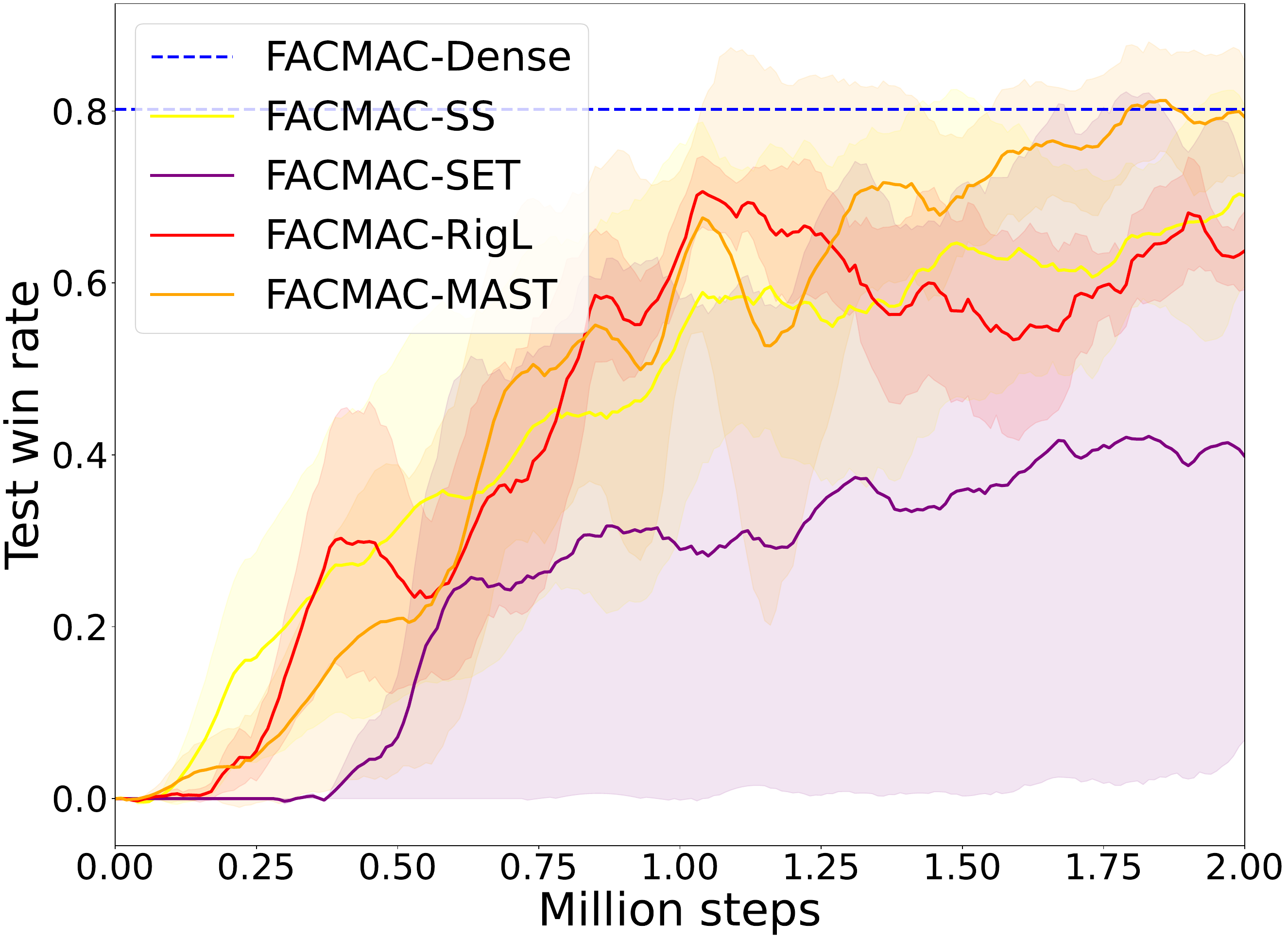}}
\subfigure[bane\_vs\_bane]{\includegraphics[width=.3\linewidth]{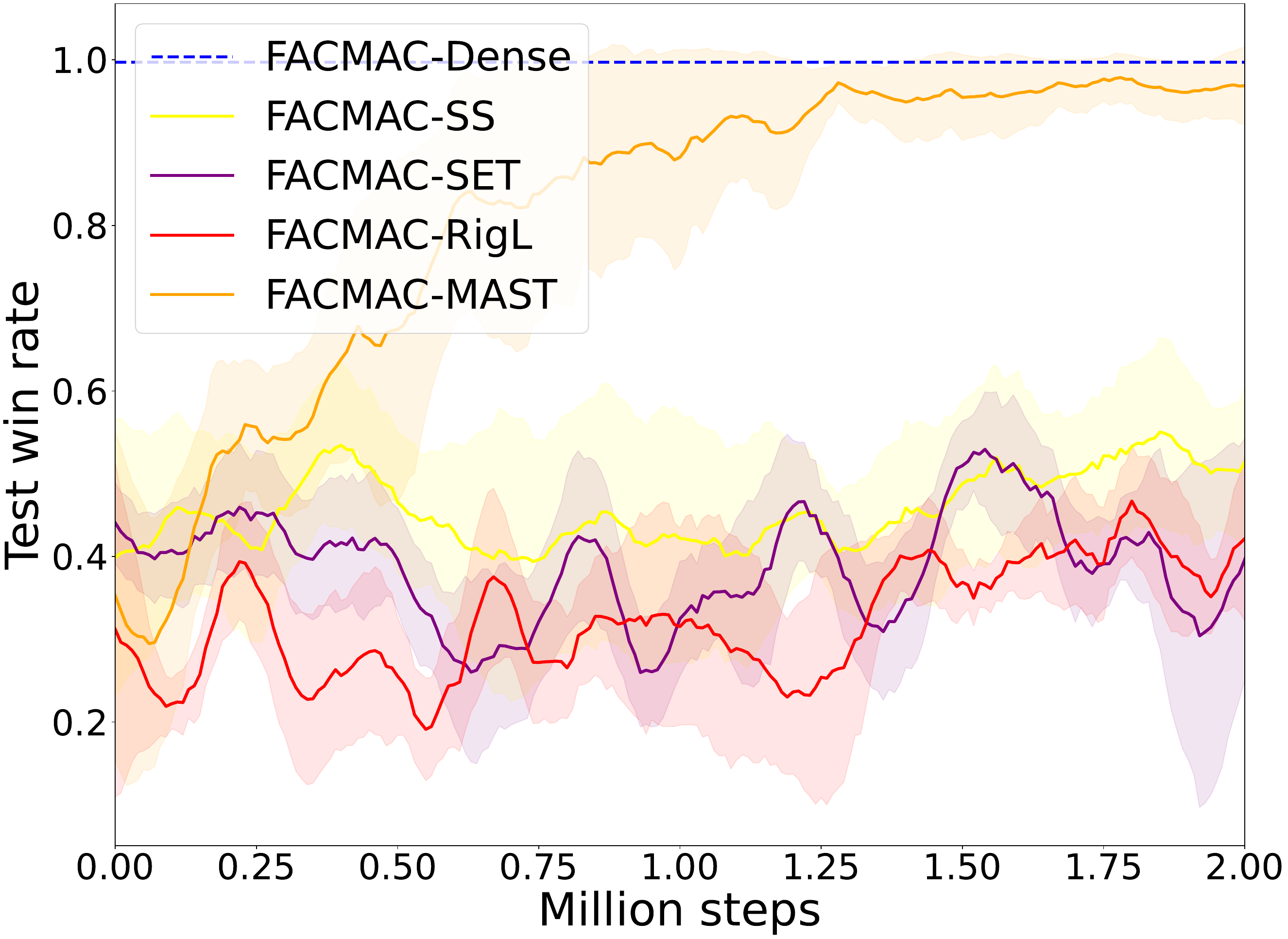}}
\vspace{-.3cm}
\caption{Mean episode win rates on different SMAC tasks with MAST-FACMAC.}
\label{fig:facmac}
\vspace{-.6cm}
\end{figure}

\subsection{Visualization of Sparse Masks}\label{app:vs}
\begin{figure}[H]
\centering
\subfigure[$T=0$M (Input)]{\includegraphics[width=.24\linewidth]{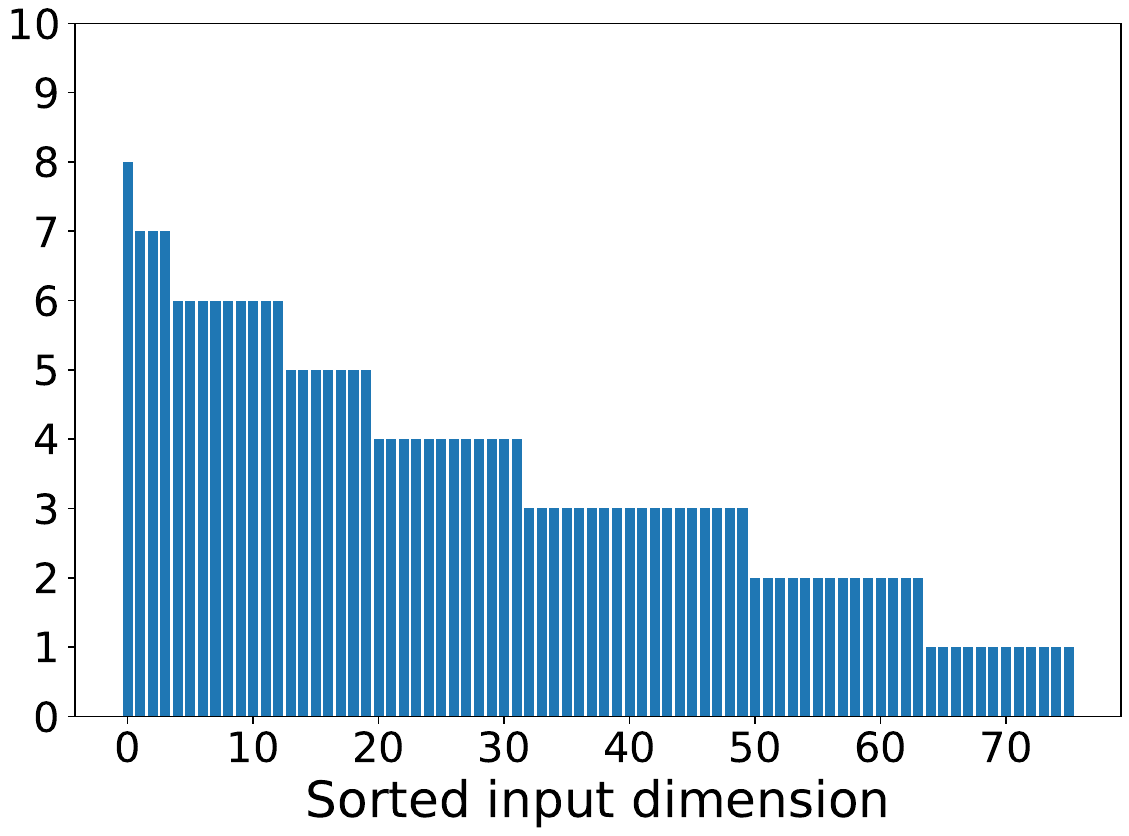}}
\subfigure[$T=0.5$M (Input)]{\includegraphics[width=.24\linewidth]{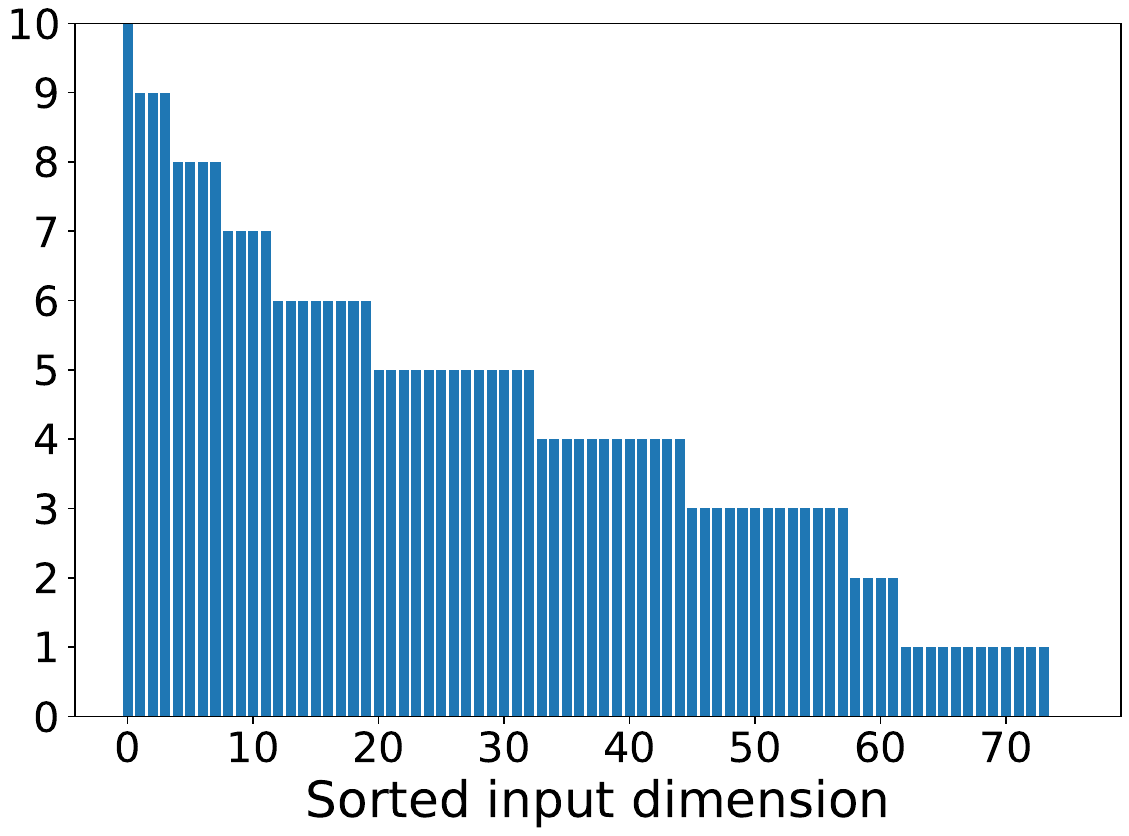}}
\subfigure[$T=1$M (Input)]{\includegraphics[width=.24\linewidth]{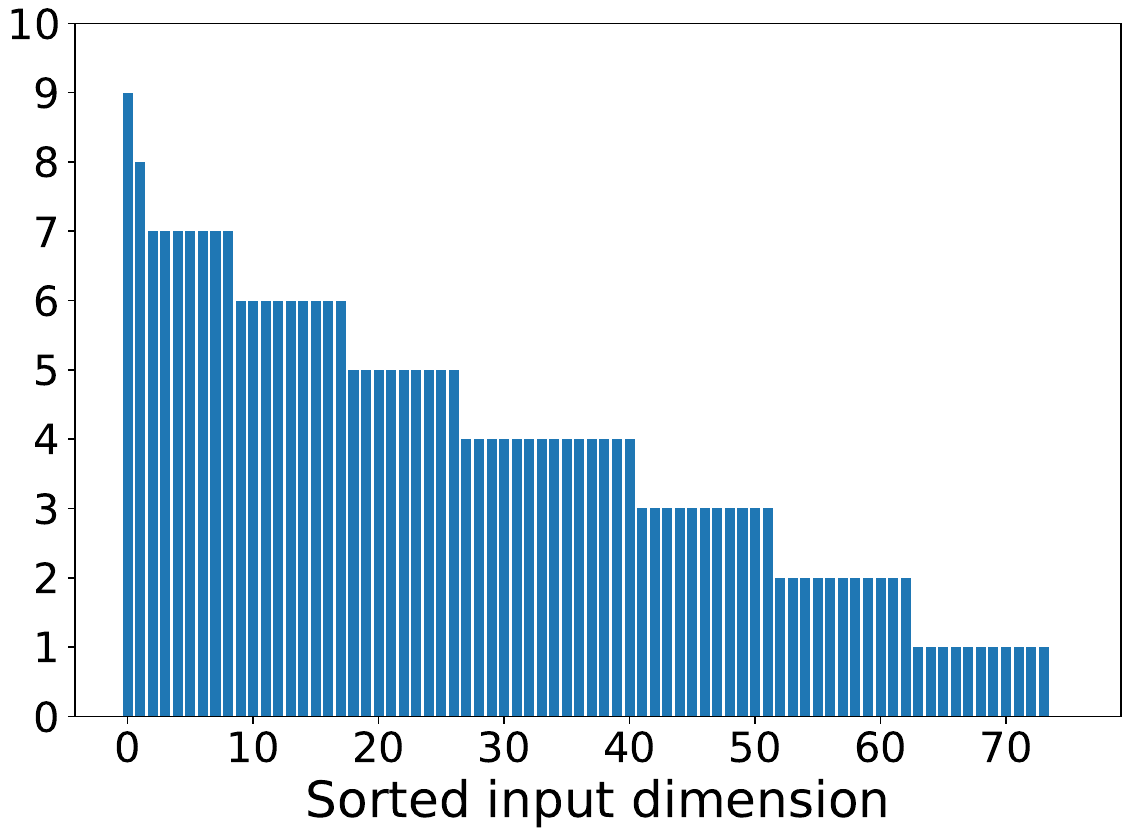}}
\subfigure[$T=2$M (Input)]{\includegraphics[width=.24\linewidth]{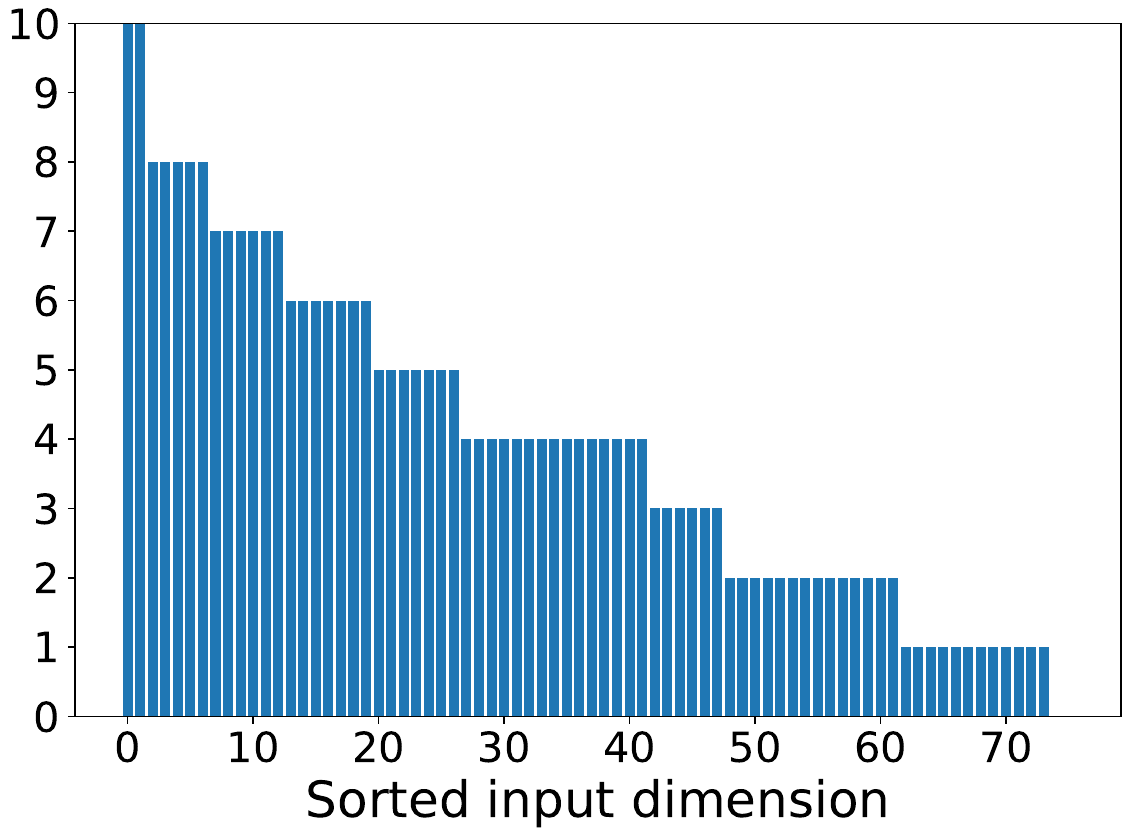}}
\subfigure[$T=0$M (Output)]{\includegraphics[width=.24\linewidth]{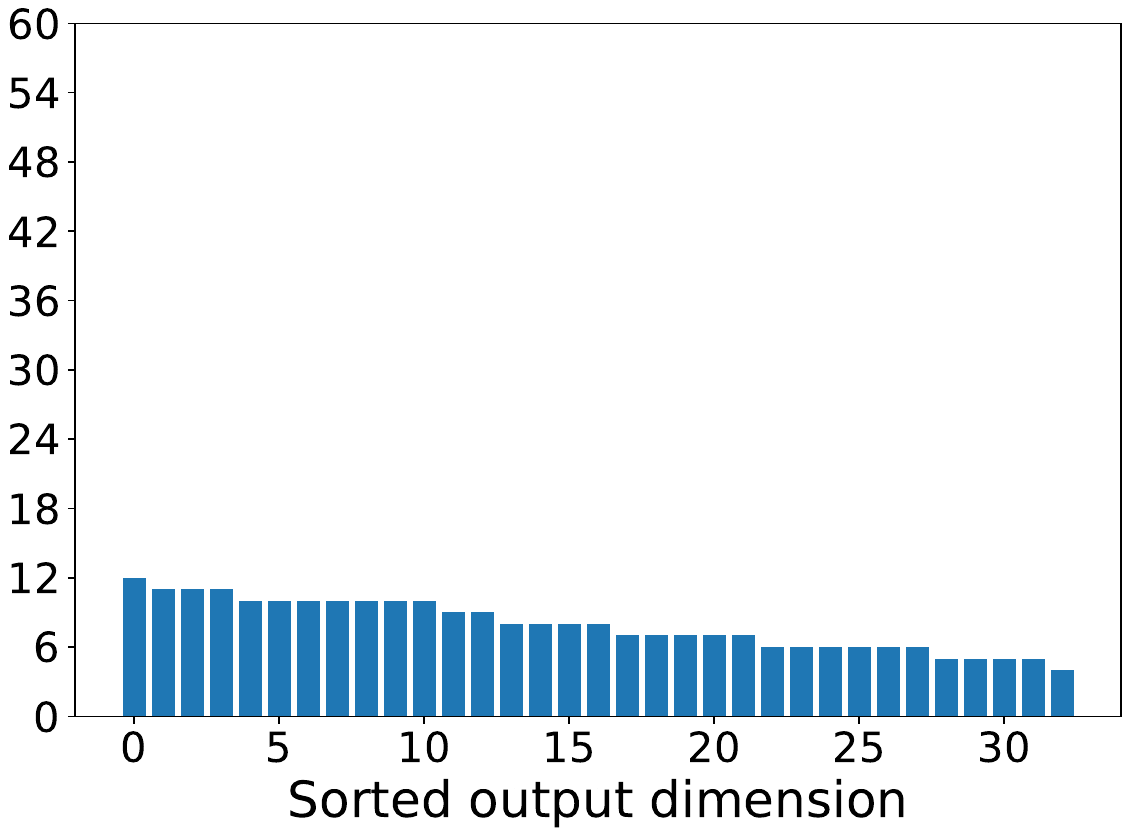}}
\subfigure[$T=0.5$M (Output)]{\includegraphics[width=.24\linewidth]{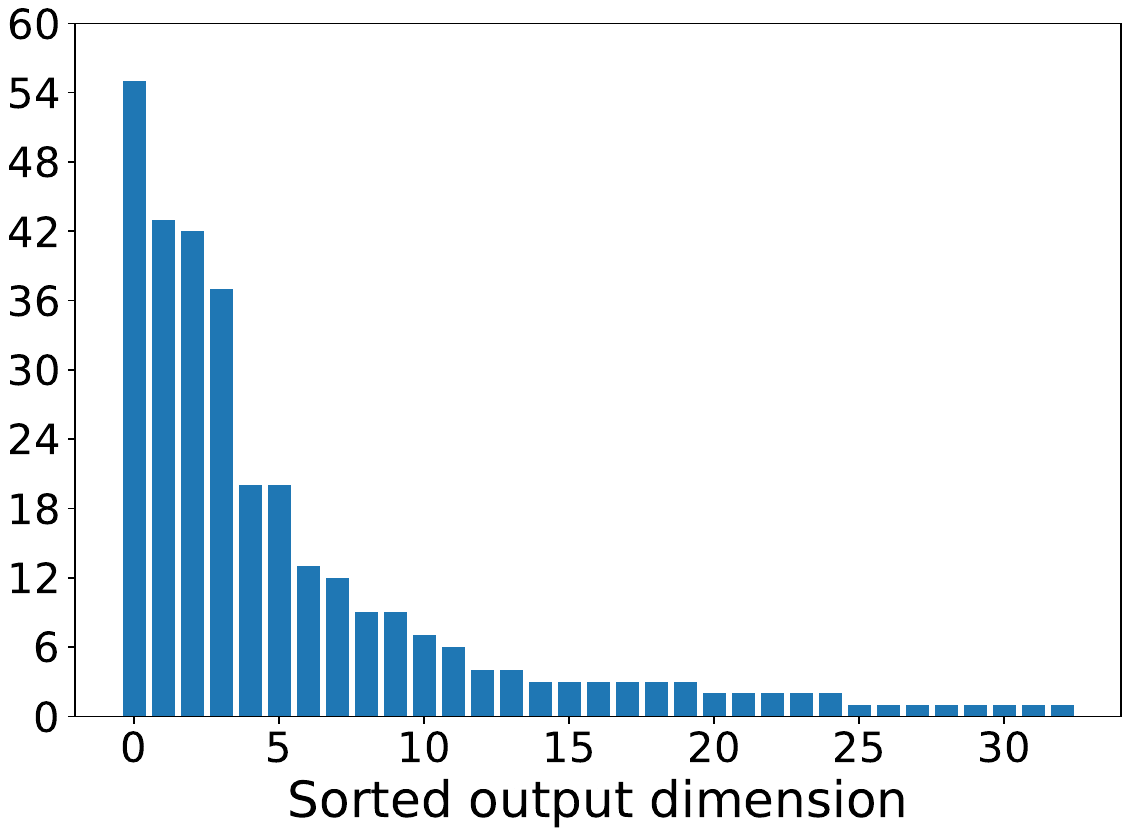}}
\subfigure[$T=1$M (Output)]{\includegraphics[width=.24\linewidth]{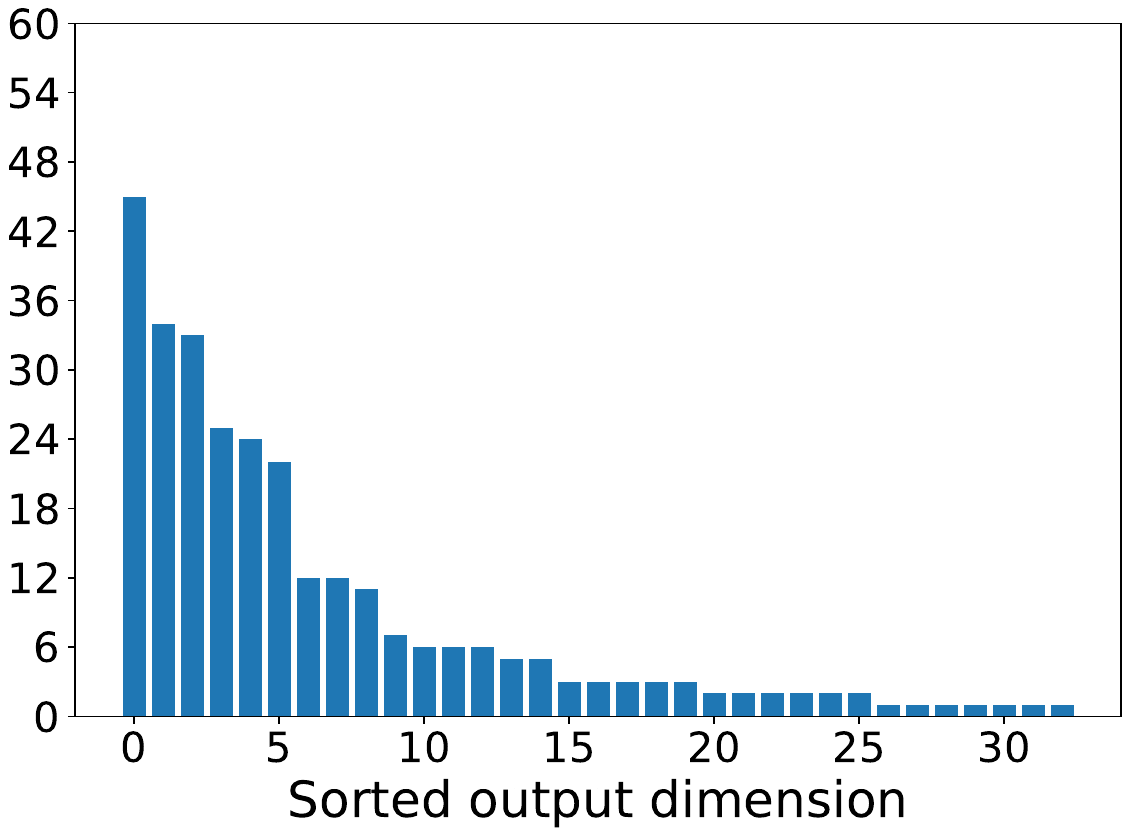}}
\subfigure[$T=2$M (Output)]{\includegraphics[width=.24\linewidth]{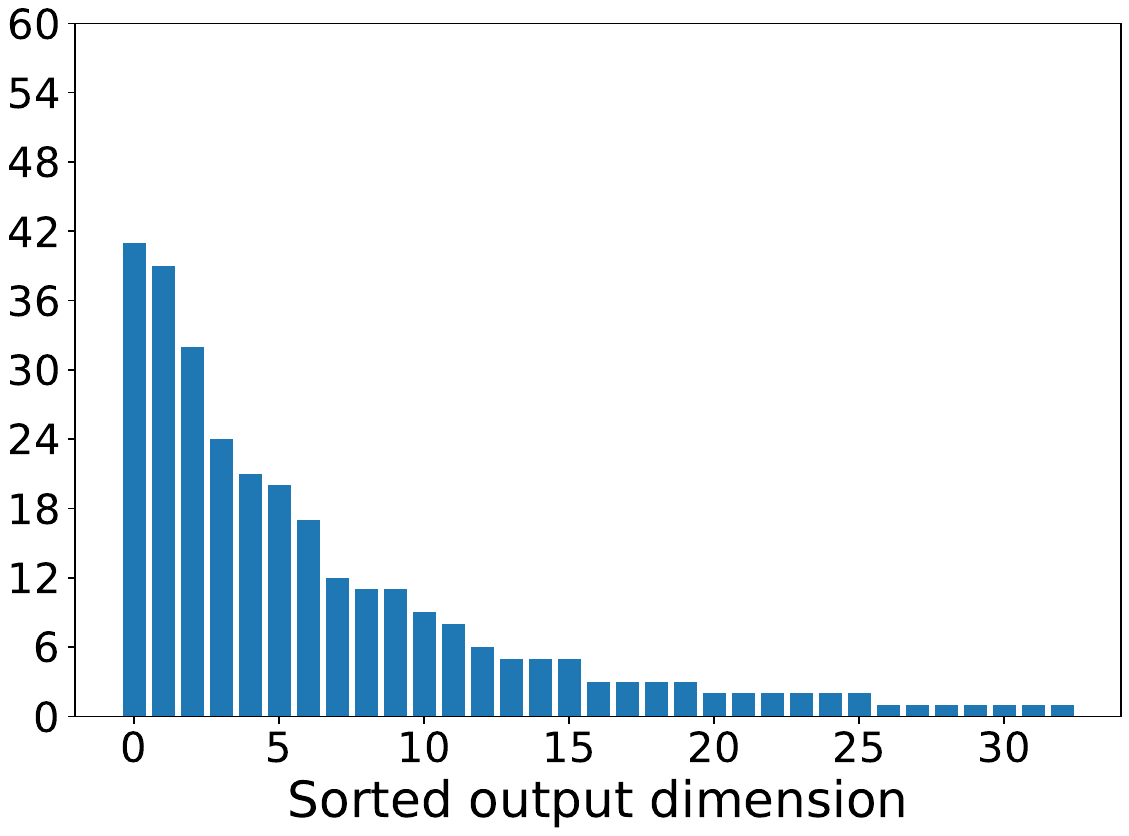}}
\caption{Number of nonzero connections for input and output dimensions in descending order of the sparse layer visualized in Figure~\ref{fig:vs1}.}
\label{fig:count1}
\end{figure}

\begin{figure}[H]
\centering
\includegraphics[width=\linewidth]{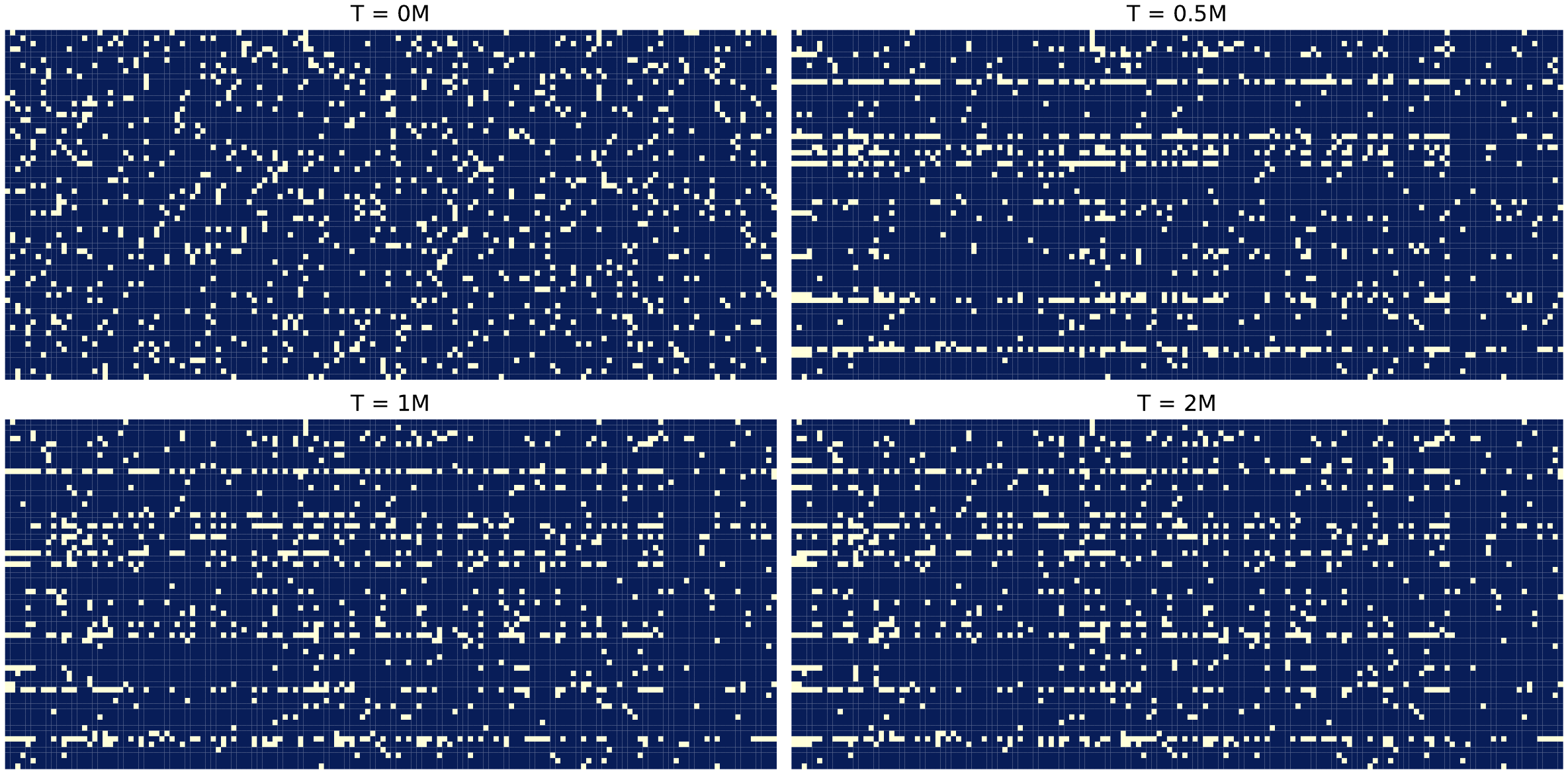}
\caption{The learned mask of the input layer weight of $Q_1$. Light pixels in row $i$ and column $j$ indicate the existence of the connection for input dimension $j$ and output dimension $i$, while the dark pixel represents the empty connection.}
\label{fig:qi_input}
\end{figure}

\begin{figure}[H]
\centering
\includegraphics[width=\linewidth]{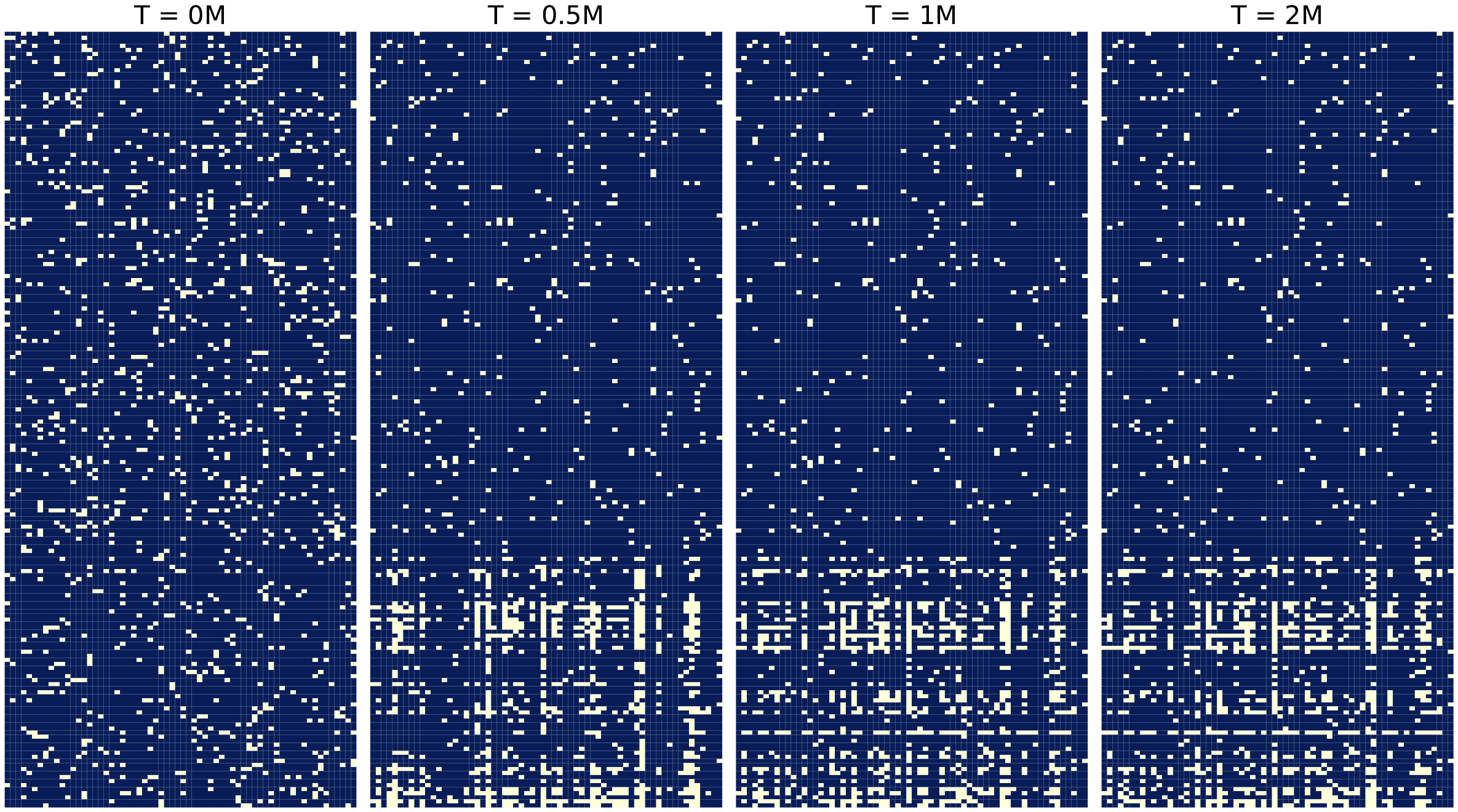}
\caption{The learned mask of the GRU layer weight of $Q_1$. Light pixels in row $i$ and column $j$ indicate the existence of the connection for input dimension $j$ and output dimension $i$, while the dark pixel represents the empty connection.}
\label{fig:qi_hidden}
\end{figure}

\begin{figure}[H]
\centering
\includegraphics[width=\linewidth]{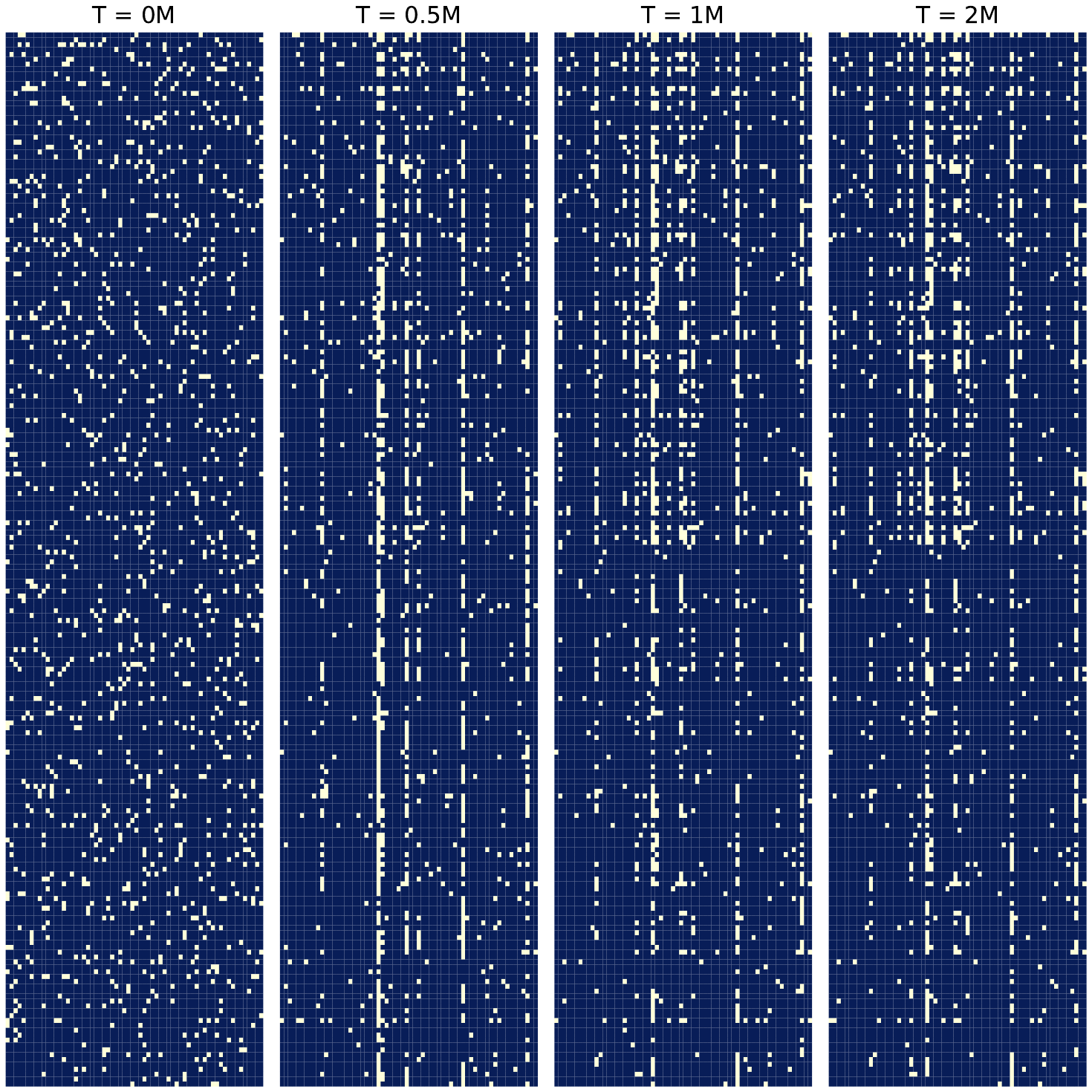}
\caption{The learned mask of the first layer weight of Hypernetwork. Light pixels in row $i$ and column $j$ indicate the existence of the connection for input dimension $j$ and output dimension $i$, while the dark pixel represents the empty connection.}
\label{fig:mix0}
\end{figure}

\begin{figure}[H]
\centering
\subfigure[$T=0$M (Output)]{\includegraphics[width=.24\linewidth]{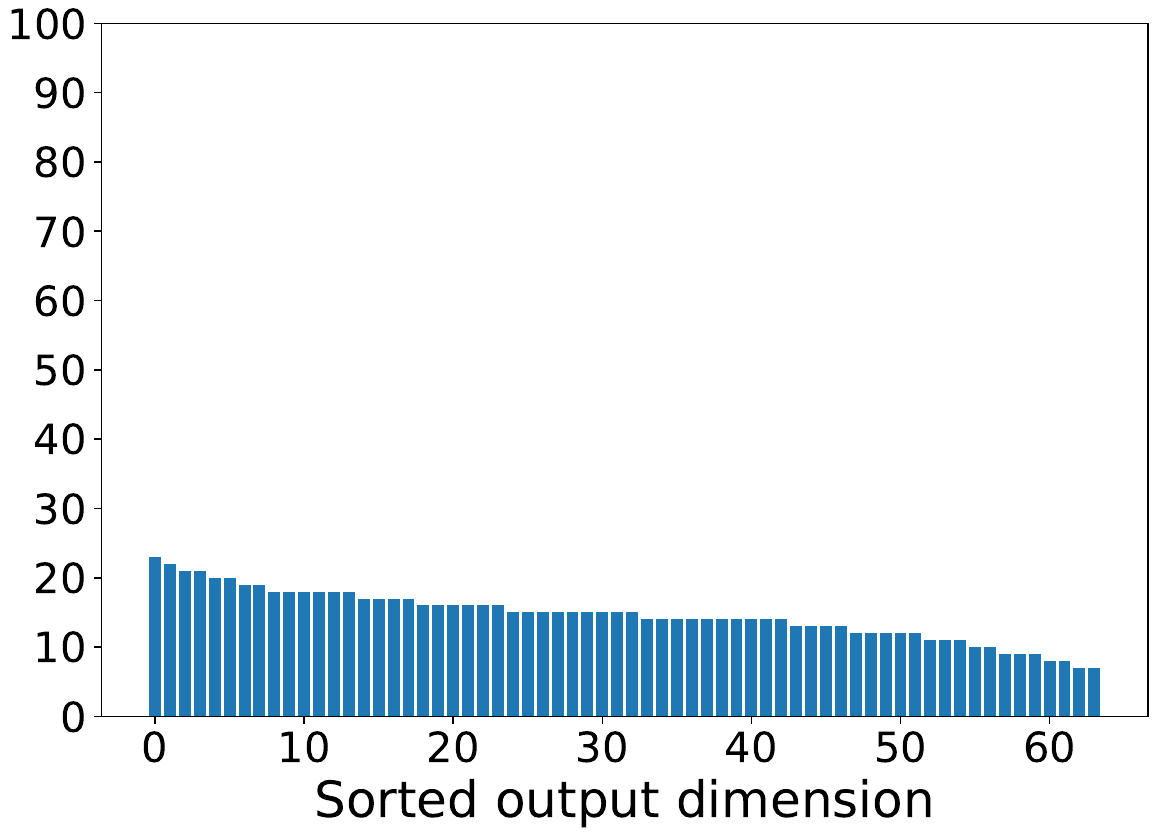}}
\subfigure[$T=0.5$M (Output)]{\includegraphics[width=.24\linewidth]{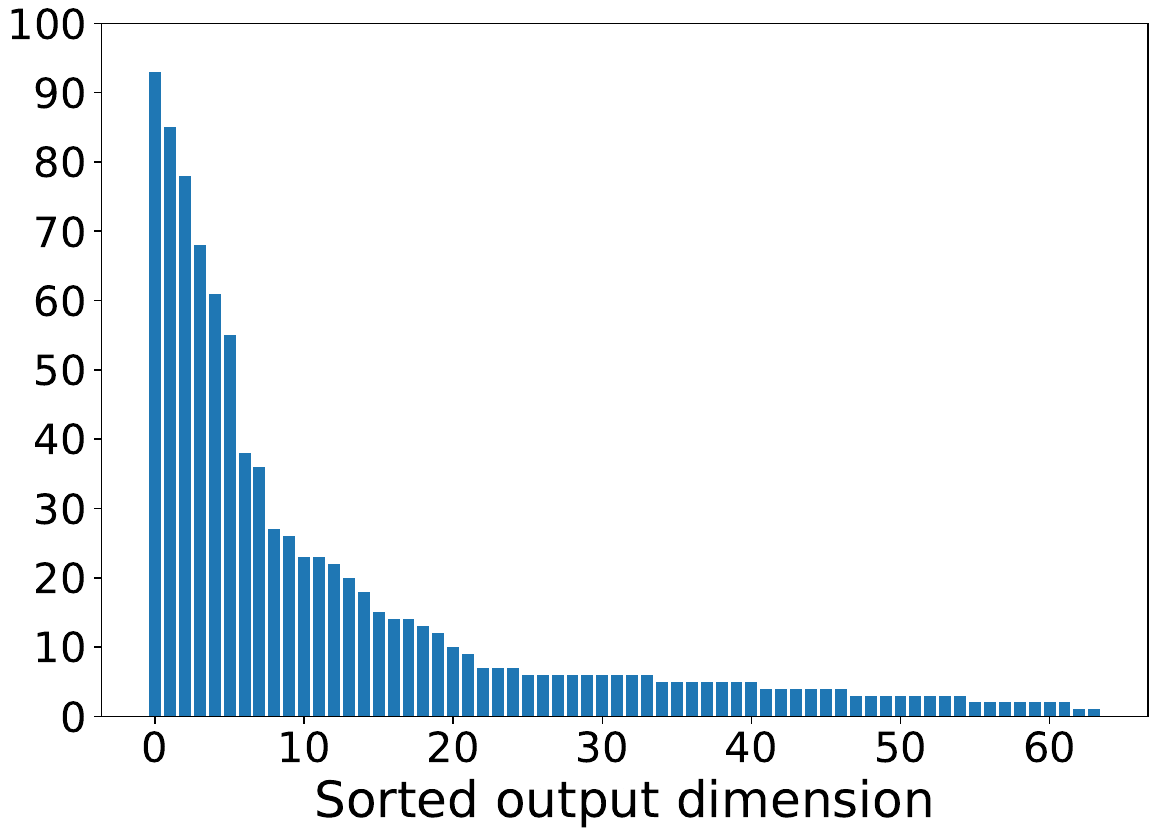}}
\subfigure[$T=1$M (Output)]{\includegraphics[width=.24\linewidth]{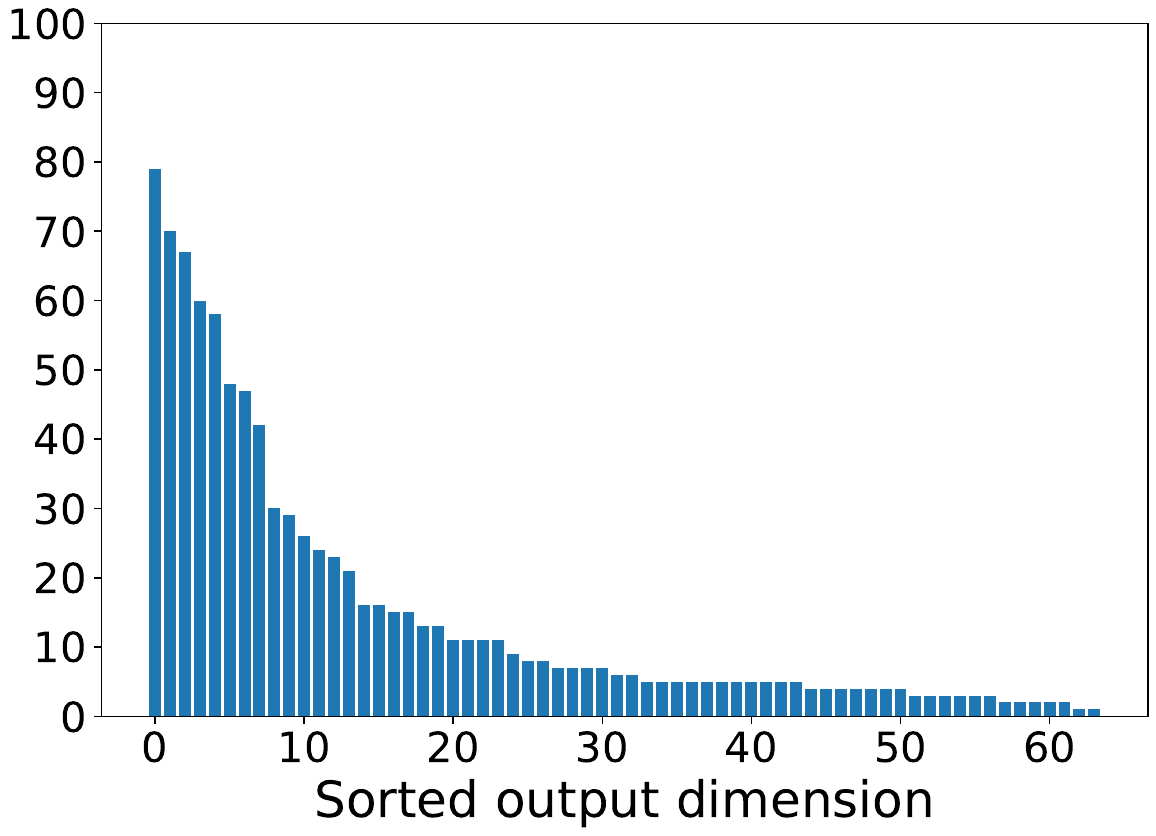}}
\subfigure[$T=2$M (Output)]{\includegraphics[width=.24\linewidth]{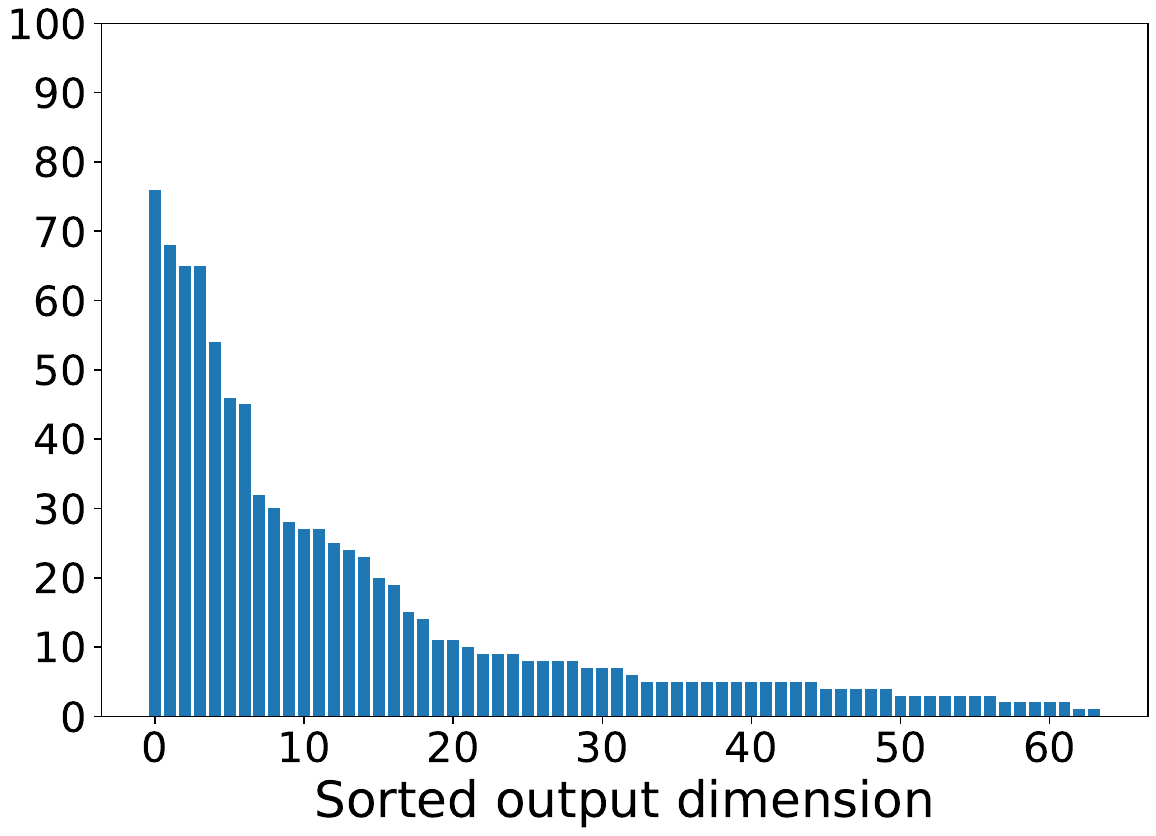}}
\vspace{-.2cm}
\caption{Number of nonzero connections for input and output dimensions in descending order of the sparse layer visualized in Figure~\ref{fig:qi_input}.}
\label{fig:count11}
\end{figure}

\begin{figure}[H]
\centering
\subfigure[$T=0$M (Output)]{\includegraphics[width=.24\linewidth]{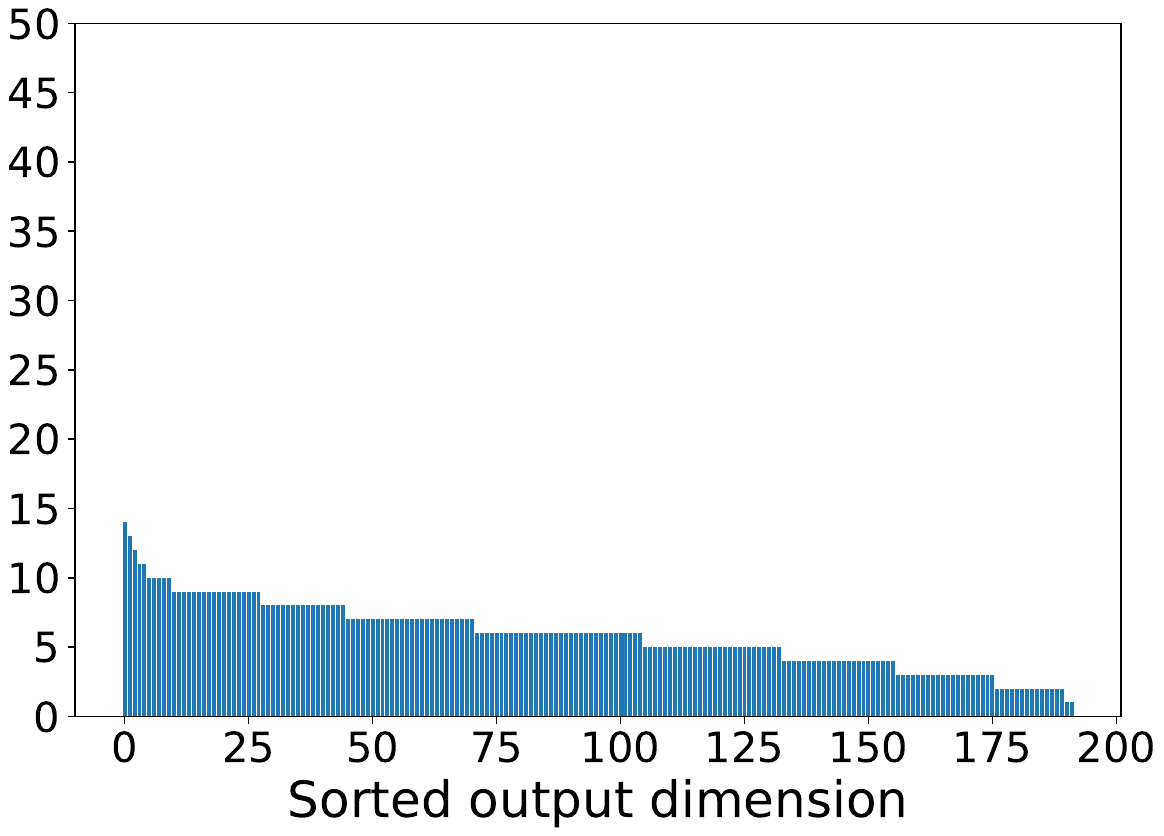}}
\subfigure[$T=0.5$M (Output)]{\includegraphics[width=.24\linewidth]{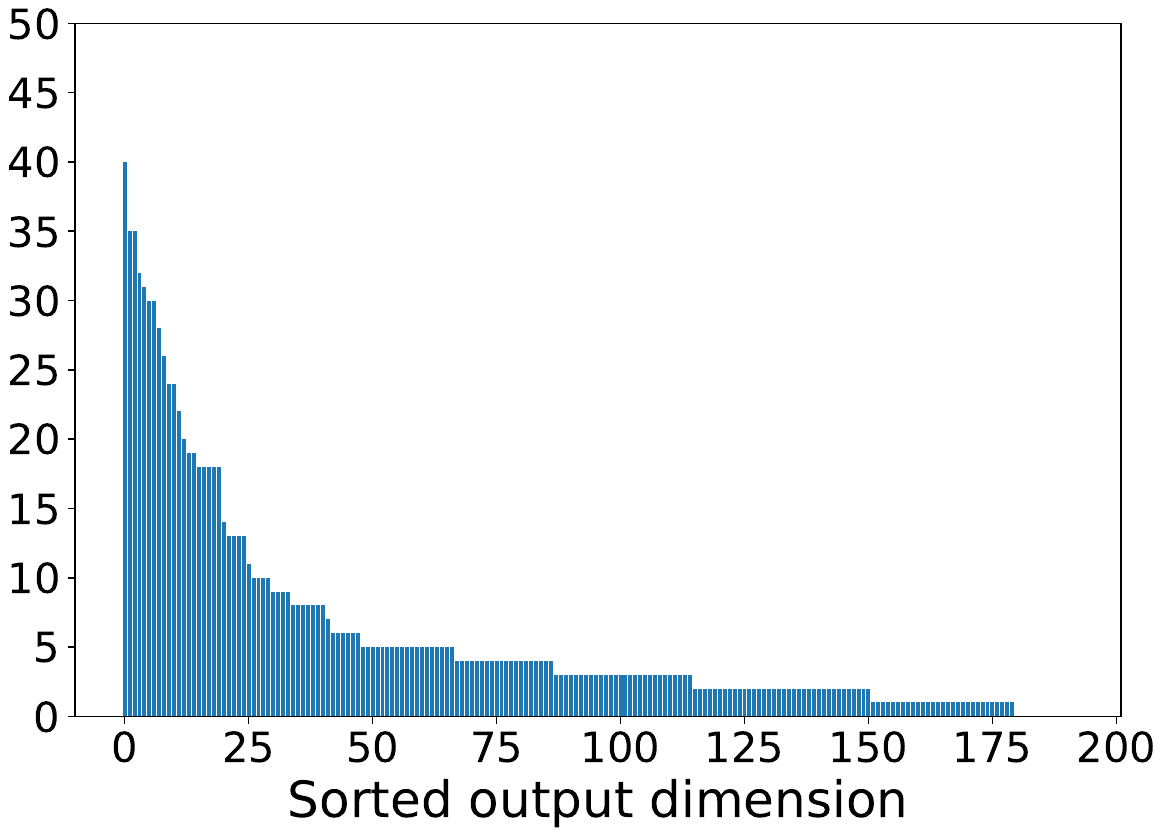}}
\subfigure[$T=1$M (Output)]{\includegraphics[width=.24\linewidth]{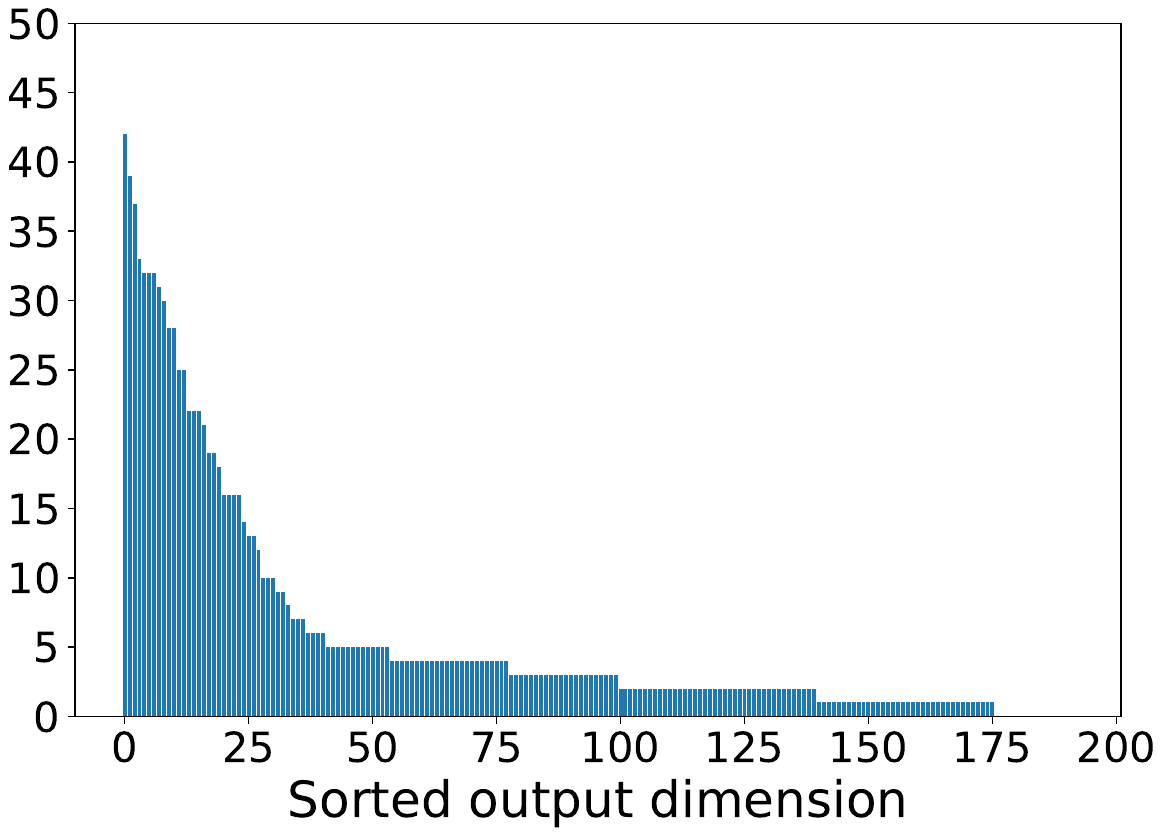}}
\subfigure[$T=2$M (Output)]{\includegraphics[width=.24\linewidth]{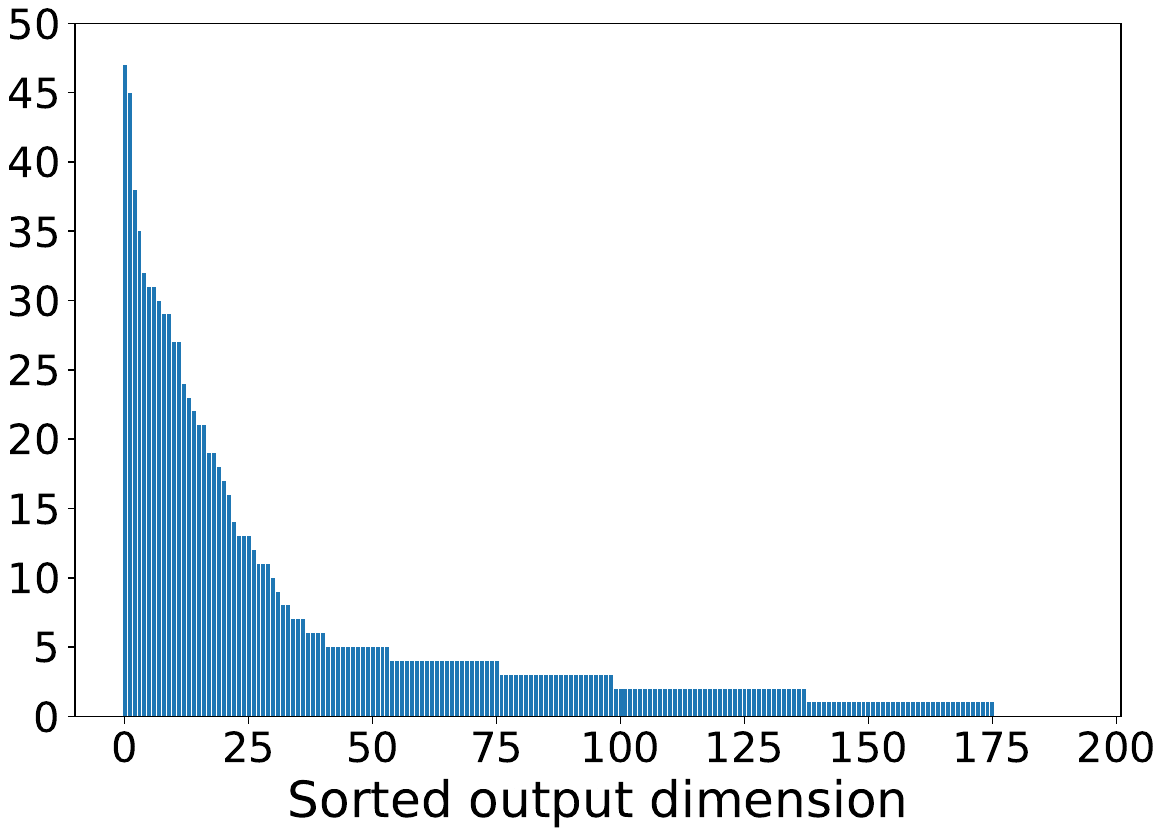}}
\vspace{-.2cm}
\caption{Number of nonzero connections for input and output dimensions in descending order of the sparse layer visualized in Figure~\ref{fig:qi_hidden}.}
\label{fig:count21}
\end{figure}

\begin{figure}[H]
\centering
\subfigure[$T=0$M]{\includegraphics[width=.24\linewidth]{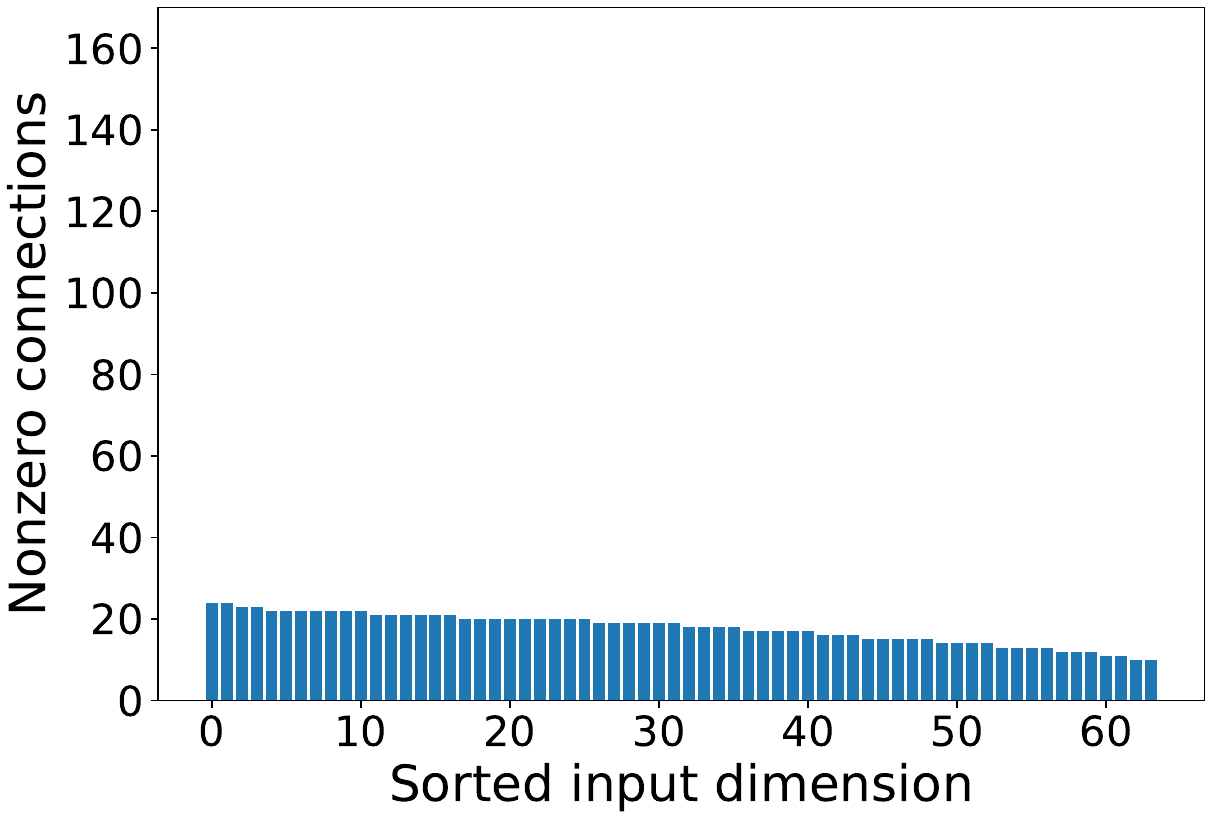}}
\subfigure[$T=0.5$M]{\includegraphics[width=.24\linewidth]{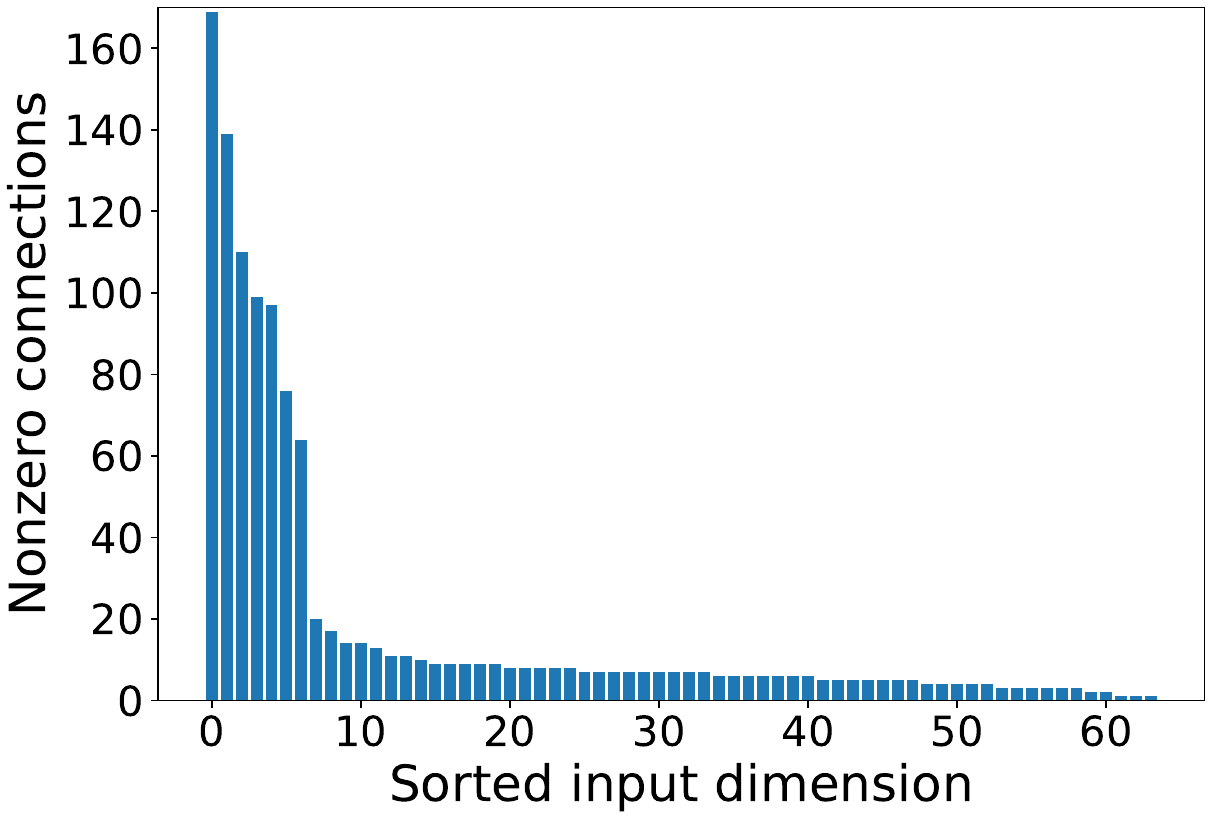}}
\subfigure[$T=1$M]{\includegraphics[width=.24\linewidth]{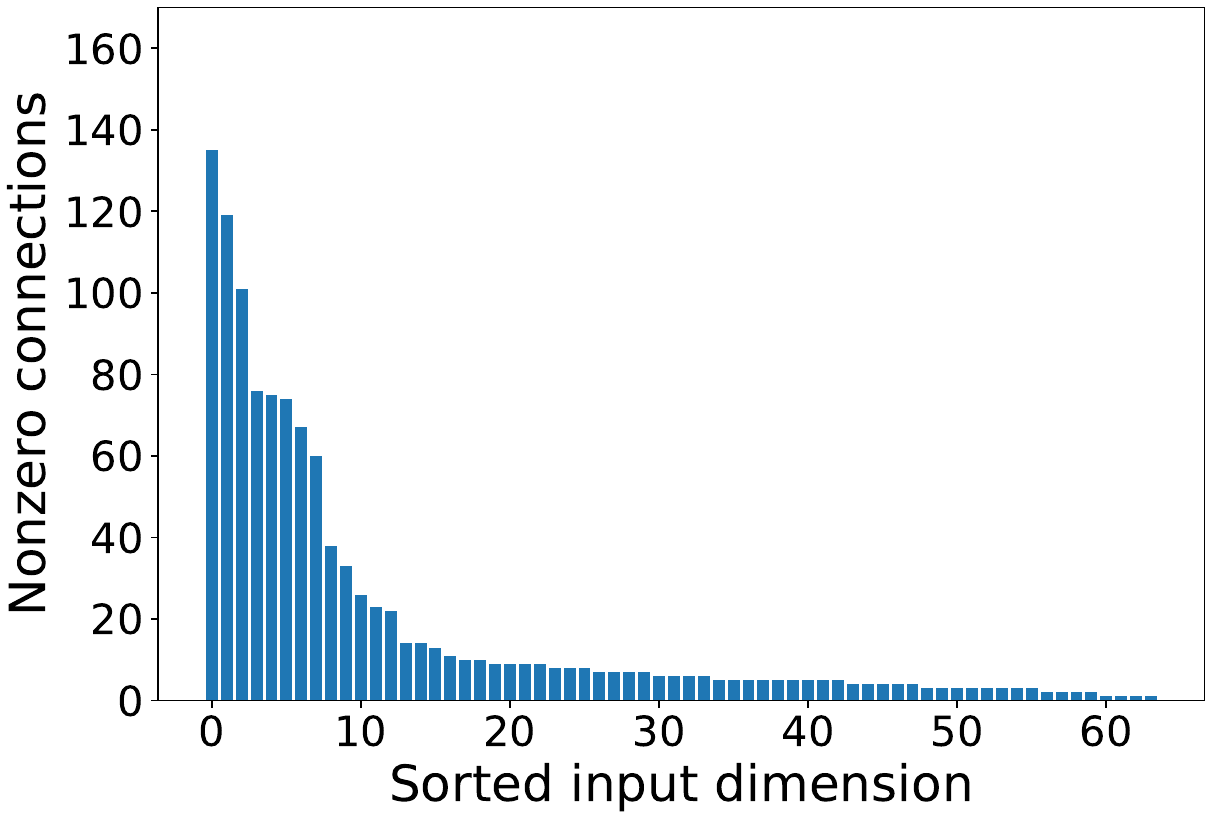}}
\subfigure[$T=2$M]{\includegraphics[width=.24\linewidth]{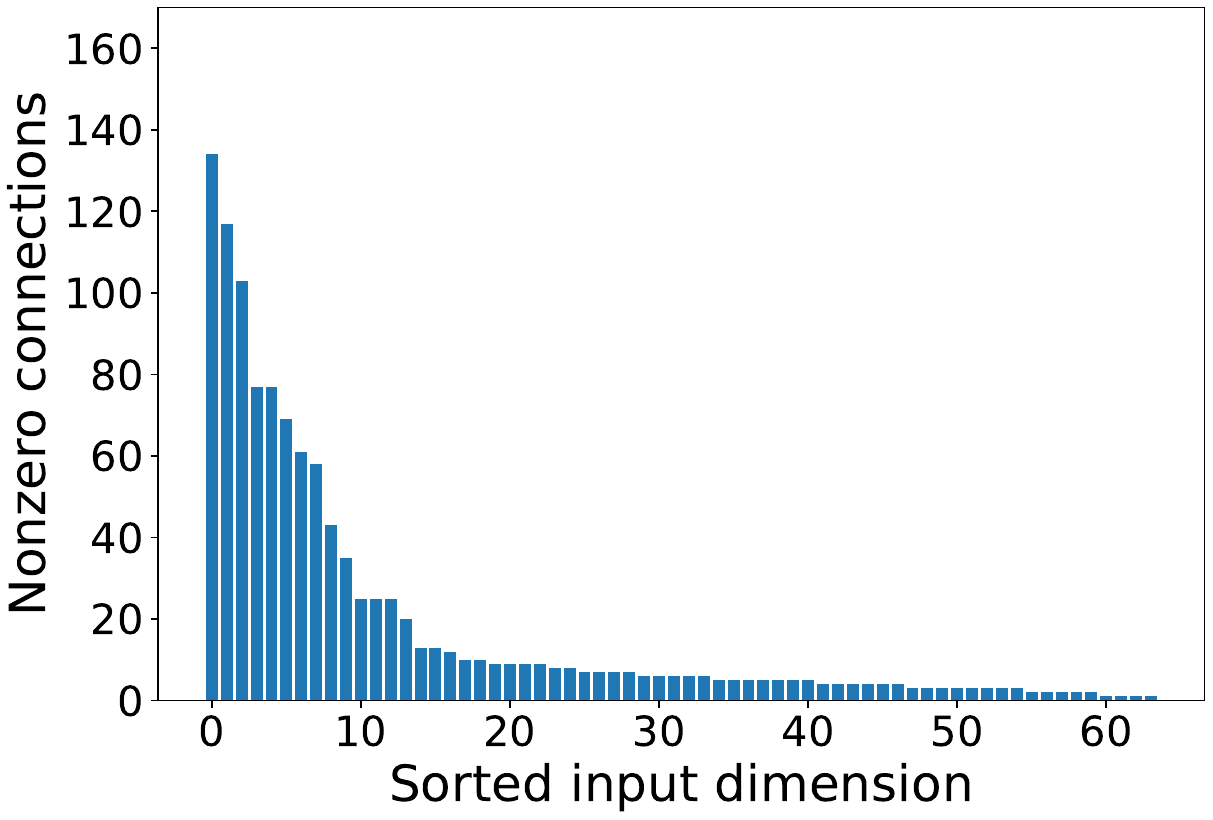}}
\subfigure[$T=0$M]{\includegraphics[width=.24\linewidth]{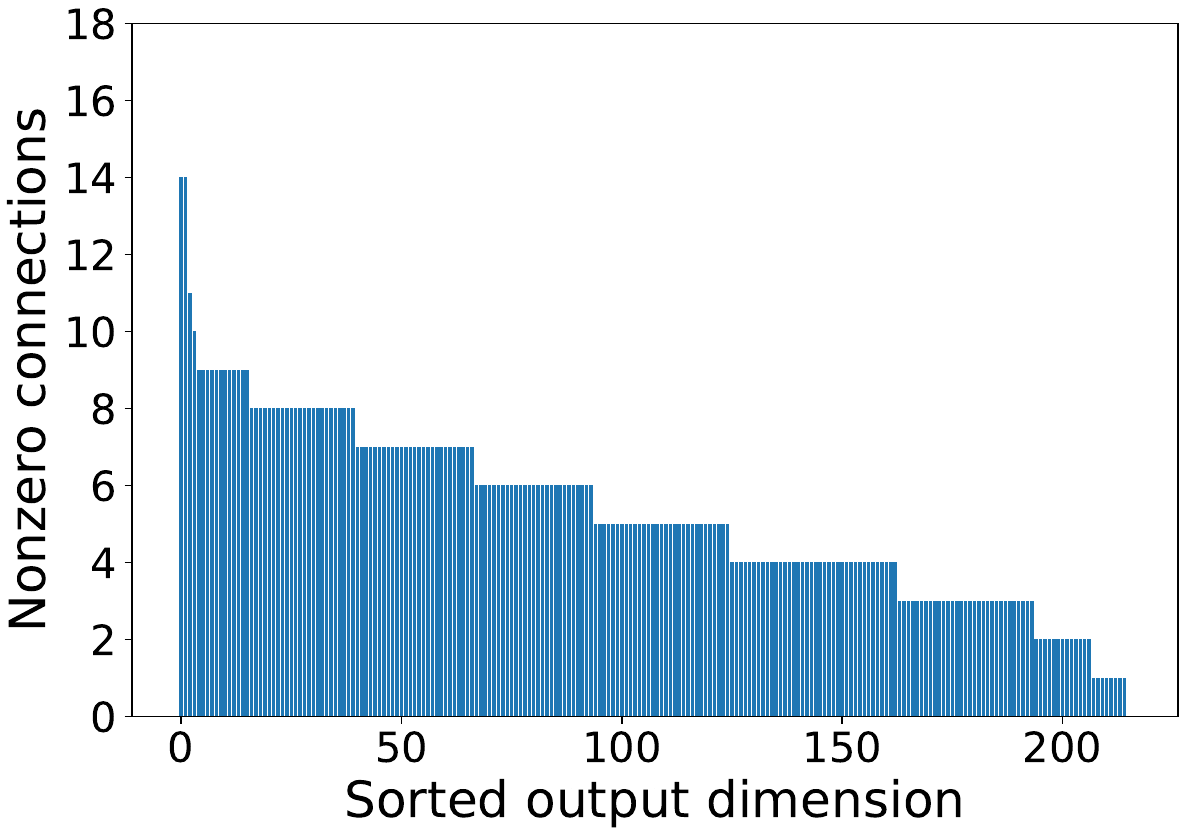}}
\subfigure[$T=0.5$M]{\includegraphics[width=.24\linewidth]{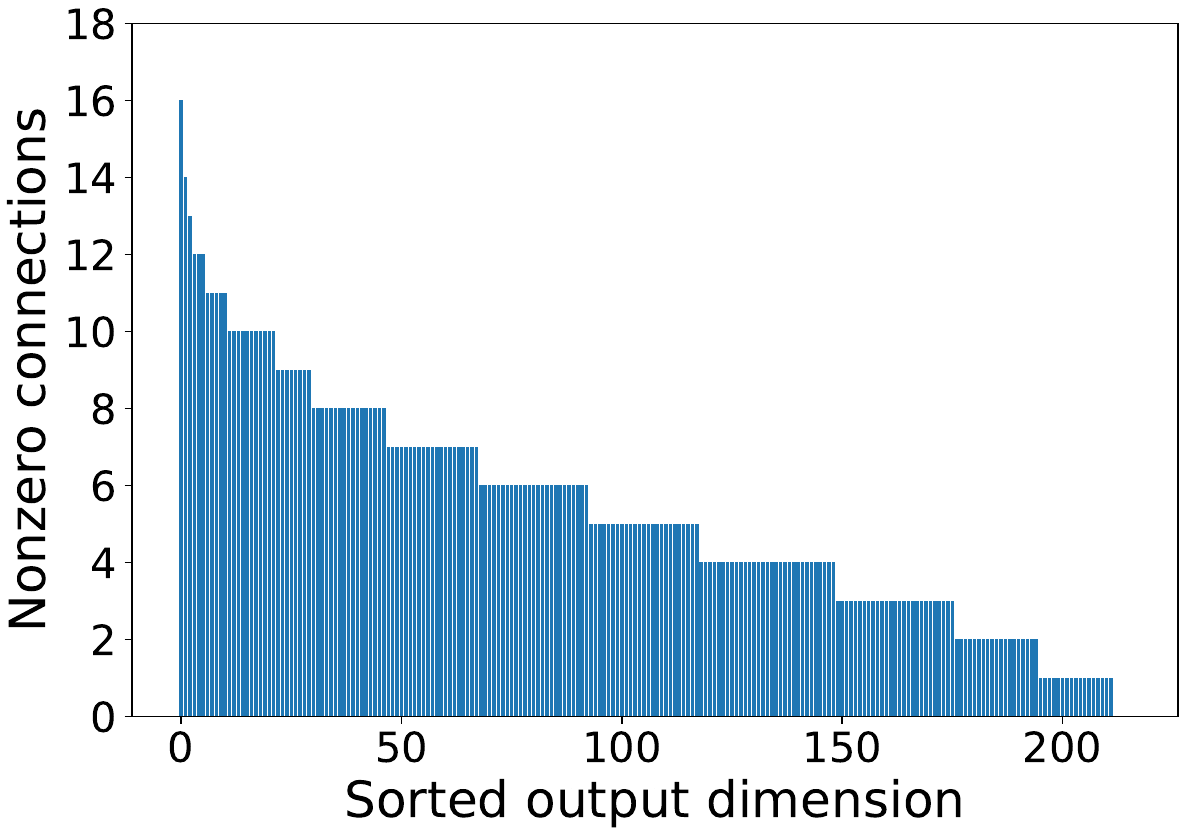}}
\subfigure[$T=1$M]{\includegraphics[width=.24\linewidth]{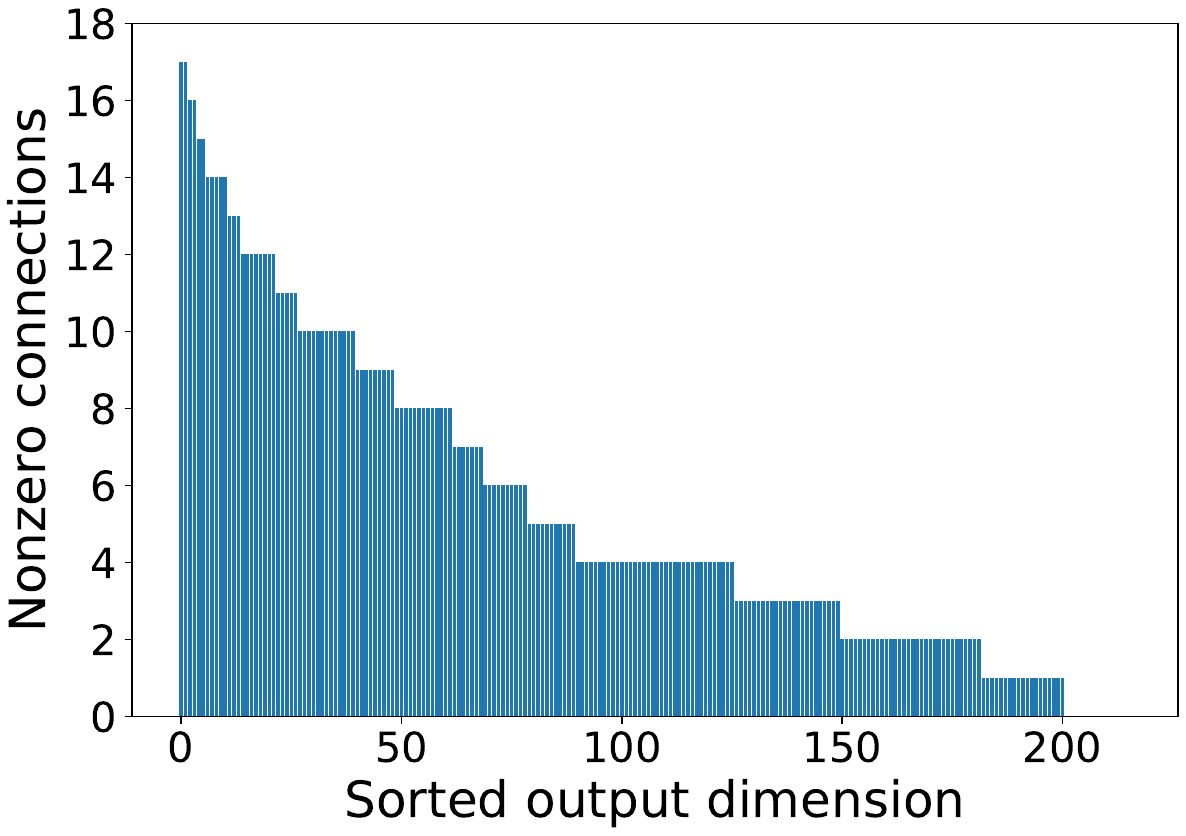}}
\subfigure[$T=2$M]{\includegraphics[width=.24\linewidth]{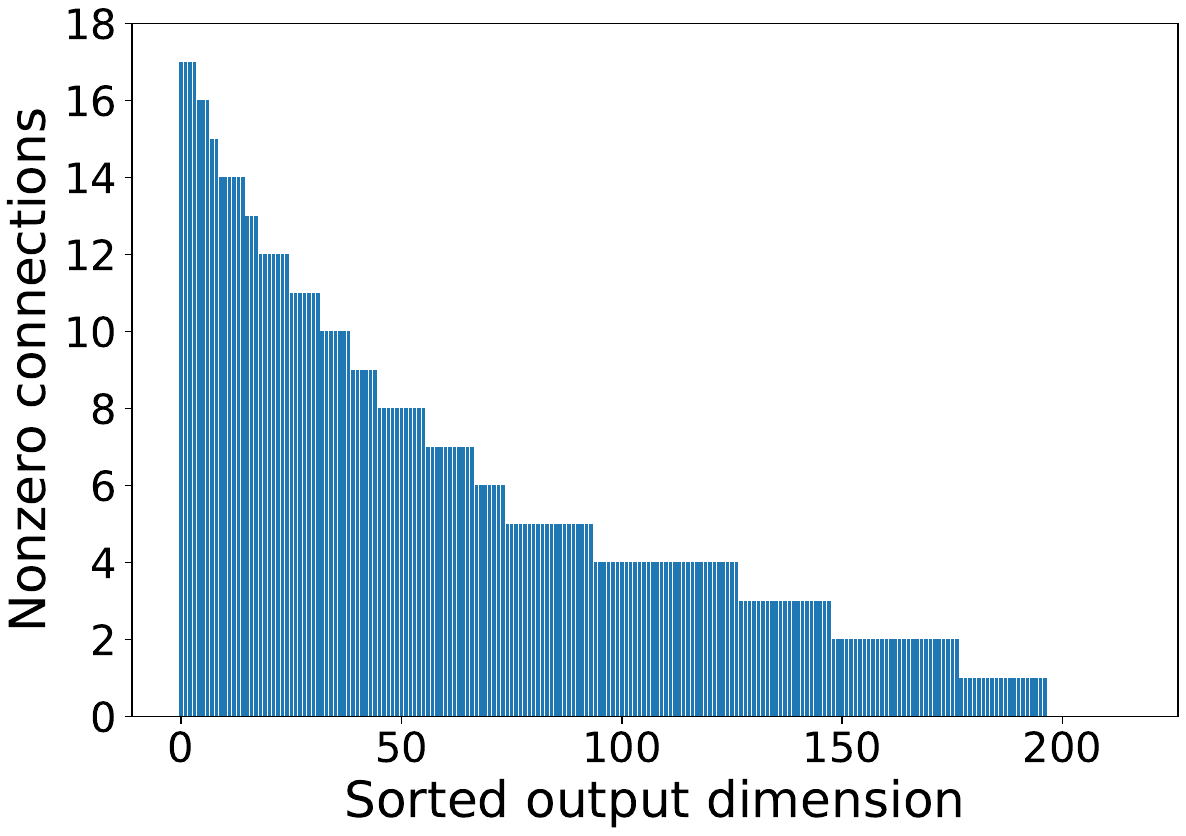}}
\vspace{-.2cm}
\caption{Number of nonzero connections for input and output dimensions in descending order of the sparse layer visualized in Figure~\ref{fig:mix0}.}
\label{fig:count31}
\end{figure}

\vspace{-.2cm}
We present a series of visualizations capturing the evolution of masks within network layers during the MAST-QMIX training in the {\ttfamily 3s5z} scenario. These figures, specifically Figure~\ref{fig:qi_input} (a detailed view of Figure~\ref{fig:vs1}, Figure~\ref{fig:qi_hidden}, and Figure~\ref{fig:mix0}, offer intriguing insights. Additionally, we provide connection counts in Figure~\ref{fig:count11}, \ref{fig:count21} and \ref{fig:count31}, for input and output dimensions in each sparse mask, highlighting pruned dimensions. To facilitate a clearer perspective on connection distributions, we sort dimensions based on the descending order of nonzero connections, focusing on the distribution rather than specific dimension ordering. The connection counts associated with Figure~\ref{fig:vs1} in the main paper s given in Figure~\ref{fig:count1}.
\vspace{-.6cm}
\vspace{-.6cm}
During the initial phases of training, a noticeable shift in the mask configuration becomes evident, signifying a dynamic restructuring process. As the training progresses, connections within the hidden layers gradually coalesce into a subset of neurons. This intriguing phenomenon underscores the distinct roles assumed by individual neurons in the representation process, thereby accentuating the significant redundancy prevalent in dense models.
\begin{itemize}
    \item Figure~\ref{fig:qi_input} provides insights into the input layer, revealing that certain output dimensions can be omitted while preserving the necessity of each input dimension.
    \item Figure~\ref{fig:qi_hidden} showcases analogous observations, reinforcing the idea that only a subset of output neurons is indispensable, even within the hidden layer of the GRU.
    \item Figure~\ref{fig:mix0} presents distinct findings, shedding light on the potential redundancy of certain input dimensions in learning the hyperparameters within the hypernetwork.
\end{itemize}
\newpage
\vspace{-1cm}
\vspace{-.6cm}
\newpage
\begin{figure}[H]
\centering
\subfigure[$T=0$M (Input)]{\includegraphics[width=.24\linewidth]{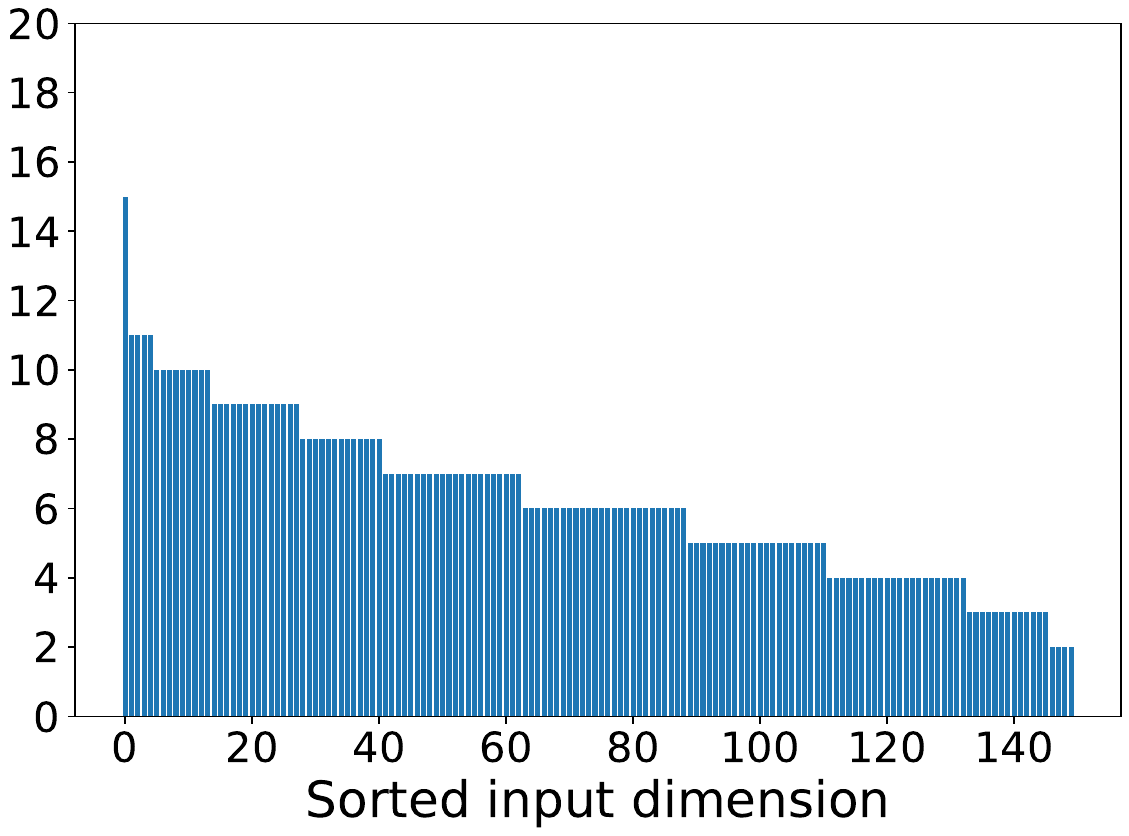}}
\subfigure[$T=0.5$M (Input)]{\includegraphics[width=.24\linewidth]{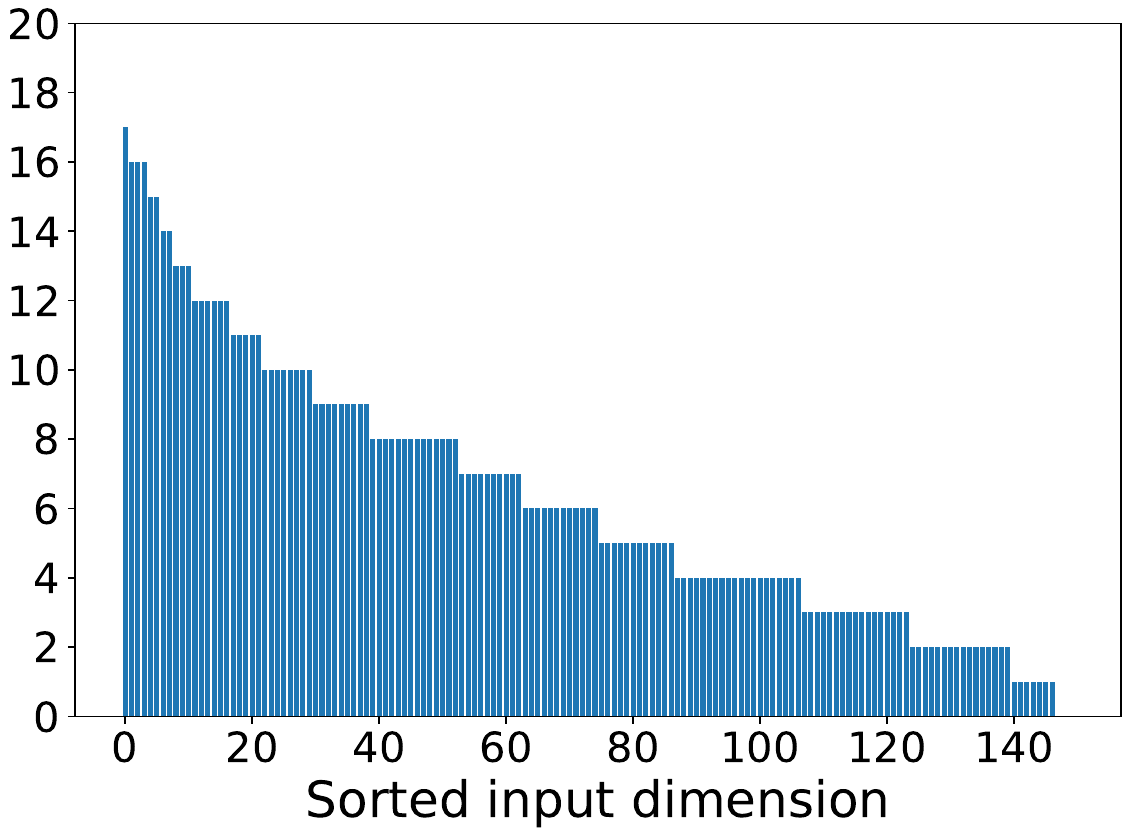}}
\subfigure[$T=1$M (Input)]{\includegraphics[width=.24\linewidth]{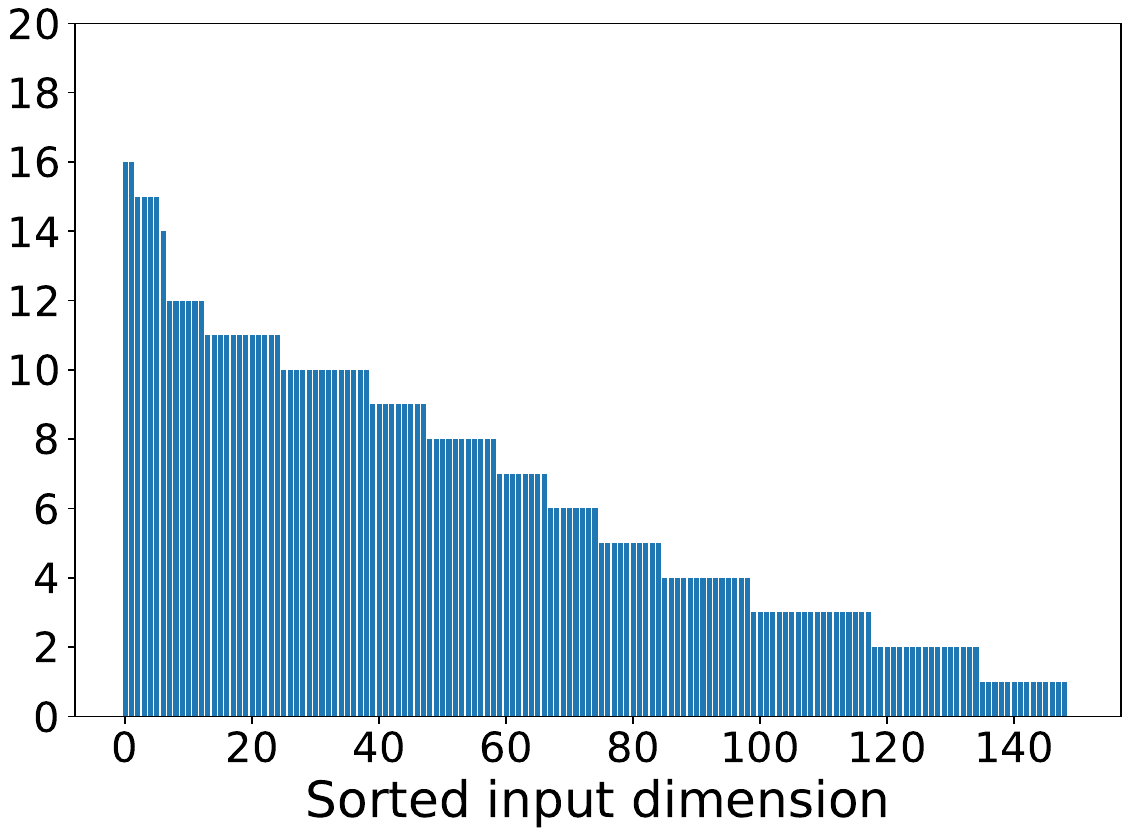}}
\subfigure[$T=2$M (Input)]{\includegraphics[width=.24\linewidth]{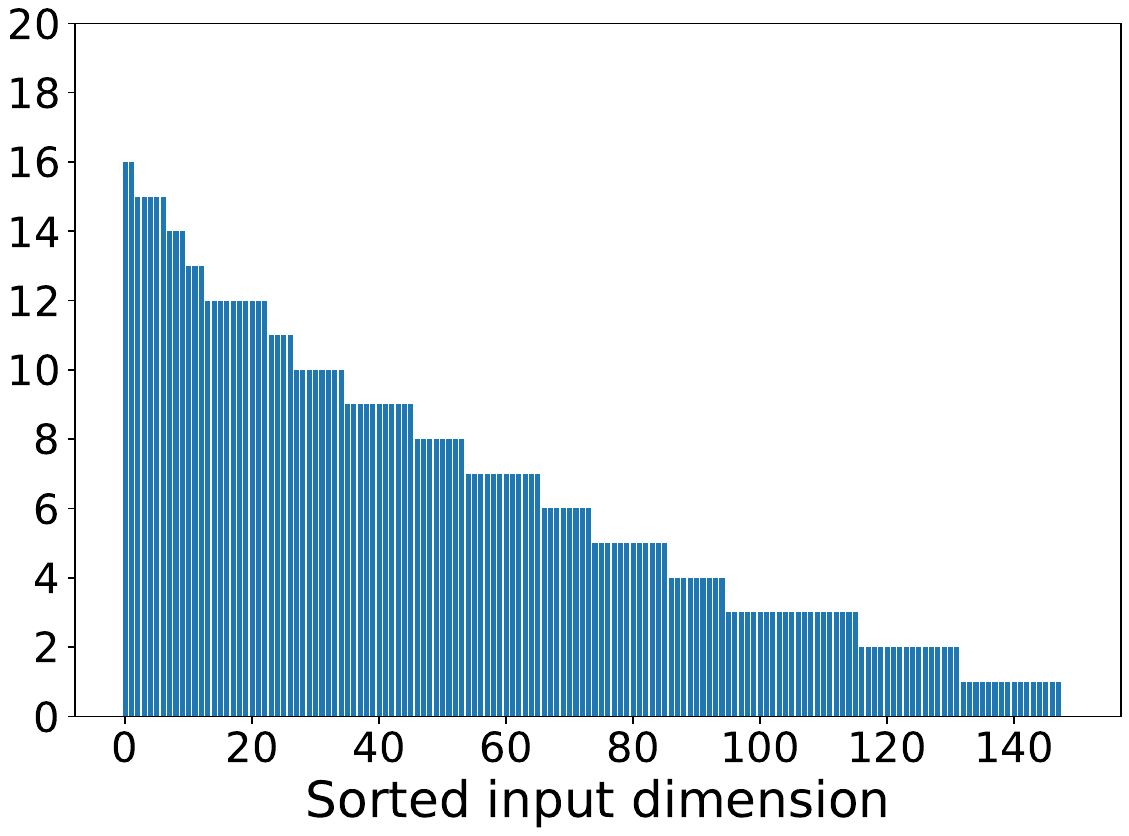}}
\end{figure}
\vspace{-1.1cm}
\vspace{-.5cm}
\begin{figure}[H]
\centering
\subfigure[$T=0$M (Input)]{\includegraphics[width=.24\linewidth]{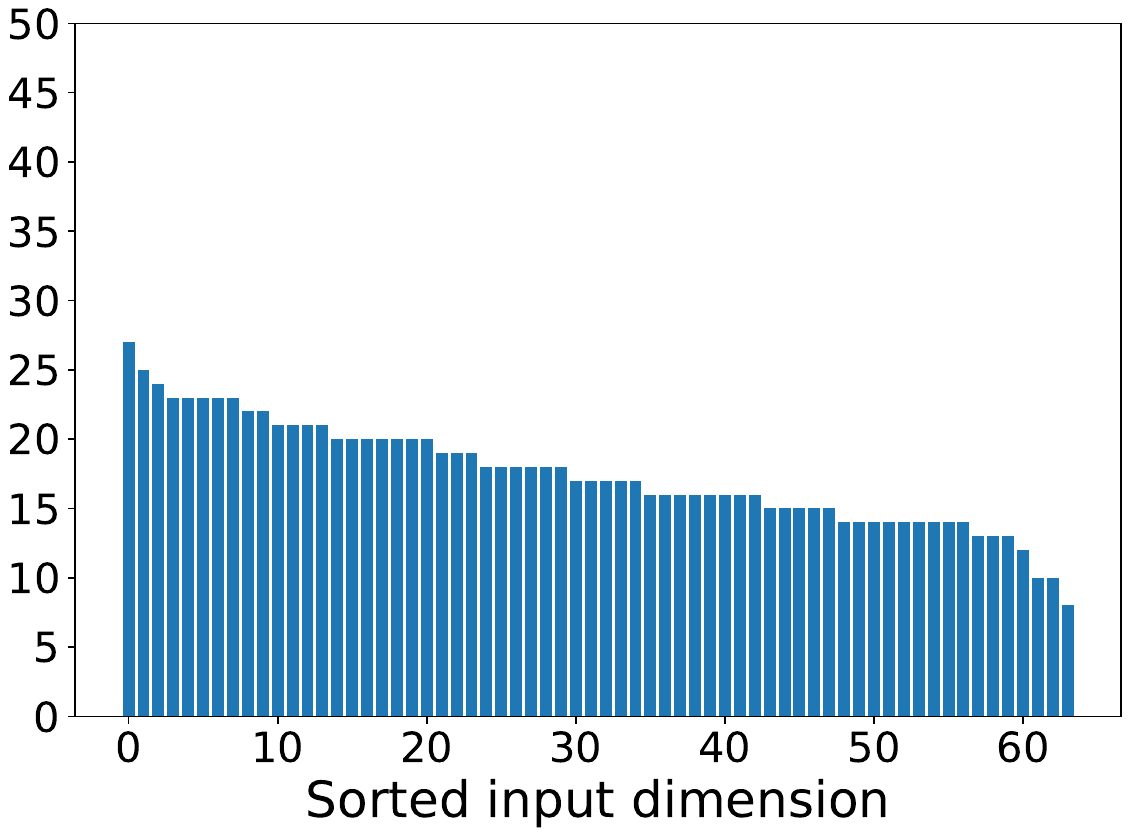}}
\subfigure[$T=0.5$M (Input)]{\includegraphics[width=.24\linewidth]{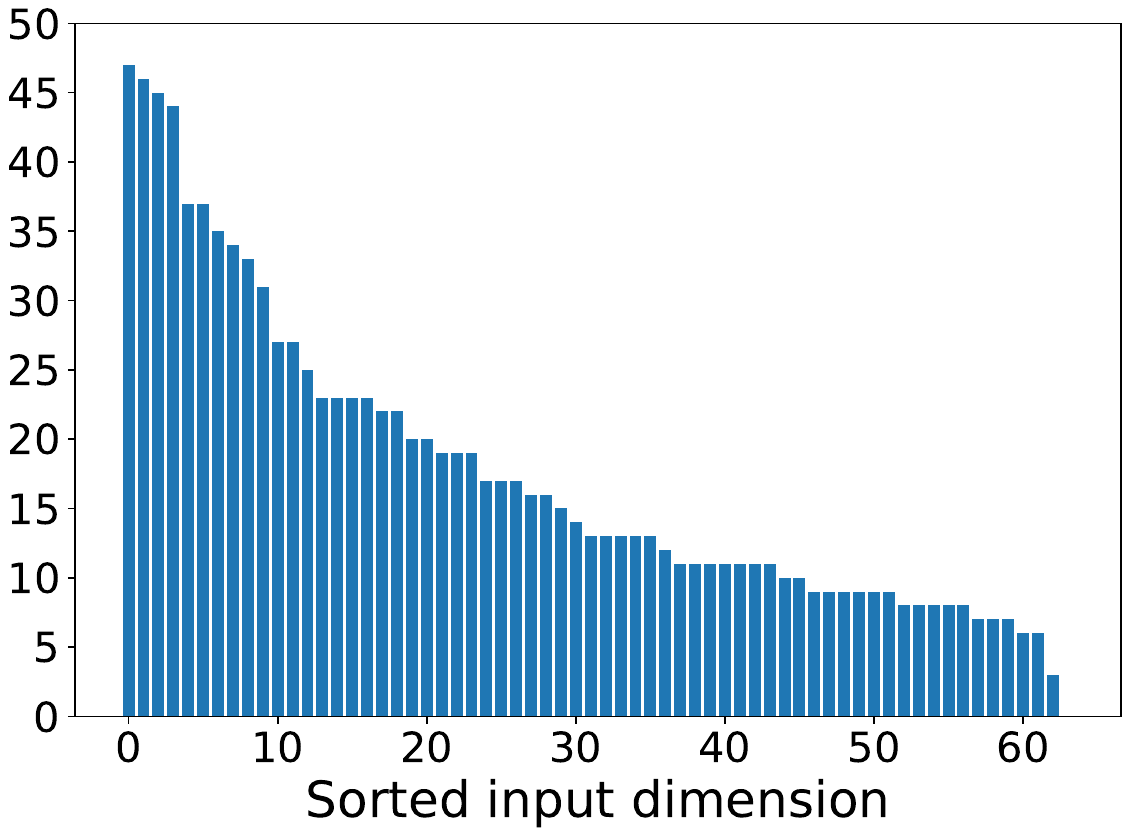}}
\subfigure[$T=1$M (Input)]{\includegraphics[width=.24\linewidth]{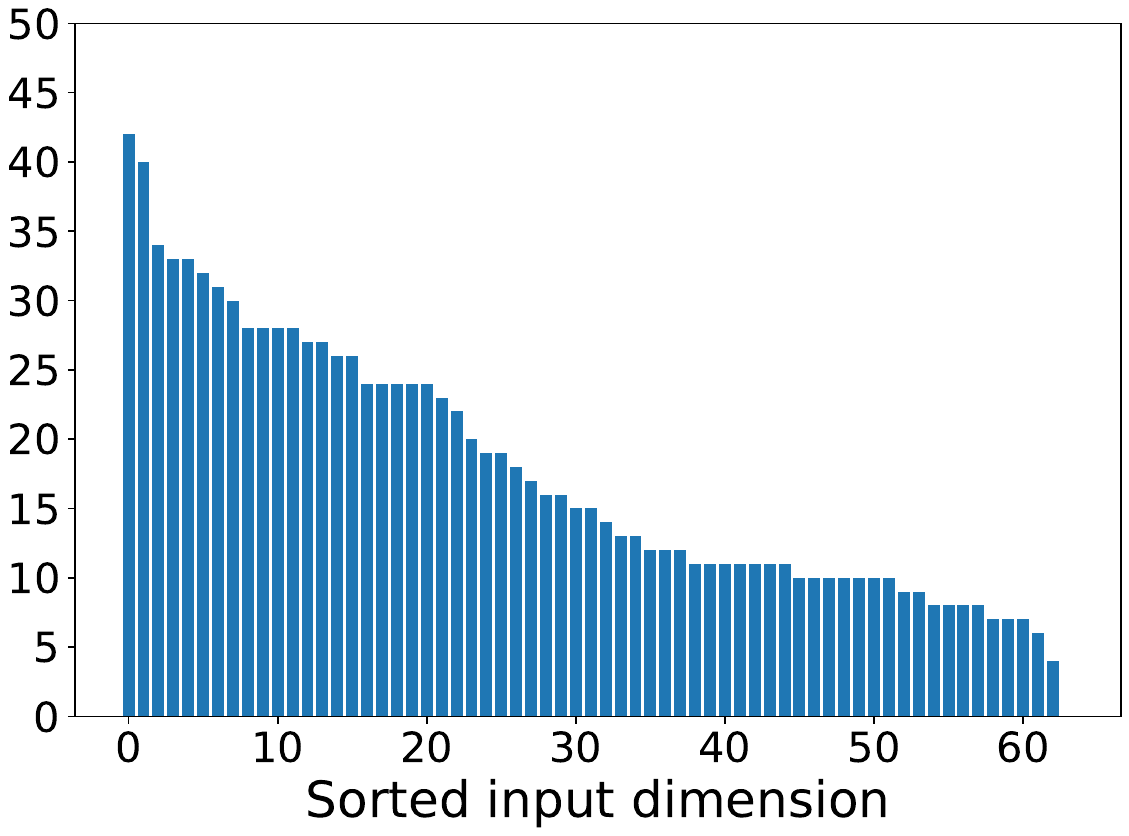}}
\subfigure[$T=2$M (Input)]{\includegraphics[width=.24\linewidth]{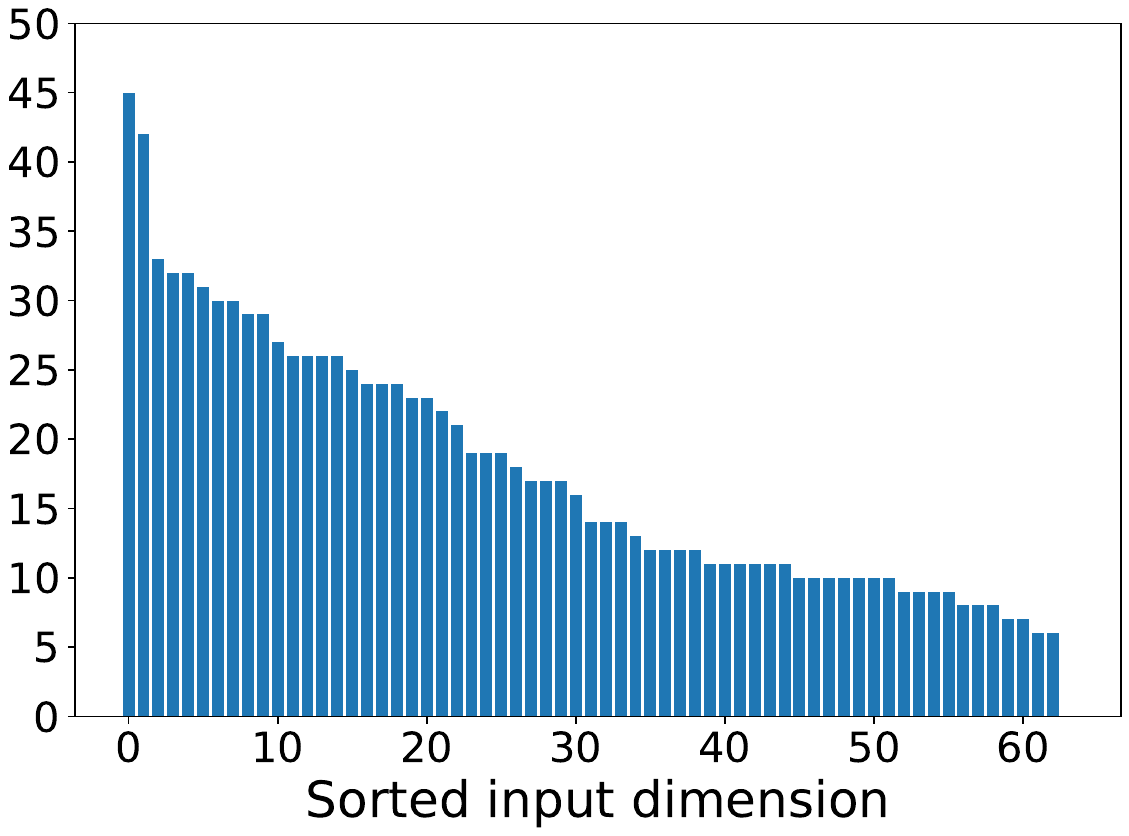}}    
\end{figure}
\vspace{-1.1cm}
\end{document}